\documentclass[12pt]{article}
\usepackage{amsmath}
\usepackage{graphicx,psfrag,epsf,color,subfigure}
\usepackage{enumerate}
\usepackage{natbib}

\usepackage[unicode=true,pdfusetitle,
   bookmarks=true, 
   bookmarksnumbered=false, 
   bookmarksopen=false,
   breaklinks=false, 
   pdfborder={0 0 1}, 
   backref=false, 
   colorlinks=false]{hyperref}

\usepackage{amssymb}
\usepackage{multirow}
\usepackage{array}
\usepackage{bm}
\usepackage{booktabs}
\usepackage{float}
\usepackage[font=small,labelfont=bf]{caption}
\setcitestyle{citesep={;}}
\usepackage[total={6.4in,8.6in}]{geometry}
\newcommand{\Mean}{{\mathbb{E}}}
\newcommand{\Var}{{\mbox{Var}}}

\newcommand{\prob}{{\mbox{Pr}}}
\newtheorem{coro}{Corollary}

\newtheorem{thm}{Theorem}  
\newtheorem{lemma}{Lemma}[section]

\newtheorem{assumption}{Assumption}[section]

\DeclareMathOperator*{\argmax}{arg\,max}

\usepackage{authblk}

\usepackage{algorithm}
\usepackage{algorithmicx}
\usepackage{algpseudocode}

\begin{document}

\def\spacingset#1{\renewcommand{\baselinestretch}%
{#1}\small\normalsize} \spacingset{1}
%
%
\title{\bf \Large Doubly Robust Interval Estimation for Optimal Policy Evaluation in Online Learning}
\author[1]{Ye Shen\thanks{Equal contribution.} } 
\author[2]{Hengrui Cai$^*$ } 
\author[1]{Rui Song }
\affil[1]{Department of Statistics, North Carolina State University} 
\affil[2]{Department of Statistics, University of California Irvine} 
 \date{}
 \maketitle  

\baselineskip=21pt

\begin{abstract} 
Evaluating the performance of an ongoing policy plays a vital role in many areas such as medicine and economics, to provide crucial instructions on the early-stop of the online experiment and timely feedback from the environment. Policy evaluation in online learning thus attracts increasing attention by inferring the mean outcome of the optimal policy (i.e., the value) in real-time. Yet, such a problem is particularly challenging due to the dependent data generated in the online environment, the unknown optimal policy, and the complex exploration and exploitation trade-off in the adaptive experiment. In this paper, we aim to overcome these difficulties in policy evaluation for online learning. We explicitly derive the probability of exploration that quantifies the probability of exploring non-optimal actions under commonly used bandit algorithms. We use this probability to conduct valid inference on the online conditional mean estimator under each action and develop the \underline{d}oubly \underline{r}obust int\underline{e}rv\underline{a}l esti\underline{m}ation (DREAM) method to infer the value under the estimated optimal policy in online learning. The proposed value estimator provides double protection for consistency and is asymptotically normal with a Wald-type confidence interval provided. Extensive simulation studies and real data applications are conducted to demonstrate the empirical validity of the proposed DREAM method.
\end{abstract}


\section{Introduction}\label{sec:1}

Sequential decision-making is one of the essential components of modern artificial intelligence that considers the dynamics of the real world. By maintaining the trade-off between exploration and exploitation based on historical information, bandit algorithms aim to maximize the cumulative outcome of interest and are thus popular in dynamic decision optimization with a wide variety of applications, such as precision medicine \citep{lu2021bandit} and dynamic pricing \citep{turvey2017optimal}. There has been a vast literature on bandit optimization established over recent decades \citep[see e.g., ][ and the references therein]{sutton2018reinforcement, lattimore2020bandit}. Most of these theoretical works focus on the regret analysis of bandit algorithms. 
When properly designed and implemented to address the exploration-and-exploitation trade-off, a powerful bandit policy could achieve a sub-linear regret, and thus eventually approximate the underlying optimal policy that maximizes the expected outcome. However, such a regret analysis only shows the convergence rate of the averaged cumulative regret  (difference between the outcome under the optimal policy and that under the bandit policy)  but provides limited information on the {expected outcome under this bandit policy} (referred to as the value in 
\citet{dudik2011doubly}). 

\begin{figure}[!t]
\centering
  
\begin{subfigure} 
  \centering
  \includegraphics[width=.5\textwidth]{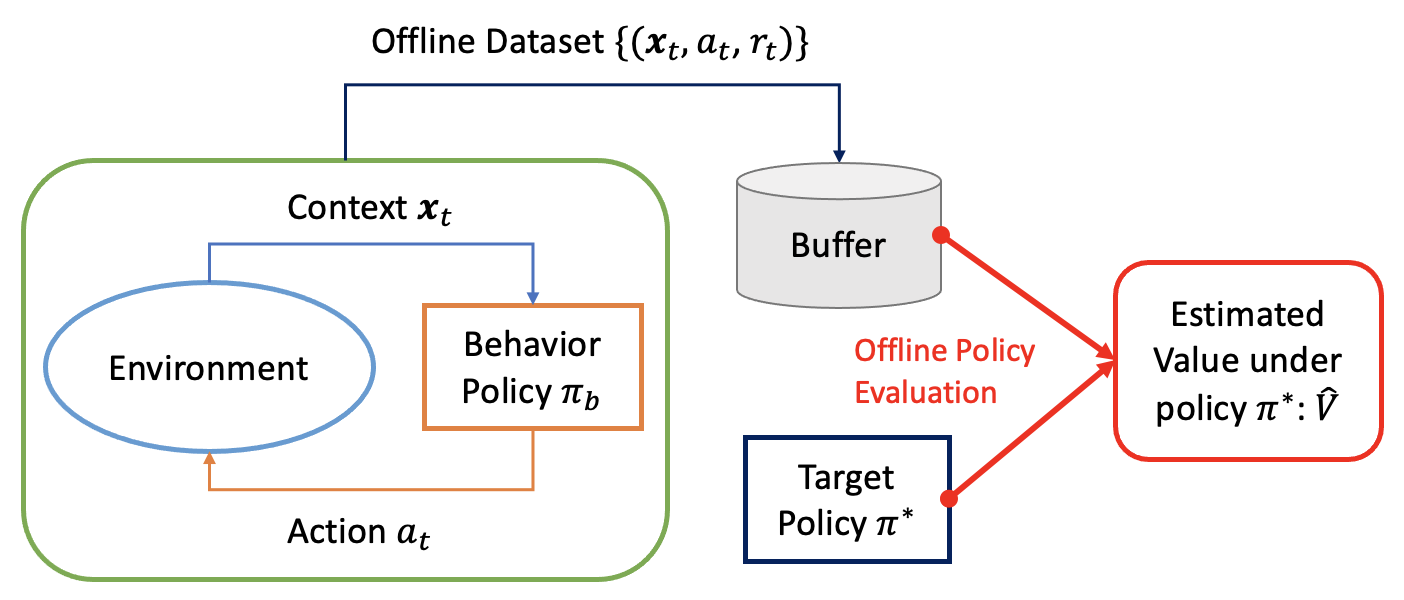}
\end{subfigure}  
\begin{subfigure} 
  \centering
  \includegraphics[width=.48\textwidth]{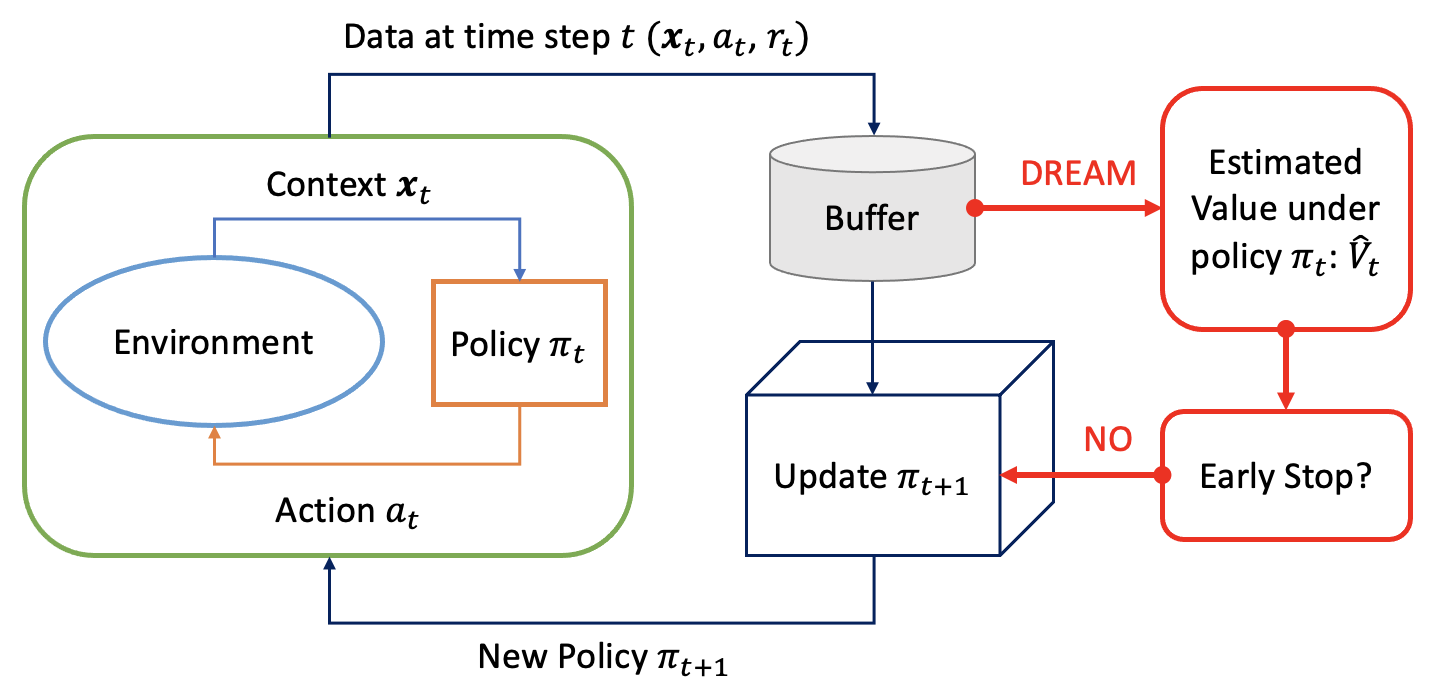}
\end{subfigure} 
  
       \caption{Left panel: the architecture of offline policy evaluation, with offline context-action-outcome triples $\{({\boldsymbol{x}_t},a_t,r_t)\}$ stored in the buffer to learn the value under a target policy $\pi^*$ with data generated by a behavior policy $\pi_b$. Right panel:
       the architecture of doubly robust interval estimation (DREAM) method for policy evaluation in online learning, where the context-action-outcome triple at time $t$, $({\boldsymbol{x}_t},a_t,r_t)$, is stored in the buffer to update the bandit policy $\pi_t$ and in the meantime to evaluate its performance.}
    \label{fig_illu} 
      
\end{figure}

The evaluation of the performance of bandit policies plays a vital role in many areas, including medicine and economics \citep[see e.g., ][]{chakraborty2013statistical,athey201921}. By evaluation, we aim to unbiasedly estimate the value of the optimal policy that the bandit policy is approaching and infer the corresponding estimate. Although there is an increasing trend in policy evaluation \citep[see e.g., ][]{li2011unbiased,dudik2011doubly,swaminathan2017off,wang2017optimal,kallus2018policy,su2019doubly}, we note that all of these works focus on learning the value of a target policy offline using historical log data. See the architecture of offline policy evaluation illustrated in the left panel of Figure \ref{fig_illu}. Instead of a post-experiment investigation,  it has attracted more attention recently to evaluate the ongoing policy in real-time.  In precision medicine, the physician aims to make the best treatment decision for each patient sequentially according to their baseline covariates. Estimating the mean outcome of the current treatment decision rule is crucial to answering several fundamental questions in health care, such as whether the current strategy significantly optimizes patient outcomes over some baseline strategies.  When the value under the ongoing rule is much lower than the desired average curative effect, the online trial must be terminated until more effective treatment options are available for the next round. Thus, policy evaluation in online learning is a new idea to provide the early stop of the online experiment and timely feedback from the environment,  as demonstrated in the right panel of Figure \ref{fig_illu}.

\subsection{Related Works and Challenges}\label{sec:challenges}
  
Despite the importance of policy evaluation in online learning, the current bandit literature suffers from three main challenges. 
First, the data, such as the actions and rewards sequentially collected from the online environment, are not independent and identically distributed (i.i.d.) since they depend on the previous history and the running policy (see the right panel of Figure \ref{fig_illu}). In contrast, the existing methods for the offline policy evaluation \citep[see e.g., ][]{li2011unbiased,dudik2011doubly} primarily assumed that the data are generated by the same behavior policy and i.i.d. across different individuals and time points. Such assumptions allow them to evaluate a new policy using offline data by modeling the behavior policy or the conditional mean outcome.   In addition, we note that the target policy to be evaluated in offline policy evaluation is fixed and generally known, whereas for policy evaluation in online learning, {the optimal policy of interest needs to be estimated and updated in real time}. 

The second challenge lies in {estimating the mean outcome under the optimal policy online}. Although numerous methods have recently been proposed to evaluate the online sample mean for a fixed action \citep[see e.g., ][]{nie2018adaptively,neel2018mitigating,deshpande2018accurate,shin2019sample,shin2019bias,waisman2019online,hadad2019confidence,zhang2020inference}, none of these methods is directly applicable to our problem, as the sample mean only provides the impact of one particular arm, not the value of the optimal policy in bandits that considers the dynamics of the online environment. For instance, in the contextual bandits, we aim to select an action for each subject based on its context/feature to optimize the overall outcome of interest. However, there may not exist a unified best action for all subjects due to heterogeneity, and thus evaluating the value of one single optimal action cannot fully address the policy evaluation in such a setting. However, although commonly used in the regret analysis, the average of collected outcomes is not a good estimator of the value under the optimal policy in the sense that it does not possess statistical efficiency (see details in Section \ref{sec:DREAMinf}).  

Third, given data generated by a bandit algorithm that maintains the exploration-and-exploitation trade-off sequentially, inferring the value of the optimal policy online should consider such a trade-off and quantify the probability of exploration and exploitation. The probability of exploring non-optimal actions is essential  in two ways. First, it determines the convergence rate of the online conditional mean estimator under each action. Second, it indicates the data points used to match the value under the optimal policy. To our knowledge, the regret analysis in the current bandit literature is based on the binding of variance information \citep[see e.g., ][]{auer2002using,srinivas2009gaussian,chu2011contextual,abbasi2011improved,bubeck2012regret,zhou2015survey}, yet little effort has been made in formally quantifying the probability of exploration over time.

There are very few studies directly related to our topic. \cite{chambaz2017targeted} established the asymptotic normality for the conditional mean outcome under an optimal policy for sequential decision making.  Later, \cite{chen2020statistical} proposed an inverse probability weighted value estimator to infer the value of optimal policy using the $\epsilon$-Greedy (EG) method. These two works did not discuss how to account for the exploration-and-exploitation trade-off under commonly used bandit algorithms, such as Upper Confidence Bound (UCB) and Thompson Sampling (TS), as considered in this paper.  Recently, to evaluate the value of a known policy based on the adaptive data, 
\citet{bibaut2021post} and \citet{zhan2021off} proposed to utilize the  stabilized doubly robust estimator and the adaptive weighting doubly robust estimator, respectively. However, both methods focused on obtaining a valid inference of the value estimator under a fixed policy by conveniently assuming a desired exploration rate to ensure sufficient sampling of different arms. Such an assumption can be violated in many commonly used bandits (see details shown in Theorem \ref{BoundPE}). Although there are other works that focus on statistical inference for adaptively collected data \citep{dimakopoulou2021online,zhang2021statistical,khamaru2021near,ramprasad2022online} in bandit or reinforcement learning setting, our work handles policy evaluation from a completely unique angle to infer the value of optimal policy by investigating the exploration rate in online learning.

\subsection{Our Contributions} \label{sec:contribution}

In this paper, we aim to overcome the aforementioned difficulties of policy evaluation in online decision-making. 
Our contributions are expressed in the following folds.

\textbf{The first contribution of this work is to explicitly characterize the trade-off between exploration and exploitation in the online policy optimization,  we derive the probability of exploration in bandit algorithms}. Such a probability is new to the literature by quantifying the chance of taking the nongreedy policy (i.e., a nonoptimal action) given the current information over time, in contrast to the probability of exploitation for taking greedy actions. Specifically, we consider three commonly used bandit algorithms for exposition, including the UCB, TS, and EG methods. We note that the probability of exploration is prespecified by users in EG while remaining implicit in UCB and TS.  We use this probability to conduct valid inferences on the online conditional mean estimator under each action. 
\textbf{The second contribution of this work is to propose the \underline{d}oubly \underline{r}obust int\underline{e}rv\underline{a}l esti\underline{m}ation (DREAM) to infer the mean outcome of the optimal online policy}. The DREAM provides double protection on the consistency of the proposed value estimator to the true value, given the product of the nuisance error rates of the probability of exploitation and the conditional mean outcome as $o_p(T^{-1/ 2})$ for $T$ as the termination time. Under standard assumptions for inferring the online sample mean, we show that the value estimator under DREAM is asymptotically normal with a Wald-type confidence interval provided.  To the best of our knowledge, this is the first work to establish the inference for the value under the optimal policy by taking the exploration-and-exploitation trade-off into thorough account and thus fills a crucial gap in the policy evaluation of online learning.

The remainder of this paper is organized as follows. We introduce notation and formulate our problem, followed by preliminaries of standard contextual bandit algorithms and the formal definition of probability of exploration.  
In Section \ref{sec:DREAMinf}, we introduce the DREAM method and its implementation details. 
Then in Section \ref{sec:4}, we derive theoretical results of DREAM under contextual bandits by establishing the convergence rate of the probability of exploration and deriving the asymptotic normality of the online mean estimator. 
Extensive simulation studies are conducted to demonstrate the empirical performance of the proposed method in Section \ref{sec:simu}, followed by a real application using OpenML datasets in Section \ref{sec:realdata}. We conclude our paper in Section \ref{sec:discussion} by discussing the performance of the proposed DREAM in terms of the regret bound and a direct extension of our method for  policy evaluation of any known policy in online learning. All the additional results and technical proofs are given in the appendix.

 \section{Problem Formulation}\label{sec:2}
  
 In this section, we formulate the problem of policy evaluation in online learning. We first build the framework based on the contextual bandits in Section \ref{sec:frame}. 
 We then introduce three commonly used bandit algorithms, including UCB, TS, and EG, to generate data online, in Section \ref{sec:bandits}. Lastly, we define the probability of exploration in Section \ref{sec:prob_expl}.  
In this paper, we use bold symbols for vectors and matrices.
 
 \subsection{Framework}\label{sec:frame}
  
In contextual bandits, at each time step $t \in \mathcal{T}\equiv\{1,2,3,\cdots\}$, we observe a $d$-dimensional context $\boldsymbol{x}_t$ drawn from a distribution $P_{\mathcal{X}}$  which includes 1 for the intercept, choose an action $a_t\in \mathcal{A}$, and then observe a reward $r_t \in \mathcal{R}$.  Denote the history observations prior to time step $t$ as $\mathcal{H}_{t-1}=\{\boldsymbol{x}_i,a_i,r_i\}_{1\leq i\leq t-1}$. Suppose that the reward given $\boldsymbol{x}$ and $a$ follows $r \equiv \mu(\boldsymbol{x} ,a)+e$, where $\mu(\boldsymbol{x},a)\equiv \mathbb{E}(r| \boldsymbol{x}, a)$ is the conditional mean outcome function (also known as the Q-function in the literature \citep{murphy2003optimal,sutton2018reinforcement}), and is bounded by $|\mu(\boldsymbol{x},a)| \leq U$. The noise term $e$ is independent $\sigma$-subgaussian at the time step $t$ independently of $\mathcal{H}_{t-1}$ given $a_t$  for $t \in  \mathcal{T} $. 
Let the conditional variance be $\Mean(e^2|a) = \sigma^2_a$. The value \citep{dudik2011doubly} of a given policy $\pi(\cdot)$ is defined as
\begin{equation*}
 V(\pi)\equiv \mathbb{E}_{\boldsymbol{x}  \sim P_{\mathcal{X}}}\left[\mathbb{E}\{r|\boldsymbol{x}, a =\pi(\boldsymbol{x} )\}\right] =\mathbb{E}_{\boldsymbol{x}  \sim P_{\mathcal{X}}}\left[\mu\{ \boldsymbol{x} ,  \pi(\boldsymbol{x} )\}\right] .
\end{equation*}
We define the optimal policy as  $\pi^*(\boldsymbol{x}) \equiv \argmax_{a\in \mathcal{A}} \mu(\boldsymbol{x },a),   \forall \boldsymbol{x} \in \mathcal{X}$, which finds the optimal action based on the conditional mean outcome function given a context $\boldsymbol{x}$. 
Thus, the optimal value can be defined as $V^*\equiv V(\pi^*)=\mathbb{E}_{\boldsymbol{x}  \sim P_{\mathcal{X}}}\left[\mu\{ \boldsymbol{x} ,  \pi^*(\boldsymbol{x} )\}\right]$. In the rest of this paper, to simplify the exposition, we focus on two actions, that is, $\mathcal{A}=\{0,1\}$.  Then the optimal policy is given by
\begin{equation*}
\pi^*(\boldsymbol{x}) \equiv \argmax_{a\in \mathcal{A}} \mu(\boldsymbol{x},a) = \mathbb{I}\{\mu(\boldsymbol{x},1)>\mu(\boldsymbol{x},0)\}, \quad \forall \boldsymbol{x} \in \mathcal{X}.
\end{equation*}  
Our goal is to infer the value under the optimal policy $\pi^* $ using the online data sequentially generated by a bandit algorithm. Since the optimal policy is unknown, we estimate the optimal policy from the online data as $\widehat{\pi}_t$.
As commonly assumed in the current online inference literature \citep[see e.g., ][]{deshpande2018accurate,zhang2020inference,chen2020statistical} and the bandit literature \citep[see e.g., ][]{chu2011contextual,abbasi2011improved,bubeck2012regret,zhou2015survey}, we consider the conditional mean outcome function taking a linear form, i.e., 
$\mu(\boldsymbol{x} ,a)=\boldsymbol{x}^\top\boldsymbol{\beta}(a),$ 
 where $\boldsymbol{\beta}(\cdot)$ is a smooth function and can be estimated via a ridge regression based on $\mathcal{H}_{t-1}$ as
\begin{equation} \label{eq:betahat}
 \widehat{\boldsymbol{\beta}}_{t-1}(a) = \{\boldsymbol{D}_{t-1}(a)^\top \boldsymbol{D}_{t-1}(a) +\omega \boldsymbol{I}_d\}^{-1}\boldsymbol{D}_{t-1}(a)^\top \boldsymbol{R}_{t-1}(a),
 \end{equation}  
 where $\boldsymbol{I}_d$ is a $d \times d$ identity matrix, $\boldsymbol{D}_{t-1}(a)$ is a  $\boldsymbol{N}_{t-1}(a) \times d$ design matrix at time $t-1$ with $\boldsymbol{N}_{t-1}(a)$ as the number of pulls for action $a$, $\boldsymbol{R}_{t-1}(a)$ is the $\boldsymbol{N}_{t-1}(a) \times 1$ vector of the outcomes received under action $a$  at time $t-1$, and $\omega$ is a positive and bounded constant as the regularization term.   There are two main reasons to choose the ridge estimator instead of the
ordinary least squares estimator that is considered in \cite{deshpande2018accurate,zhang2020inference,chen2020statistical}. First, the ridge estimator is well defined when $\boldsymbol{D}_{t-1}(a)^\top \boldsymbol{D}_{t-1}(a)$ is singular and its bias is negligible when the time step is large. 
  Second, the parameter estimations in the ridge method are in accordance with the linear UCB \citep{li2010contextual} and the linear TS \citep{agrawal2013thompson} methods (detailed in the next section) with $\omega=1$. 
  Based on the ridge estimator in \eqref{eq:betahat}, the online conditional mean estimator for $\mu$ is defined as $\widehat{\mu}_{t-1}(\boldsymbol{x},a) =\boldsymbol{x}^\top \widehat{\boldsymbol{\beta}}_{t-1}(a) $. With two actions, the estimated optimal policy at time step $t$ is defined by
\begin{equation} \label{eq:pihat}
\widehat{\pi}_t(\boldsymbol{x})=\mathbb{I}\left\{\widehat{\mu}_{t -1}(\boldsymbol{x},1)> \widehat{\mu}_{t -1}(\boldsymbol{x},0)\right\} = \mathbb{I} \{\boldsymbol{x}^\top\widehat{\boldsymbol{\beta}}_{t -1}(1)> \boldsymbol{x}^\top\widehat{\boldsymbol{\beta}}_{t -1}(0) \}, \quad \forall \boldsymbol{x} \in \mathcal{X}.
\end{equation} 
We note that the linear form of $\mu(\boldsymbol{x} ,a)$ 
can be relaxed to the non-linear case as $\mu(\boldsymbol{x} ,a)=f(\boldsymbol{x})^\top\boldsymbol{\beta}(a)$, where $f(\cdot)$ is a continuous function (see examples in our simulation studies in Section \ref{sec:simu}). Then the corresponding online conditional mean estimator for $\mu$ is defined as $  \widehat{\mu}_{t-1}(\boldsymbol{x},a) =f(\boldsymbol{x})^\top \widehat{\boldsymbol{\beta}}_{t-1}(a)$ based on $\mathcal{H}_{t-1}$.

  \subsection{Bandit Algorithms}\label{sec:bandits}
     
We briefly introduce three commonly used bandit algorithms in the framework of contextual bandits, to generate the online data sequentially.\\
\textbf{Upper Confidence Bound (UCB) \citep{li2010contextual}:} Let the estimated standard deviation based on $\mathcal{H}_{t-1}$ be $\widehat{\sigma}_{t-1}(\boldsymbol{x},a) = \sqrt{\boldsymbol{x}^\top \{\boldsymbol{D}_{t-1}(a)^\top\boldsymbol{D}_{t-1}(a) +\omega  \boldsymbol{I}_d\}^{-1}\boldsymbol{x}}.$ The action at time $t$ is selected by  
\begin{equation*}
a_{t} = \argmax_{a\in \mathcal{A}} \widehat{\mu}_{t -1}(\boldsymbol{x}_t ,a)+c_t \widehat{\sigma}_{t-1}(\boldsymbol{x}_t ,a),
\end{equation*}
where $c_t$ is a non-increasing positive parameter that controls the level of exploration. 
With two actions, we have the action at time step $t$ as \begin{equation*}
a_{t} =\mathbb{I}\left\{\widehat{\mu}_{t -1}(\boldsymbol{x}_t ,1)+c_t \widehat{\sigma}_{t-1}(\boldsymbol{x}_t ,1) > \widehat{\mu}_{t -1}(\boldsymbol{x}_t ,0)+c_t \widehat{\sigma}_{t-1}(\boldsymbol{x}_t ,0)\right\}.
\end{equation*}
\textbf{Thompson Sampling (TS) \citep{agrawal2013thompson}:}
Suppose a normal likelihood function for the reward given $\boldsymbol{x}$ and $a$  such that $R \sim \mathcal{N}\{\boldsymbol{x}^\top\boldsymbol{\beta}(a), \rho^2\}$ with a known parameter $\rho^2$. If the prior for $\boldsymbol{\beta}(a)$ at time $t$ is 
\begin{equation*}
\boldsymbol{\beta}(a)\sim \mathcal{N}_d [\widehat{\boldsymbol{\beta}}_{t-1}(a), \rho^2\{\boldsymbol{D}_{t-1}(a)^\top\boldsymbol{D}_{t-1}(a) +\omega \boldsymbol{I}_d\}^{-1}],
\end{equation*}
where $\mathcal{N}_d$ is  the $d$-dimensional multivariate normal distribution, we have the posterior distribution of $\boldsymbol{\beta}(a)$ as 
\begin{equation*}
\boldsymbol{\beta}(a)|\mathcal{H}_{t-1} \sim \mathcal{N}_d[\widehat{\boldsymbol{\beta}}_{t}(a), \rho^2\{\boldsymbol{D}_{t}(a)^\top\boldsymbol{D}_{t}(a) +\omega \boldsymbol{I}_d\}^{-1}],
\end{equation*}
for $a\in \{0,1\}$. 
At each time step $t$, we draw a sample from the posterior distribution as $\boldsymbol{\beta}_t(a)$ for $a\in \{0,1\}$, 
and select the next action within two arms by $a_{t} =\mathbb{I}\left\{\boldsymbol{x}_t^\top\boldsymbol{\beta}_t(1) > \boldsymbol{x}_t^\top\boldsymbol{\beta}_t(0) \right\}.$\\
\textbf{$\epsilon$-Greedy (EG) \citep{sutton2018reinforcement}:} Recall the estimated conditional mean outcome for action $a$ as $\widehat{\mu}_t(\boldsymbol{x},a)$ given a context $\boldsymbol{x}$. 
Under EG method, the action at time $t$ is selected by  \begin{equation*}
a_{t}  = \delta_t \argmax_{a\in \mathcal{A}} \widehat{\mu}_{t-1}(\boldsymbol{x}_t ,a) + (1-\delta_t)\text{Bernoulli}(0.5),
\end{equation*}
where $\delta_t\sim \text{Bernoulli}(1-\epsilon_t)$ and the parameter $\epsilon_t$ controls the level of exploration as pre-specified by users.

\subsection{Probability of Exploration}\label{sec:prob_expl}
  
We next quantify the probability of exploring non-optimal actions at each time step.  To
be specific, define the status of exploration as $\mathbb{I}\{a_t\not = \widehat{\pi}_t(\boldsymbol{x}_t )\}$,
indicating whether the action taken by the bandit algorithm is different from the estimated optimal action that exploits the historical information, given the context information. Here, $\widehat{\pi}_t$ can be viewed as the greedy policy at time step $t$. Thus the probability of exploration is defined by 
 \begin{equation}\label{def_kappa}
\kappa_t(\boldsymbol{x}_t)\equiv  \prob\{a_t\not = \widehat{\pi}_t(\boldsymbol{x}_t )\} =  \Mean [ \mathbb{I}\{a_t\not = \widehat{\pi}_t(\boldsymbol{x}_t )\}],
\end{equation}
where the expectation in the last term is taken respect to $a_t \in \mathcal{A}$ and history $\mathcal{H}_{t-1}$.  According to \eqref{def_kappa}, $\kappa_t$ is determined by the given context information and the current time point. We clarify the connections and distinctions of the defined probability of exploration concerning the bandit literature here. First, the exploration rate used in the current bandit works mainly refers to the probability of exploring non-optimal actions given an optimal policy  and contextual information $\boldsymbol{x}_t $ at time step $t$, i.e., $\prob\{a_t\not = \pi^*(\boldsymbol{x}_t)\}$. This is different from our proposed probability of exploration $\kappa_t(\boldsymbol{x}_t) \equiv \prob\{a_t\not = \widehat{\pi}_t(\boldsymbol{x}_t )\}$. 
The main difference lies in the estimated optimal policy from the collected data at the current time step $t$, i.e., $\widehat{\pi}_t$, in $\kappa_t(\boldsymbol{x}_t)$. If properly designed and implemented, the estimated optimal policy under a bandit algorithm $\widehat{\pi}_t$ will eventually converge to the true optimal policy $\pi^*$ when $t$ is large, which yields $\kappa_t (\boldsymbol{x} ) \to \lim_{t\rightarrow \infty}\prob\{a_t\not = \pi^*(\boldsymbol{x})\}$. In practice, we can estimate $\kappa_t(\boldsymbol{x}_t) $ sequentially by using  $\{i,\boldsymbol{x}_i \}_{1\leq i\leq t}$ as inputs and $\{ \mathbb{I}\{a_i = \widehat{\pi}_i(\boldsymbol{x}_i)\}  \}_{1\leq i\leq t}$ as outputs via parametric or non-parametric tools. We denote the corresponding estimator as $\widehat{\kappa}_t (\boldsymbol{x}_t)$ for time step $t$.  Such implementation conditions are explicitly described in Section \ref{sec:DREAMinf}.

\section{Doubly Robust Interval Estimation}\label{sec:DREAMinf}

We present the proposed DREAM method in this section. We first detail why the average of outcomes received in bandits fails to process statistical efficiency. In view of the results established in regret analysis for the contextual bandits \citep[see e.g., ][]{abbasi2011improved,chu2011contextual,zhou2015survey}, we have the cumulative regret as $ |\sum_{t=1}^T  (r_t-  V^*) |=\tilde{\mathcal{O}}(\sqrt{dT})$,  where $\tilde{\mathcal{O}}$ is the asymptotic order up to some logarithm factor. Therefore, it is immediate that the average of rewards follows ${\sqrt{T}}( T^{-1}\sum_{t=1}^T r_t - V^*  )= \tilde{\mathcal{O}}(1)$, since the dimension $d$ is finite. These results indicate that a simple average of the total outcome under the bandit algorithm is not a good estimator for the optimal value, since it does not own the asymptotic normality for a valid confidence interval construction. Instead of the simple aggregation, intuitively, we should select the outcome when the action taken under a bandit policy accords with the optimal policy, i.e., $a_t=\widehat{\pi}_t(\boldsymbol{x}_t )$, defined as the status of exploitation. In contrast to the probability of exploration, we define the probability of exploitation as
 \begin{equation*}
\prob\{a_t = \widehat{\pi}_t(\boldsymbol{x}_t )\} = 1-{\kappa}_t (\boldsymbol{x}_t ) =  \Mean[ \mathbb{I}\{a_t = \widehat{\pi}_t(\boldsymbol{x}_t )\}].
\end{equation*}   
Following the doubly robust value estimator in \cite{dudik2011doubly}, we propose the doubly robust mean outcome estimator as the value estimator under the optimal policy as
 \begin{equation}\label{dr_est}
\widehat{V}_T=\frac{1}{T}\sum_{t=1}^T \frac{\mathbb{I}\{a_t=\widehat{\pi}_t(\boldsymbol{x}_t )\}}{1-\widehat{\kappa}_{t}(\boldsymbol{x}_t )} \Big[ r_t -\widehat{\mu}_{t-1}\{\boldsymbol{x}_t ,\widehat{\pi}_t(\boldsymbol{x}_t )\}\Big] + \widehat{\mu}_{t-1}\{\boldsymbol{x}_t ,\widehat{\pi}_t(\boldsymbol{x}_t ) \},
\end{equation}
where $T$ is the current/termination time, and $1-\widehat{\kappa}_{t}(\boldsymbol{x}_t )$ is the estimated matching probability between the chosen action $a_t$ and estimated optimal action given $\boldsymbol{x}_t $, which captures the probability of exploitation.  
Our value estimator provides double protection on the consistency to the true value, given the product of the nuisance error rates of the probability of exploitation and the conditional mean outcome as $\mathcal{O}(T^{-1/ 2})$, as discussed in Theorem \ref{thm:asym}. We further propose a variance estimator for $\widehat{V}_T$ as 
\begin{equation} \label{eq:sigmahat}
\resizebox{1\hsize}{!}{$
\begin{aligned}
\widehat{\sigma}_{T}^2  & = \frac{1}{T}\sum_{t=1}^{T} \left( \frac{\widehat{\pi}_{t}(\boldsymbol{x}_t )  \widehat{\sigma}_{1,t-1}^2  +\{1-\widehat{\pi}_{t}(\boldsymbol{x}_t ) \}\widehat{\sigma}_{0,t-1}^2}{1-\widehat{\kappa}_{t}(\boldsymbol{x}_t)}+  \left[  \widehat{\mu}_{T} \{\boldsymbol{x}_t ,\widehat{\pi}_{T}(\boldsymbol{x}_t ) \} -  \frac{1}{T}\sum_{t=1}^{T} \widehat{\mu}_{T} \{\boldsymbol{x}_t ,\widehat{\pi}_{T}(\boldsymbol{x}_t ) \}  \right]^2\right),
 \end{aligned}$}
\end{equation}
where $ \widehat{\sigma}_{a,t}^2 = \{\sum_{i=1}^{t}\mathbb{I}(a_i =a)-d\}^{-1}\sum_{a_i =a}^{1\leq i\leq t}[\widehat{\mu}_{t} \{\boldsymbol{x}_i ,a_i\} - r_i]^2$ is an estimator for $\sigma^2_a$, for $a=0,1$.  The proposed variance estimator is consistent to the true variance of the value shown in Theorem \ref{thm:asym}. 
We officially name our method as \underline{d}oubly \underline{r}obust int\underline{e}rv\underline{a}l esti\underline{m}ation (DREAM), 
with the detailed pseudocode provided in Algorithm \ref{algo_cb}. 
To ensure sufficient exploration and a valid inference, we force to pull non-optimal actions given greedy samples
with a pre-specified clipping rate $p_t$ in step (4) of Algorithm \ref{algo_cb} of an order larger than $\mathcal{O}(t^{-1/2})$ as required by Theorem \ref{thm:asym}. In step (4), if the unchosen action $1-a_t$ does not satisfy the clipping condition detailed in Assumption \ref{Clipping}, then we will force to pull this non-greedy action for additional exploration, as required for a valid online inference. 
The estimated probability of exploration is used in step (6) in Algorithm \ref{algo_cb} to learn the value and variance. A potential limitation of DREAM rises when the agent resists taking extra non-greedy actions in online learning. Yet, we discuss in Section \ref{sec:reg} that the regret from such an extra exploration is negligible compared to the regret from exploitation, thus we can still maintain the sub-linear regret under DREAM. The theoretical validity of the proposed DREAM is detailed in Section \ref{sec:4}, with its empirical out-performance over baseline methods demonstrated in Section \ref{sec:simu}. Finally, we remark that our method is not overly sensitive to the choice of $p_t$ with additional sensitivity analyses provided in Appendix A.

 \begin{algorithm}[th]  
   \caption{DREAM under Contextual Bandits with Clipping}\label{algo_cb}
  \begin{algorithmic}
    \State \textbf{Input}: termination time $T$, and the clipping rate $p_t>\mathcal{O}(t^{-1/2})$;
    \For{Time $t = 1,2,\cdots T$}
       \State (1) Sample $d$-dimensional context $\boldsymbol{x}_t \in \mathcal{X}$;
       \State (2) Update $\widehat{\boldsymbol{\beta}}_{t-1}(\cdot)$ using Equation \eqref{eq:betahat} and $\widehat{\mu}_{t-1}(\cdot,\cdot)$;
       \State (3) Update $\widehat{\pi}_t(\cdot)$ and $a_t$ using Equation \eqref{eq:pihat} and  the contextual bandit algorithms in Section \ref{sec:bandits};
      {\If{$\lambda_{\min}({t}^{-1}\sum_{i=1}^{t} \mathbb{I}(a_i = 1- a_t) \boldsymbol{x}_i \boldsymbol{x}_i^{\top})$ $< p_t \lambda_{\min}({t}^{-1} \sum_{i=1}^{t}  \boldsymbol{x}_i \boldsymbol{x}_i^{\top})$}
     	\State (4)  Choose action $1- a_t$;
      \EndIf}
     
         \State  \hspace{-0.8cm} (5) Use the history $\{\mathbb{I}\{\widehat{\pi}_i(\boldsymbol{x}_i)=a_i\},\boldsymbol{x}_i \}_{1\leq i\leq t}$ to estimate $\widehat{\kappa}_t(\cdot)$;
       \State \hspace{-0.8cm} (6) Get the value and its variance under the optimal policy by Equations \eqref{dr_est} and \eqref{eq:sigmahat}.
       \State \hspace{-0.8cm} (7)  A two-sided $1-\alpha$ CI for $V^{*}$ under the online optimization is given by \\
       \hspace{1cm} $\Big [ \widehat{V}_T -z_{\alpha/2}\widehat{\sigma}_{T}/\sqrt{T},\quad \widehat{V}_T+z_{\alpha/2}\widehat{\sigma}_{T}/\sqrt{T} \Big]$.
     \EndFor
 \end{algorithmic}  
\end{algorithm}

\section{Theoretical Results}\label{sec:4}

We formally present our theoretical results. 
In Section \ref{BoundPE}, we first derive the bound of the probability of exploration under the three commonly used bandit algorithms introduced in Section \ref{sec:bandits}. This allows us to further establish the asymptotic normality of the online conditional mean estimator under a specific action in Section \ref{sec:4.2}. Next, we establish the theoretical properties of DREAM with a Wald-type confidence interval given in Section \ref{sec:4.3}. All the proofs are provided in Appendix B. The following assumptions are required to establish our theories.

\begin{assumption}(Boundness) \label{ALx}
There exists a positive constant $L_{\boldsymbol{x}}$ such that $\|\boldsymbol{x}\|_{\infty} \leq L_{\boldsymbol{x}}$ for all $\boldsymbol{x} \in \mathcal{X}$, and $\boldsymbol{\Sigma}=\mathbb{E}\left(\boldsymbol{x} \boldsymbol{x} ^{\top}\right)$ has minimum eigenvalue $\lambda_{\min }(\boldsymbol{\Sigma})>\lambda$ for some $\lambda>0$.
\end{assumption}
\begin{assumption}(Clipping)\label{Clipping}  For any action $a \in \mathcal{A}$ and  time step $t \geq 1$, there exists an sequence positive and non-increasing $\{p_i\}_{i=1}^t$ , such that $ \lambda_{\min} \{  {t}^{-1} \sum_{i=1}^{t} \mathbb{I}(a_i = a) \boldsymbol{x}_i \boldsymbol{x}_i^{\top} \} > p_t\lambda_{\min }(\boldsymbol{\Sigma}).$
\end{assumption} 
\begin{assumption}(Margin Condition) \label{AMargin}
Assume there exist some constants $\gamma$ and $\delta$ such that $\prob\{0 \leq  |\mu(\boldsymbol{x},1) - \mu(\boldsymbol{x},0)|\leq M\} =\mathcal{O}(M^\gamma)$, $\forall \boldsymbol{x} \in \mathcal{X},$  where the big-O term is uniform in $0 < M \leq \delta$.
\end{assumption}

Assumption \ref{ALx} is a technical condition on bounded contexts such that the mean of the martingale differences will converge to zero \citep[see e.g., ][]{zhang2020inference,chen2020statistical}.   
Assumption \ref{Clipping} is a technical requirement for the uniqueness and convergence of the least squares estimators, which requires the bandit algorithm to explore all actions sufficiently such that the asymptotic properties for the online conditional mean estimator under different actions hold \citep[see e.g., ][]{deshpande2018accurate,hadad2019confidence,zhang2020inference}. The parameter  $\{p_i\}_{i=1}^t$ defined in Assumption \ref{Clipping} characterizes the boundary of the probability of taking one action and we name it as the clipping  rate. We ensure Assumption \ref{Clipping} is satisfied using step (4) in Algorithm \ref{algo_cb}.  We establish the relationship between $p_t$ and $\kappa_t$  and discuss the requirement for $p_t$ to consistently estimate $\widehat{\boldsymbol{\beta}}_{t}(a)$ in this section. Assumption \ref{AMargin} is well known as the margin condition, which is commonly assumed in the literature to derive a sharp convergence rate for the value under the estimated optimal policy \citep[see e.g., ][]{luedtke2016statistical,chambaz2017targeted,chen2020statistical}.

\subsection{Bounding the Probability of Exploration} \label{BoundPE}
 
The probability of exploration not only shows the signal of success in finding the global optimization in online learning, but also connects to optimal policy evaluation by quantifying the rate of executing the greedy actions. Instead of directly specifying this probability \citep[see e.g., ][]{zhang2020inference,chen2020statistical,bibaut2021post,zhan2021off}, we explicitly bound this rate based on the updating parameters in bandit algorithms, by which we conduct a valid follow-up inference. Since the probability of exploration involves the estimation of the mean outcome function, we need to first derive a tail bound for the online ridge estimator and the estimated difference between the mean outcomes.

\begin{lemma}{(Tail bound for the online ridge estimator)}.\label{thmTBOLS} In the online contextual bandit under UCB, TS, or EG, with  Assumptions \ref{ALx} and \ref{Clipping} hold,  we have that for any $h>0$, the probability of the online ridge estimator bounded within its true as 
 \begin{equation*}
{\prob \left( \left\|\widehat{\boldsymbol{\beta}}_t(a)-\boldsymbol{\beta}(a)\right\|_{1} > h \right)   \leq 2 d \exp \left\{-tp_t^{2}\lambda^{2} \left(h   - \sqrt{d}\left\| \boldsymbol{\beta}(a)\right\|_2\right)^{2}/\left(8d^{2} \sigma^{2} L_{\boldsymbol{x}}^{2}\right)\right\}.}
\end{equation*}
\end{lemma}
The results in Lemma \ref{thmTBOLS} establish the tail bound of the online ridge estimator $\widehat{\boldsymbol{\beta}}_t(a)$, which can be simplified as  $C_1 \exp (-C_2 tp_t^{2})$, for some constants $\{C_j\}_{1\leq j \leq 2}$, by noticing the constant $h$ and $\lambda$, the dimension $d$, the subgaussian parameter $\sigma$, and the bound $L_{\boldsymbol{x}}$ are positive and bounded, under bounded true coefficients $\boldsymbol{\beta}(a)$. Recall the clipping constraint that $0<p_t<1$ is non-increasing sequence. This tail bound is asymptotically equivalent to $\exp (- tp_t^{2})$. The established results in Lemma \ref{thmTBOLS} work for general bandit algorithms including UCB, TS, and EG. These tail bounds are aligned with the bound $\exp (- t \epsilon_t^{2})$ derived in \cite{chen2020statistical} for the EG method only. By Lemma \ref{thmTBOLS}, it is immediate to obtain the consistency of the online ridge estimator $\widehat{\boldsymbol{\beta}}_t(a)$, if $tp_t^{2}\rightarrow \infty$ as $t\rightarrow \infty$. 
We can further obtain the tail bound for the estimated difference between the conditional mean outcomes under two actions by Lemma \ref{thmTBOLS} as detailed in the following corollary. 

\begin{coro}{(Tail bound for the online mean estimator)} \label{coro:thmTBOLS} 
Suppose conditions in Lemma \ref{thmTBOLS} hold. Denote $\Delta_{\boldsymbol{x}_t} \equiv \mu(\boldsymbol{x}_t ,1)-\mu(\boldsymbol{x}_t ,0)$, then for any $\xi >0$, we have the probability of the online conditional mean estimator bounded within its true as 
\begin{equation*}
\prob \left\{  \left| \widehat{\mu}_t(\boldsymbol{x}_t,1)-\widehat{\mu}_t(\boldsymbol{x}_t,0) -  \Delta_{\boldsymbol{x}_t} \right|   > \xi \right \} \leq 4 d \exp \left\{-tp_t^{2} c_{\xi}\right\},
\end{equation*}
with 
\begin{equation*}
c_{\xi} = \lambda^{2} \left[ \min \left\{  \left( \xi /2    - \sqrt{d} L_{\boldsymbol{x}} \left\| \boldsymbol{\beta}(1)\right\|_2 \right)^{2} , \left( \xi /2    - \sqrt{d} L_{\boldsymbol{x}} \left\| \boldsymbol{\beta}(0)\right\|_2\right)^{2}\right\}  \right]/\left( 8d^{2} \sigma^{2} L_{\boldsymbol{x}}^{4}\right)
\end{equation*}
being a constant of time $t$.
\end{coro}
The above corollary quantifies the uncertainty of the online estimation of the conditional mean outcomes, and thus provides a crucial middle result to further access the probability of exploration by noting $\widehat{\pi}_t(\boldsymbol{x})=\mathbb{I}\left\{\widehat{\mu}_{t -1}(\boldsymbol{x},1)> \widehat{\mu}_{t -1}(\boldsymbol{x},0)\right\}$  in \eqref{eq:pihat}. More specifically, we derive the probability of exploration at each time step under the three discussed bandit algorithms for exposition in the following theorem.
\begin{thm}(Probability of exploration)\label{thm1}
In the online contextual bandit algorithms using UCB, TS, or EG, given a context $\boldsymbol{x}_t$ at time step $t$, assuming Assumptions \ref{ALx}, \ref{Clipping}, and \ref{AMargin}  hold  with $tp_t \rightarrow \infty$,  then for any $0<\xi <\left| \Delta_{\boldsymbol{x}_t} \right|/2 $ with  $c_{\xi}$ specified in Corollary \ref{coro:thmTBOLS},\\
(i) under UCB, there exists some constant $C>0$ such that 
\begin{eqnarray*}
  \kappa_t(\boldsymbol{x}_t) \leq {C \left( \frac{ 2 c_tL_{\boldsymbol{x}}}{\sqrt{(t-1)p_{t-1}\lambda }} + \xi  \right)^\gamma +  4 d \exp \left\{-(t-1)p_{t-1}^{2} c_{\xi}\right\}};
    \end{eqnarray*} 
(ii) under TS, we have 
\begin{eqnarray*}
\kappa_t(\boldsymbol{x}_t)  \leq {\exp\left(- \frac{\left(\left| \Delta_{\boldsymbol{x}_t} \right| - \xi   \right)^2  { {(t-1)p_{t-1} \lambda}}}{4\rho^2 L_{\boldsymbol{x}}^2 }\right)+  4 d \exp \left\{-(t-1)p_{t-1}^{2} c_{\xi}\right\}};
    \end{eqnarray*} 
(iii) under EG,  we have $ \kappa_t(\boldsymbol{x}_t)  = \epsilon_t/2$.
\end{thm}
The theoretical order of $\kappa_t(\cdot)$ is new to the literature, which quantifies the probability of exploring the non-optimal actions under a bandit policy over time. We note that the probability of exploration under EG is pre-specified by users, which directly implies its exploring rate as $\epsilon_t/2$ (for the two-arm setting) as shown in 
results (iii). The results (i) and (ii) in Theorem \ref{thm1} show that the probabilities of exploration under UCB and TS are non-increasing under certain conditions. For instance, if $(t-1)p_{t-1}^2 \rightarrow \infty$ for TS, as $t\rightarrow \infty$, the upper bound for $ \kappa_t(\boldsymbol{x}_t)$ decays to zero with an asymptotically equivalent convergence rate as $\mathcal{O}\{\exp(-(t-1)p_{t-1}^{2})\}$ up to some constant. Similarly, for UCB, with an arbitrarily small $\xi$ at an order of $\mathcal{O}(1/\sqrt{tp_t})$,  as $t\rightarrow \infty$, the upper bound for $ \kappa_t(\boldsymbol{x}_t)$ decays to zero as long as $(t-1)p_{t-1}^2 \rightarrow \infty$.
Theorem \ref{thm1} also indicates that without the clipping required in Assumption \ref{Clipping} and implemented in step (4) of Algorithm \ref{algo_cb}, the exploration for non-optimal arms might be insufficient, leading to a possibly invalid and biased inference for the online conditional mean.

  \subsection{Asymptotic Normality of Online Ridge Estimator}\label{sec:4.2}
   
We use the established bounds for the probability of exploration in Theorem \ref{thm1} to further obtain the asymptotic normality for the online conditional mean estimator under each action.
Specifically, denote $ \kappa_{\infty} (\boldsymbol{x})\equiv \lim_{t \to \infty}\kappa_t(\boldsymbol{x}) = \lim_{t\rightarrow \infty}\prob\{a_t\not = \pi^*(\boldsymbol{x})\}$ given a context $\boldsymbol{x}$. Assume the conditions in Theorem \ref{thm1} hold and   $tp_t^2 \rightarrow \infty$  as $t\rightarrow \infty$, we have the upper bound of the probability of exploration decays to zero in Theorem \ref{thm1} for UCB and TS. Since  $\kappa_t(\cdot)$ is nonnegative by its definition, it follows from Sandwich Theorem immediately that $ \kappa_{\infty} (\cdot)$  exist for UCB and TS, and $ \kappa_{\infty} (\cdot) = \lim_{t \to \infty} \epsilon_t/2$ for EG. By using this limit, we can further characterize 
the following theorem for asymptotics and inference. 
 
\begin{thm}{(Asymptotics and inference)} \label{thm:beta}
Supposing the conditions in Theorem \ref{thm1} hold with 
  $tp_t^2 \rightarrow \infty$ as $t\rightarrow \infty$, we have $\sqrt{t}\{\widehat{\boldsymbol{\beta}}_t(a)-\boldsymbol{\beta}(a)\}  \stackrel{D}{\longrightarrow}  \mathcal{N}_d\{ \boldsymbol{0}_d, \sigma_{\boldsymbol{\beta}(a)}^2\}$ and  $\sqrt{t}\{\widehat{\mu}_t(\boldsymbol{x},a) -\mu(\boldsymbol{x},a) \} \stackrel{D}{\longrightarrow}  \mathcal{N}\{ 0, \boldsymbol{x}^{\top} \sigma_{\boldsymbol{\beta}(a)}^2 \boldsymbol{x}\}$ for $\forall \boldsymbol{x}\in \mathcal{X}$ and  $\forall a \in \mathcal{A}$, with the  variance given by
 \begin{equation*}
 \begin{aligned}
\sigma_{\boldsymbol{\beta}(a)}^2  = \sigma_a^2  &\left[  \int   \kappa_{\infty} (\boldsymbol{x})\mathbb{I}\{\boldsymbol{x}^{\top} \boldsymbol{\beta}(a) < \boldsymbol{x}^{\top}\boldsymbol{\beta}(1-a) \}   \boldsymbol{x} \boldsymbol{x}^{\top}   d P_{\mathcal{X}} \right. \\
& \left.  + \int \{1-  \kappa_{\infty} (\boldsymbol{x})\} \mathbb{I}\left\{\boldsymbol{x}^{\top} \boldsymbol{\beta}(a) \geq \boldsymbol{x}^{\top}\boldsymbol{\beta}(1-a)\right\}   \boldsymbol{x} \boldsymbol{x}^{\top}   d P_{\mathcal{X}} \right]^{-1}.
\end{aligned}
 \end{equation*}
\end{thm}
The results in Theorem \ref{thm:beta} establish the asymptotic normality of the online ridge estimator and the online conditional mean estimator over time, with an explicit asymptotic variance derived. Here,  $\kappa_{\infty} (\boldsymbol{x})$ may differ under different bandit algorithms while the asymptotic normality holds as long as the adopted bandit algorithm explores the non-optimal actions sufficiently with a non-increasing rate, following the conclusion in Theorem \ref{thm1}, i.e., a clipping rate $p_t$ of an order larger than $\mathcal{O}(t^{-1/2})$ as described in Algorithm \ref{algo_cb}.  The proof of Theorem \ref{thm:beta} generalizes Theorem 3.1 in \cite{chen2020statistical} by additionally considering the online ridge estimator and general bandit algorithms.

\subsection{Asymptotic Normality and Robustness for DREAM}\label{sec:4.3}
   
In this section, we further derive the asymptotic normality of $\sqrt{T}(\widehat{V}_T -V^*)$. 
The following additional assumption is required to establish the double robustness of DREAM.
 
\begin{assumption}(Rate Double Robustness) \label{ADouble}
Define $||z_t||_{2,T}= \sqrt{{1\over T}\sum_{t=1}^T z_t^2 } $ as the $L_2$ norm. Let $||\mu(\boldsymbol{x} ,a)-\widehat{\mu}_t(\boldsymbol{x} ,a)||_{2,T}=\mathcal{O}_p(c_{\mu,T})\text{ for } a \in \mathcal{A}$,  and  $||\kappa_t(\boldsymbol{x}  )-\widehat{\kappa}_t(\boldsymbol{x}  )||_{2,T}=\mathcal{O}_p(c_{\kappa,T})$. Assume the product of two rates satisfies $c_{\mu,T} c_{\kappa,T} = o(T^{-1/ 2})$.
\end{assumption}
Assumption \ref{ADouble} requires the estimated conditional mean function and the estimated probability of exploration to converge at certain rates in online learning. This assumption is frequently studied in the causal inference literature \citep[see e.g., ][]{farrell2015robust, luedtke2016statistical,smucler2019unifying, hou2021treatment,kennedy2022semiparametric} to derive the asymptotic distribution of the estimated average treatment effect with either parametric estimators or non-parametric estimators \citep[see e.g., ][]{wager2018estimation,farrell2021deep}. We remark that under the conditions required in Theorem \ref{thm:beta}, we have 
$||\mu(\boldsymbol{x} ,a)-\widehat{\mu}_t(\boldsymbol{x} ,a)||_{2,T}=\mathcal{O}_p(T^{-1/ 2})$ for all $a\in \mathcal{A}$, thus Assumption \ref{ADouble} holds 
as along as 
$||\kappa_t(\boldsymbol{x}  )-\widehat{\kappa}_t(\boldsymbol{x}  )||_{2,T}=o_p(1)$.   We then have the asymptotic normality of $\sqrt{T}(\widehat{V}_T -V^*)$ as stated in the following theorem.
 
\begin{thm}{(Asymptotic normality for DREAM)} \label{thm:asym}
Suppose conditions in Theorem \ref{thm:beta} hold. Assuming
Assumption \ref{ADouble}, with  $\left\|\widehat{\kappa}_{t}(\cdot)-\kappa_{t}\right\|_{\infty}=o_{p}(1)$ , we have 
$\sqrt{T}(\widehat{V}_T -V^*) \stackrel{D}{\longrightarrow}  \mathcal{N}\left(0,  \sigma_{DR}^2 \right),$ 
with 
\begin{equation*} 
\widehat{\sigma}_{T}^2 \overset{p}{\longrightarrow} \sigma_{DR}^2 = \int_{\boldsymbol{x}} \frac{ \pi^*(\boldsymbol{x} ) \sigma_{1}^2 +\{1-\pi^*(\boldsymbol{x} )\}\sigma_{0}^2 }{1-\kappa_\infty(\boldsymbol{x})} d {P_{\mathcal{X}}} +\Var \left[  {\mu} \{\boldsymbol{x} ,\pi^*(\boldsymbol{x} ) \} \right] <\infty.
\end{equation*}
\end{thm}
The above theorem shows that the value estimator under DREAM is doubly robust to the true value, given the product of the nuisance error rates of the probability of exploitation and the conditional mean outcome as $o_p(T^{-1/ 2})$. We use the Martingale Central Limit Theorem to overcome the non-i.i.d. sample problem for asymptotic normality. The asymptotic variance of the proposed estimator is caused by two sources. One is the variance due to the context information, and the other is a weighted average of the variance under the optimal arm and the variance under the non-optimal arms. The weight is determined by the probability of exploration when $t\rightarrow \infty$ under the adopted bandit algorithm. To the best of our knowledge, this is the first work that studies the asymptotic distribution of the mean outcome of the estimated optimal policy under general online bandit algorithms, which takes the exploration-and-exploitation trade-off into a thorough account. Our method thus fills a crucial gap in policy evaluation of online learning. By Theorem \ref{thm:asym}, a two-sided $1-\alpha$ confidence interval (CI) for $V^*$ is $ [ \widehat{V}_T-z_{\alpha/2}\widehat{\sigma}_{T}/\sqrt{T},\quad  \widehat{V}_T+z_{\alpha/2}\widehat{\sigma}_{T}/\sqrt{T} ]$,  where $z_{\alpha/2}$ denotes the upper $\alpha/2-$th quantile of a standard normal distribution.

 \section{Simulation Studies}\label{sec:simu} 

We investigate the finite sample performance and demonstrate the coverage probabilities of the policy value of DREAM in this section. 
The computing infrastructure used is a virtual machine in the AWS Platform with 72 processor cores and
144GB memory.

Consider the 2-dimensional context $\boldsymbol{x} =[x_1,x_2]^{\top}$, with  $x_1,x_2 {\sim}_{i.i.d.} \text{Uniform}(0,2\pi)$. Suppose the outcome of interest given $\boldsymbol{x}$ and $a$ is generated from $ \mathcal{N}\{\mu(\boldsymbol{x},a), \sigma_a^2\}$, where the conditional mean function takes a non-linear form as $\mu(\boldsymbol{x},a) = 2-a+(5a-1)\cos(x_1)+(1.5-3a)\cos(x_2)$, with equal variances as $\sigma_1=\sigma_0= 0.1$ and $\boldsymbol{\beta}(a) = [2-a, 5a-1, 1.5-3a]^{\top}$. Then the optimal policy is given by $\mathbb{I}\{-1 + 5\cos(x_1)-3\cos(x_2)>0\}$, and the optimal value is 3.27 calculated by integration.

We employ the three bandit algorithms described in Section \ref{sec:bandits} to generate the online data with  $\omega =1$ \citep{li2010contextual}, where $c_t = 1$ for UCB, $ \boldsymbol{\beta}(a)\sim  \mathcal{N}_d(\boldsymbol{0}_d,\boldsymbol{I}_d)$ with $\rho = 2$ for TS, and  $\epsilon_t=0.1t^{-2/5}$ for EG. Set the total decision time as $T = 2000$ with a burning period $T_0=50$. 

\begin{figure}[!t] 
\centering
\includegraphics[width = \textwidth]{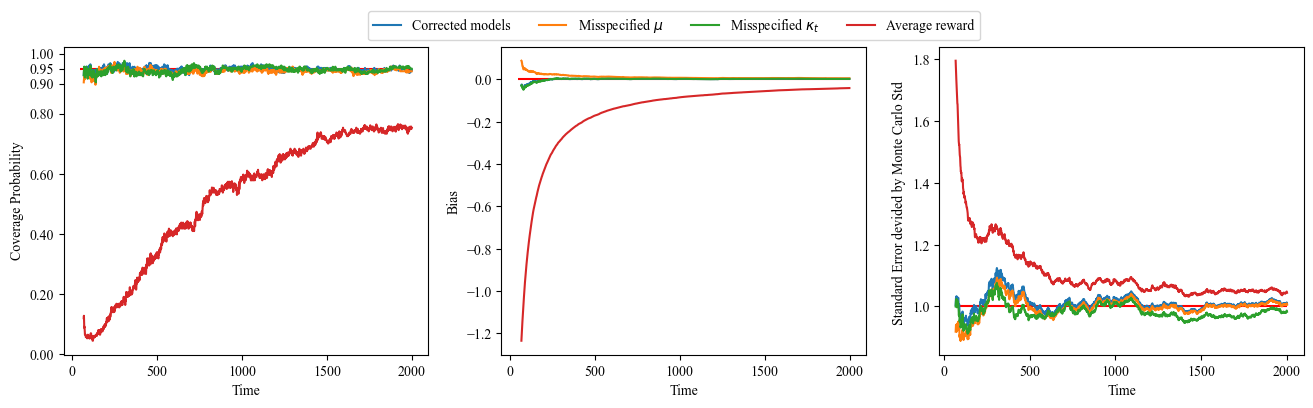}
  
     \caption{Results by DREAM with UCB under different model specifications in comparison to the averaged reward. Left panel: the coverage probabilities of the 95\% two-sided Wald-type confidence interval, with the red line representing the nominal level at 95\%. Middle panel: the bias between the estimated value and the true value. Right panel: the ratio between the standard error and the Monte Carlo standard deviation, with the red line representing the nominal level at 1. }
    \label{fig4ucb} 
\end{figure}

\begin{figure}[!t] 
\centering
\includegraphics[width = \textwidth]{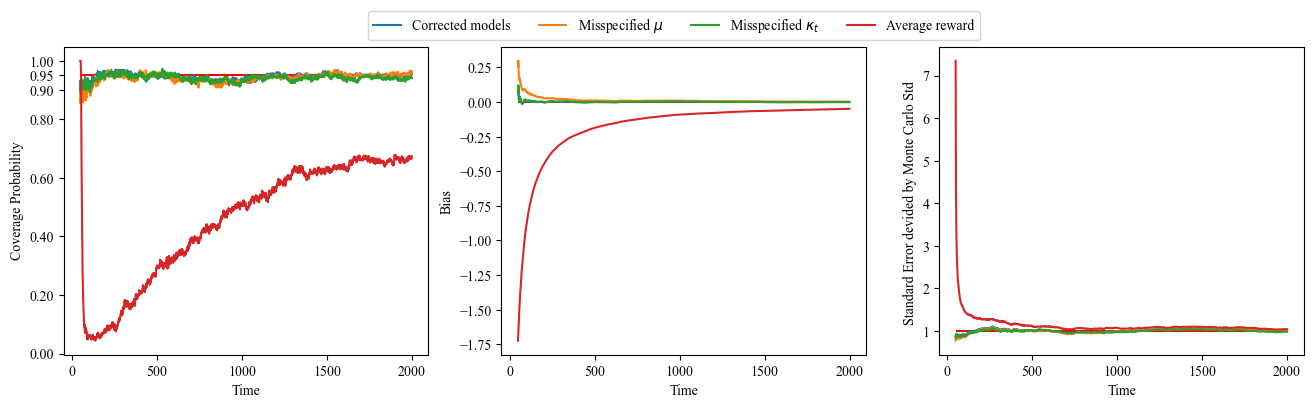}
  
     \caption{Results by DREAM with TS under different model specifications in comparison to the averaged reward. Left panel: the coverage probabilities of the 95\% two-sided Wald-type confidence interval, with the red line representing the nominal level at 95\%. Middle panel: the bias between the estimated value and the true value. Right panel: the ratio between the standard error and the Monte Carlo standard deviation, with the red line representing the nominal level at 1. }
    \label{fig4ts} 
\end{figure}

\begin{figure}[!t] 
\centering
\includegraphics[width = \textwidth]{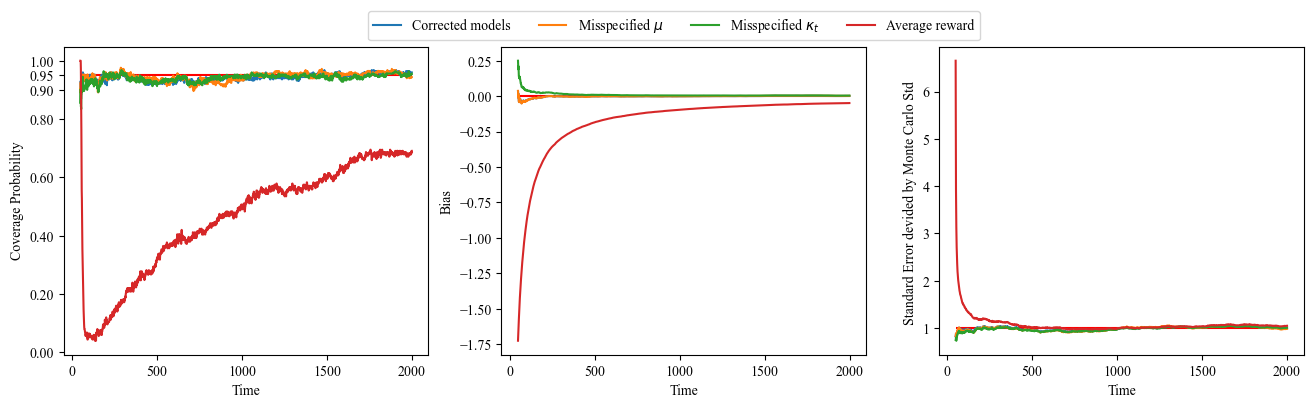}
  
     \caption{Results by DREAM with EG under different model specifications in comparison to the averaged reward. Left panel: the coverage probabilities of the 95\% two-sided Wald-type confidence interval, with the red line representing the nominal level at 95\%. Middle panel: the bias between the estimated value and the true value. Right panel: the ratio between the standard error and the Monte Carlo standard deviation, with the red line representing the nominal level at 1. }
    \label{fig4eg} 
\end{figure}

We evaluate the doubly robustness property of the proposed value estimator under DREAM in comparison to the simple average of the total reward for contextual bandits. To be specific, we consider the following four methods: 1. the conditional mean function $\mu$ is correctly specified, and the probability of exploration  $\kappa_t $ is estimated by a nonparametric regression  in DREAM; 2. the probability of exploration  $\kappa_t $ is estimated by a nonparametric regression while the  model of $\mu$ is misspecified with linear regression; 3. the conditional mean function $\mu$ is correctly specified while the model of  $\kappa_t  $ is misspecified by a constant 0.5; 4. using the averaged reward as the value estimator. The clipping rate of DREAM is set to be 0.01, and our method is not overly sensitive to the choice of $p_t$ as shown in additional sensitivity analyses provided in Appendix A. 
 The above four value estimators are evaluated by the coverage probabilities of the 95\% two-sided Wald-type CI on covering the optimal value, the bias, and the ratio between the standard error and the Monte Carlo standard deviation, as shown in Figure \ref{fig4ucb} for UCB, Figure \ref{fig4ts} for TS, and Figure \ref{fig4eg} for EG, aggregated over 1000 runs.

Based on Figures \ref{fig4ucb}, \ref{fig4ts}, and \ref{fig4eg}, the performance of the proposed DREAM method is reasonably much better than the simple average estimator of the total outcome. Specifically, under different bandit algorithms, when the time $t$ increases, 
 the coverage probabilities of the proposed DREAM estimator are close to the nominal level of 95\%, with the biases approaching 0  and the ratios between the standard error and the Monte Carlo standard deviation approaching 1. In addition, our DREAM method achieves reasonably good performance when either regression model for the conditional mean function or the probability of exploration is misspecified. These findings not only validate the theoretical results in Theorem \ref{thm:asym} but also demonstrate the doubly robustness of DREAM in handling policy evaluation in online learning. In contrast, the simple average of reward can hardly maintain coverage probabilities over 80\% with much larger biases in all cases.

   \section{Real Data Application} \label{sec:realdata}
  
   In this section, we evaluate the performance of the proposed DREAM method in real datasets from  \href{https://www.openml.org/}{the OpenML database}, which is a curated, comprehensive benchmark suite for machine-learning tasks. 
  Following the contextual bandit setting considered in this paper, we select two datasets in the public OpenML Curated Classification benchmarking suite 2018 (OpenML-CC18; BSD 3-Clause license) \citep{bischl2017openml}, i.e., the SEA50 and SEA50000, to formulate the real application.  Each dataset is a collection of pairs of 3-dimensional features $\boldsymbol{x}$ and their corresponding labels $Y \in \{0,1\}$, with a total number of observations as  $n=$1,000,000.  To simulate an online environment for data generation, we turn two-class classification tasks into two-armed contextual bandit problems \citep[see e.g., ][]{dudik2011doubly,wang2017optimal,su2019doubly}, such that we can reproduce the online data to evaluate the performance of the proposed method. 
   Specifically, at each time step $t$, we draw the pair $\{\boldsymbol{x}_t, Y_t\}$ uniformly at random without replacement from the dataset with $n=$1,000,000. Given the revealed context $\boldsymbol{x}_t$, the bandit algorithm selects an action $a_t \in\{0,1\}$. The reward is generated by a normal distribution $ \mathcal{N} \{\mathbb{I}(a_t =Y_t), 0.5^2\}$. Here, the mean of reward is 1 if the selected action matches the underlying true label, and 0 otherwise. Therefore, the optimal value is 1 while the optimal policy is unknown due to the complex relationship between the collected features and the label. Our goal is to infer the value under the optimal policy in the online settings produced by datasets SEA50 and SEA50000.
 
\begin{figure}[!t] 
\centering
\includegraphics[width=\textwidth]{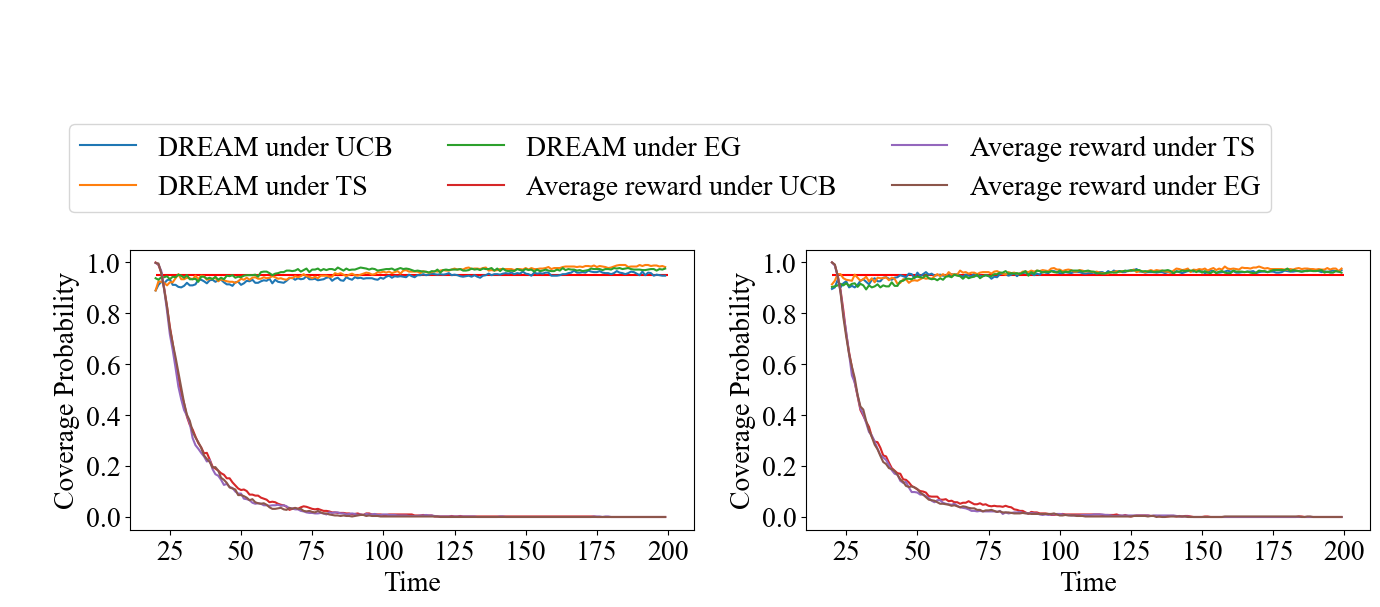}
 
     \caption{The coverage probabilities of the 95\% two-sided Wald-type confidence interval by the proposed value estimator under DREAM in comparison to the averaged reward, under different contextual bandit algorithms. Left panel: the results for the SEA50 dataset. Right panel: the results for the SEA50000 dataset. The red line represents the nominal level at 95\%.}
    \label{fig_real_CPv} 
      
\end{figure}

   Using the similar procedure as described in Section \ref{sec:simu}, we apply  DREAM in comparison to the simple average estimator, by employing three bandit algorithms with the following specifications: (i) for UCB, let $c_t=2$;  (ii) for TS, let priors $ \boldsymbol{\beta}(a)\sim \mathcal{N}_d(\boldsymbol{0}_d,\boldsymbol{I}_d)$ and parameter $\rho=0.5$; and (iii) for EG, let $\epsilon_t \equiv  t^{-1/3}$. Set the total time for the online learning as $T = 200$ with a burning period as $T_0=20$. The results are evaluated by the coverage probabilities of the 95\% two-sided Wald-type CI in Figure \ref{fig_real_CPv} for two real datasets, respectively, averaged over 500 replications. It can be observed from Figure \ref{fig_real_CPv} that our proposed DREAM method performs much better than the simple average estimator, in all cases. To be specific, the coverage probabilities of the value estimator under DREAM are close to the nominal level of 95\%, while the CI constructed by the averaged reward can hardly cover the true value with its coverage probabilities decaying to 0, under different bandit algorithms and two simulated online environments. These findings are consistent with what we have observed in simulations and consolidate the practical usefulness of the proposed DREAM.

\section{Discussion} \label{sec:discussion}

 In this paper, we propose doubly robust interval estimation (DREAM) to infer the mean outcome of the optimal policy using the online data generated from a bandit algorithm. We explicitly characterize the probability of exploring the non-optimal actions under different bandit algorithms and show the consistency and asymptotic normality of the proposed value estimator. In this section, we discuss the  performance of DREAM in terms of the regret bound and extend it to the evaluation of a known policy in online learning.

\subsection{Regret Bound under DREAM}\label{sec:reg}
In this section, we discuss the regret bound  of the proposed DREAM method. Specifically, we study the regret defined as the difference between the expected cumulative rewards under the oracle optimal policy and the bandit policy, which is
\begin{equation} \label{eq:regret}
R_T = \sum_{t=1}^{T} \Mean \left\{\mu( \boldsymbol{x}_t,\pi^*(\boldsymbol{x}_t) ) - \mu( \boldsymbol{x}_t,a_t)    \right\}.
\end{equation}
By noticing that $\mu( \boldsymbol{x}_t,\pi^*(\boldsymbol{x}_t) ) - \mu( \boldsymbol{x}_t,a_t)  = \Delta_{\boldsymbol{x}_t}$ if $a_t \neq \pi^*(\boldsymbol{x}_t) $ and 0 otherwise,  we have 
\begin{equation*}
R_T = \sum_{t=1}^{T} \Mean \left[ \Delta_{\boldsymbol{x}_t} \mathbb{I}\left\{ a_t \neq \pi^*(\boldsymbol{x}_t)\right\}\right].
\end{equation*}
We note that the indicator function is equivalent to $|a_t - \pi^*(\boldsymbol{x}_t)|$ and is  bounded by $|a_t - \widehat{\pi}(\boldsymbol{x}_t)| + |\widehat{\pi}(\boldsymbol{x}_t) - \pi^*(\boldsymbol{x}_t)|$. Thus, we can divide the regret defined in Equation \eqref{eq:regret} into two parts as $R_T \leq R_T^{(1)} +R_T^{(2)}$  
where 
\begin{equation*}
R_T^{(1)}  =  \sum_{t=1}^{T} \Mean \left[ \Delta_{\boldsymbol{x}_t} |a_t - \widehat{\pi}(\boldsymbol{x}_t)|\right],
\end{equation*}
is the regret from the exploration and $R_T^{(2)}  =  \sum_{t=1}^{T} \Mean \left[ \Delta_{\boldsymbol{x}_t} |\widehat{\pi}(\boldsymbol{x}_t) - \pi^*(\boldsymbol{x}_t)|\right]$
is the regret from the exploitation. It is well known that the regret from the exploitation $R_T^{(2)} $ is sublinear \citep{chu2011contextual, agrawal2013thompson}  and the regret for EG has been well studied by \cite{chen2020statistical}. Therefore, we focus on the analysis of the regret $R_T
^{(1)}$ from the exploration for UCB and TS here.

Since $\Delta_{\boldsymbol{x}_t}$ is bounded and  the upper bound of $\Mean |a_t - \widehat{\pi}(\boldsymbol{x}_t)| =  \Mean [ \mathbb{I}\{a_t\not = \widehat{\pi}_t(\boldsymbol{x}_t )\}]= \kappa_t(\boldsymbol{x}_t)$ has an asymptotically equivalent convergence rate as  $\mathcal{O}\{\exp(-t p_t^2)\}$ up to some constant by Theorem \ref{thm1}, there exists some constant $C$ such that the regret from the exploration  is bounded by 
\begin{equation*}
R_T^{(1)} \leq C 	 \sum_{t=1}^{T}\mathcal{O}\{\exp(-t p_t^2)\}. 
\end{equation*}
If we choose $p_t = \sqrt{ {\alpha  \log t}/{t}}$ for some $\alpha \in (0,1)$, the regret $R_T^{(1)}$ is bounded by $\sum_{t=1}^{T}\mathcal{O}\{t^{-\alpha}\} = \mathcal{O}\{T^{1-\alpha} \} $, where the equation is calculated using  Lemma 6 in \cite{luedtke2016statistical}. Thus, 
we still have sublinear regret under DREAM.

\subsection{Evaluation of Known Policies in Online Learning}

We could extend our method to evaluate a new known policy $\pi^{E}$ that is different from the bandit policy, in the online environment.
 Typically, we focus on the statistical inference of policy evaluation in online learning under the contextual bandit framework. Recall the setting and notations in Section \ref{sec:2}, given a target policy $\pi^E$, we propose its doubly robust value estimator as
\begin{equation*}
\widehat{V}_T({\pi}^E)=\frac{1}{T}\sum_{t=1}^T \frac{\mathbb{I}\{a_t={\pi}^E(\boldsymbol{x}_t )\}}{\widehat{p}_{t-1}\{a_t|\boldsymbol{x}_t\}} \Big[ r_t -\widehat{\mu}_{t-1}\{\boldsymbol{x}_t ,{\pi}^E(\boldsymbol{x}_t )\}\Big] + \widehat{\mu}_{t-1}\{\boldsymbol{x}_t ,{\pi}^E(\boldsymbol{x}_t ) \},
\end{equation*}
 where $T$ is the current time step or the termination time and $ \widehat{p}_{t-1}(a_t|\boldsymbol{x}_t)$ is the estimator for the propensity score of the chosen action $a_t$ denoted as $p_{t-1}(a_t|\boldsymbol{x}_t)$. The following theorem summarizes the asymptotic properties of $\widehat{V}_T({\pi}^E)$, built on Theorem \ref{thm:beta}.

\begin{coro}{(Asymptotic normality for evaluating a known policy)}
Suppose the conditions in Theorem \ref{thm:beta} hold. Furthermore, assuming the rate doubly robustness that $||\mu(\boldsymbol{x} ,a)-\widehat{\mu}_t(\boldsymbol{x} ,a)||_{2,T} ||p_t(a|\boldsymbol{x})-\widehat{p}_{t}(a|\boldsymbol{x})||_{2,T}=o_p(T^{-1/ 2}) \text{ for } a \in \mathcal{A},$
we have $\sqrt{T}\{\widehat{V}_T(\pi^E) -V(\pi^E)\}  \stackrel{D}{\longrightarrow}  \mathcal{N}\{ 0, \sigma(\pi^E)^2\}$, with
\begin{equation*}
\sigma(\pi^E)^2 = \int_{\boldsymbol{x}} \frac{ \pi^E(\boldsymbol{x} ) \sigma_{1}^2 +\{1-\pi^E(\boldsymbol{x} )\}\sigma_{0}^2 }{\prob\left\{\pi^E(\boldsymbol{x} )|\boldsymbol{x}\right\}} d {P_{\mathcal{X}}} +\Var \left[  {\mu} \{\boldsymbol{x} ,\pi^E(\boldsymbol{x} ) \} \right] <\infty,
\end{equation*}
where $\prob\left\{\pi^E(\boldsymbol{x} )|\boldsymbol{x}\right\} = \lim_{t\rightarrow \infty} p_{t-1}(\pi^E(\boldsymbol{x}_t )|\boldsymbol{x}_t)$.
\end{coro}
Here, we impose the same conditions on the bandit algorithms as in Theorem \ref{thm:asym} to guarantee sufficient exploration on different arms, and thus evaluating an arbitrary policy is valid. The usage of the new rate doubly robustness assumption and the margin condition (Assumption \ref{AMargin}) follows a similar logic as in Theorem \ref{thm:asym}. The estimator  of $\sigma(\pi^E)^2$  denoted as  $\widehat{\sigma}(\pi^E)^2$ can be obtained similarly to \eqref{eq:sigmahat}. Thus, a two-sided $1-\alpha$ CI for $V(\pi^E)$ under online optimization is $ [ \widehat{V}_T(\pi^E)-z_{\alpha/2}\widehat{\sigma}(\pi^E)/\sqrt{T},\quad  \widehat{V}_T(\pi^E)+z_{\alpha/2}\widehat{\sigma}(\pi^E)/\sqrt{T} ]$.

\bigskip

 There are some other extensions that we may also consider in future work. First, in this paper, we focus on settings with binary actions. Thus, a more general method with multiple actions or even a continuous action space is desirable. Second, we consider the contextual bandits in this paper and all the theoretical results are applicable to the multi-armed bandits. It would be practically interesting to extend our proposal to reinforcement learning problems. Third, instead of using the rate double robustness assumption in the current paper, it is of theoretical interest to impose the model double robustness version of DREAM in future research.

\newpage

\bibliography{mycite}
\bibliographystyle{agsm}

\newpage
\appendix
\counterwithin{figure}{section}
\counterwithin{table}{section}
\counterwithin{equation}{section}


\begin{center}
{\Large\bf Supplementary to `Doubly Robust Interval Estimation for Optimal Policy Evaluation in Online Learning'}
\end{center}

This supplementary article provides sensitivity analyses and all the technical proofs for the established theorems for policy evaluation in online learning under the contextual bandits. Note that the theoretical
results in Section 7 can be proven in a similar manner by arguments in Section \ref{sec:supproofs}. Thus we omit the details here.

\section{Sensitivity Test for the Choice of \texorpdfstring{$p_{t}$}{Lg}} \label{sec:sensitivity}

We conduct a sensitivity test for the choice of $p_{t}$ in this section. We run all the simulations with $p_t=0.01,0.05,0.1$ and find that Algorithm 1 is not sensitive to the choice of $p_{t}$.

\begin{figure}[!htp] 
\centering     
\subfigure[$p_t = 0.01$]{ \includegraphics[width=  \textwidth]{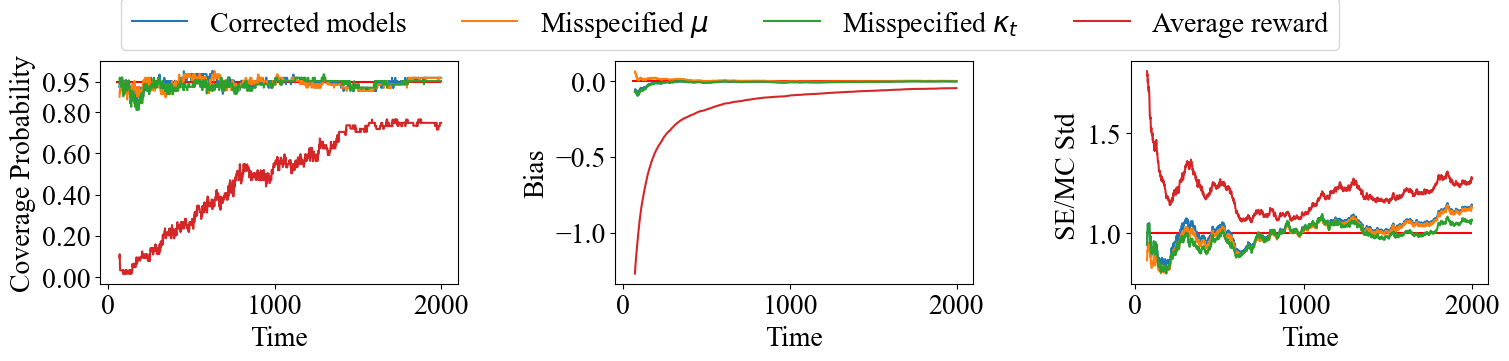}}
\caption{Results by DREAM under UCB with different model specifications  in comparison to the averaged reward. Left panel: the coverage probabilities of the 95\% two-sided Wald-type CI, with the red line representing the nominal level at 95\%. Middle panel: the bias between the estimated value and the true value. Right panel: the ratio between the standard error and the Monte Carlo standard deviation, with the red line representing the nominal level at 1. }
\end{figure}

\begin{figure}[!htp] 
\centering     
\ContinuedFloat
\subfigure[$p_t = 0.05$]{ \includegraphics[width= \textwidth]{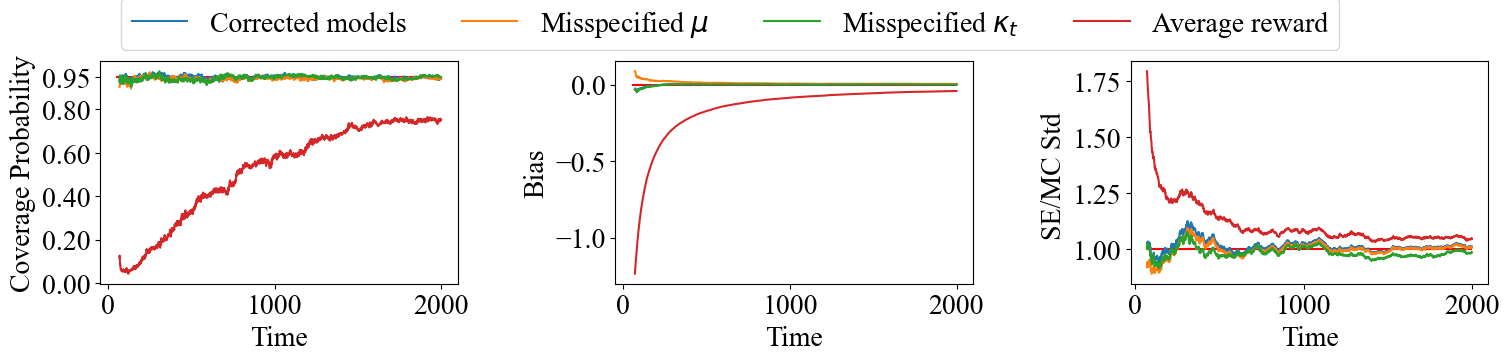}}
\subfigure[$p_t = 0.1$]{ \includegraphics[width= \textwidth]{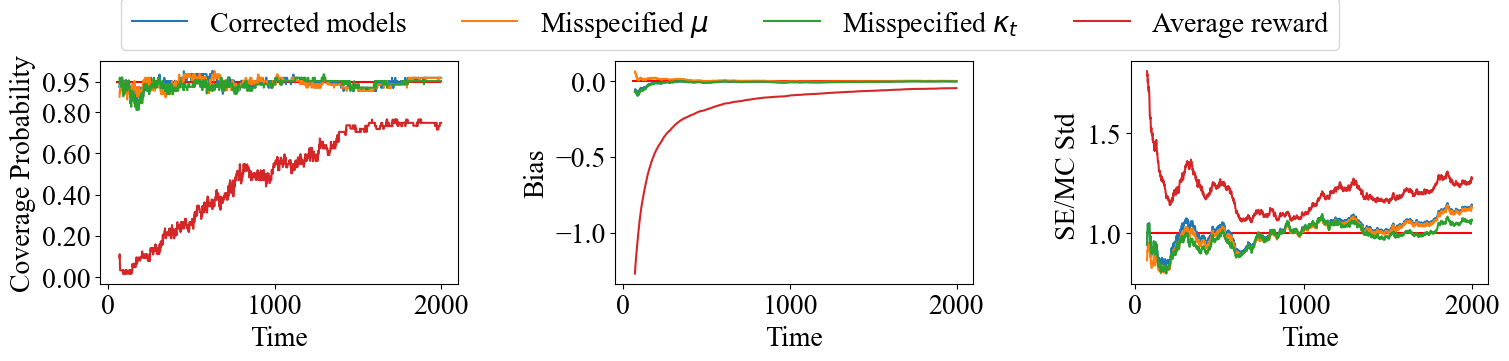}}
\caption{Results by DREAM under UCB with different model specifications  in comparison to the averaged reward. Left panel: the coverage probabilities of the 95\% two-sided Wald-type CI, with the red line representing the nominal level at 95\%. Middle panel: the bias between the estimated value and the true value. Right panel: the ratio between the standard error and the Monte Carlo standard deviation, with the red line representing the nominal level at 1. }
\end{figure}

\newpage
\begin{figure}[!h] 
\centering     
\subfigure[$p_t = 0.01$]{ \includegraphics[width= \textwidth]{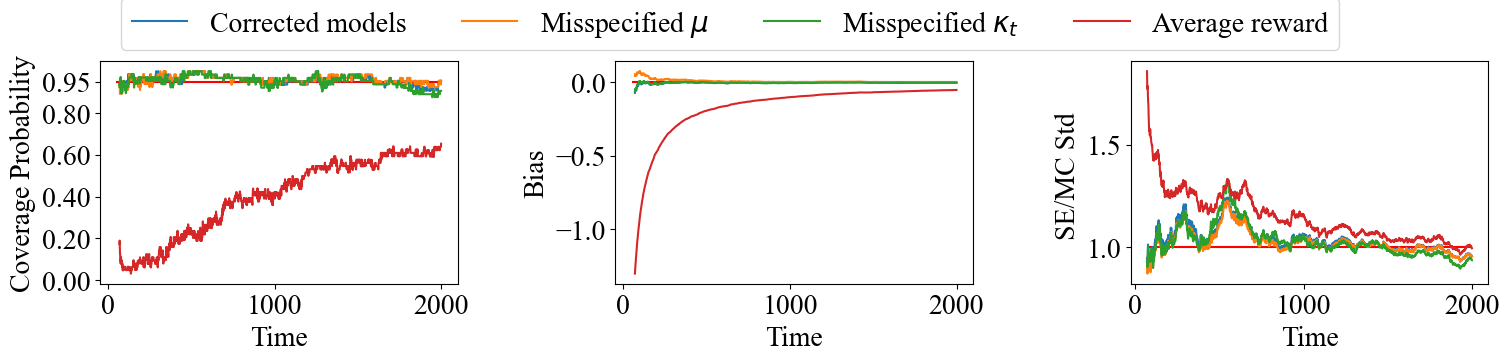}}
\subfigure[$p_t = 0.05$]{ \includegraphics[width= \textwidth]{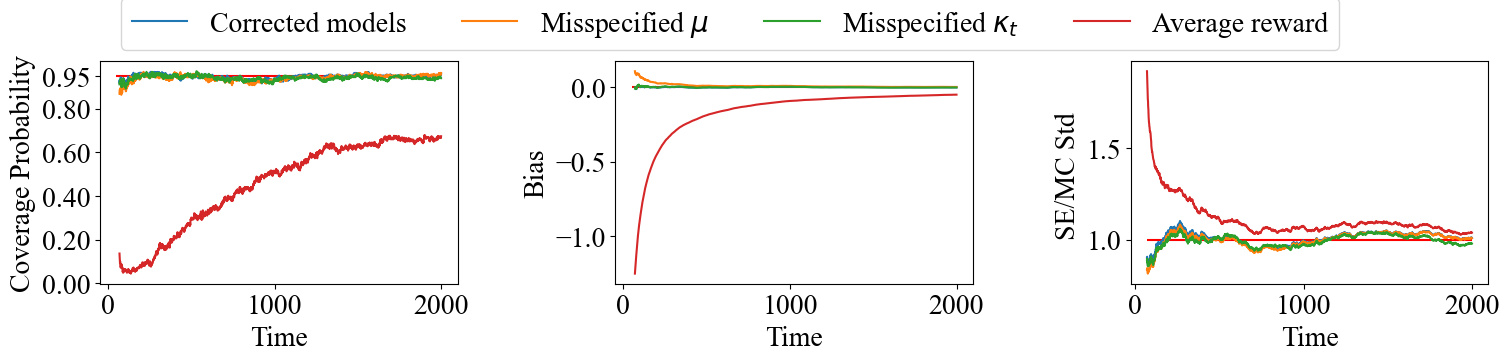}}
\subfigure[$p_t = 0.1$]{ \includegraphics[width= \textwidth]{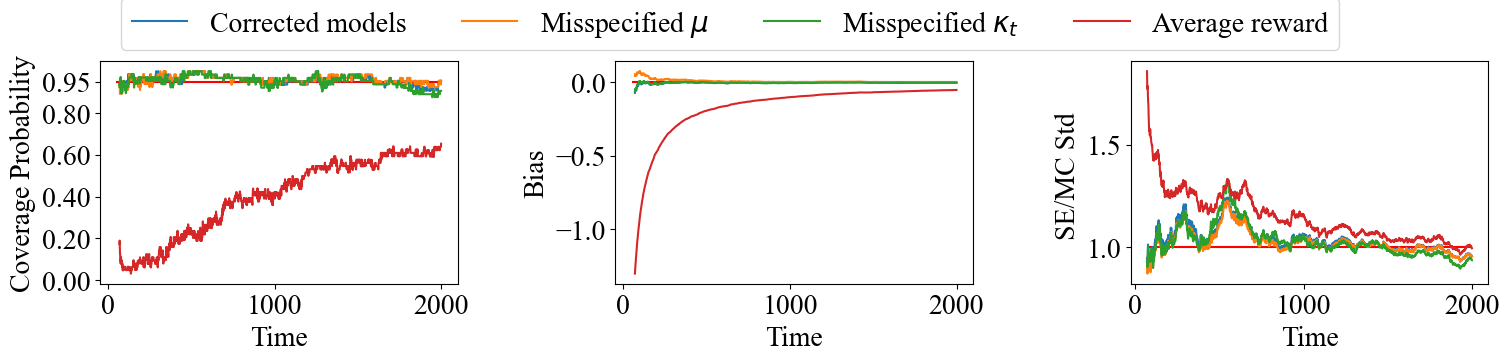}}
\caption{Results by DREAM under TS with different model specifications  in comparison to the averaged reward. Left panel: the coverage probabilities of the 95\% two-sided Wald-type CI, with the red line representing the nominal level at 95\%. Middle panel: the bias between the estimated value and the true value. Right panel: the ratio between the standard error and the Monte Carlo standard deviation, with the red line representing the nominal level at 1. }
\end{figure}

\newpage
\begin{figure}[!h] 
\centering     
\subfigure[$p_t = 0.01$]{ \includegraphics[width= \textwidth]{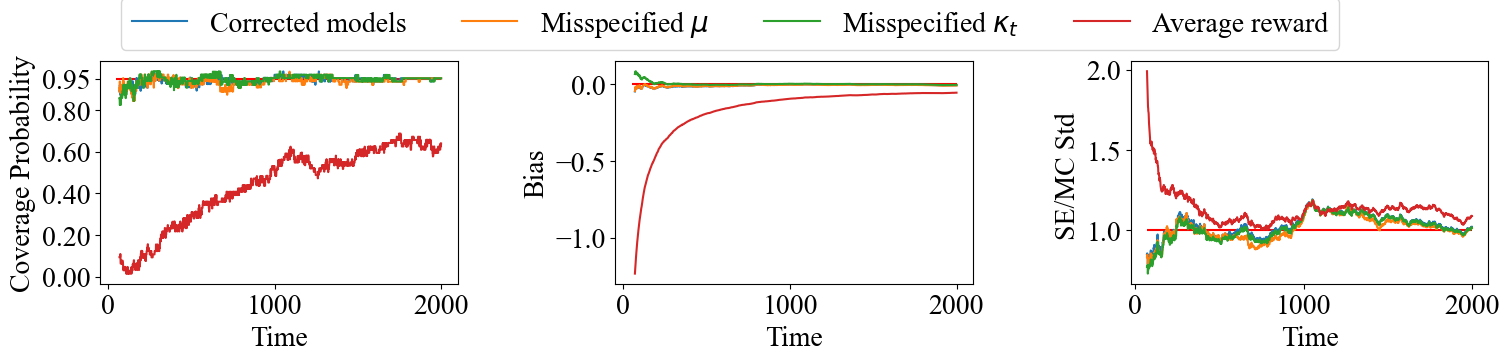}}
\subfigure[$p_t = 0.05$]{ \includegraphics[width= \textwidth]{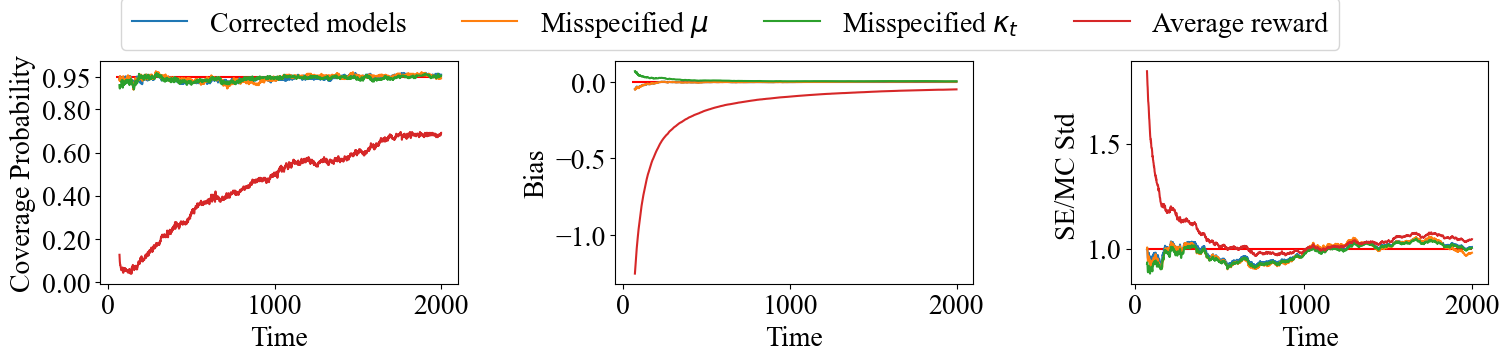}}
\subfigure[$p_t = 0.1$]{ \includegraphics[width= \textwidth]{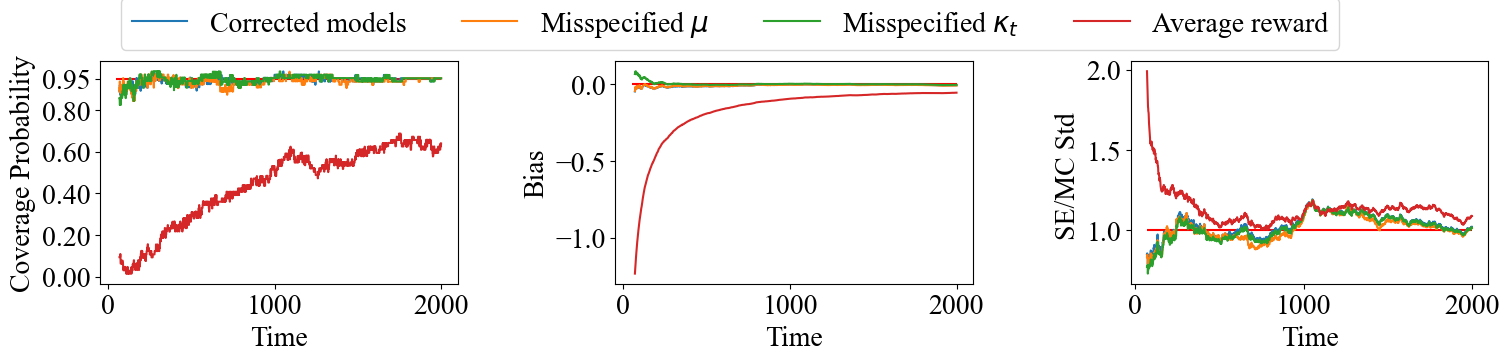}}
\caption{Results by DREAM under EG with different model specifications  in comparison to the averaged reward. Left panel: the coverage probabilities of the 95\% two-sided Wald-type CI, with the red line representing the nominal level at 95\%. Middle panel: the bias between the estimated value and the true value. Right panel: the ratio between the standard error and the Monte Carlo standard deviation, with the red line representing the nominal level at 1. }
\end{figure}

\section{Technical Proofs for Main Results}\label{sec:supproofs}
 This section provides all the technical proofs for the established theorems for policy evaluation in online learning under the contextual bandits.

\subsection{Proof of Lemma 4.1}
The proof of Lemma 4.1 consists of three main steps. To be specific, we first reconstruct the target difference $\widehat{\boldsymbol{\beta}}_t(a)- \boldsymbol{\beta}(a)$ and decompose it into two parts. Then, we establish the bound for each part and derive its  lower bound $\prob ( \|\widehat{\boldsymbol{\beta}}_t(a)- \boldsymbol{\beta}(a)\|_{1} \leq h )$. 

\smallskip \noindent \textbf{Step 1:}  
Recall Equation (1) in the main paper  with  $\boldsymbol{D}_{t-1}(a)$ being a $\boldsymbol{N}_{t-1}(a) \times d$ design matrix at time $t-1$ with $\boldsymbol{N}_{t-1}(a)$ as the number of pulls for action $a$, we have
\begin{eqnarray*}
\widehat{\boldsymbol{\beta}}_t(a) = \left \{ \frac{1}{t} \sum_{i=1}^{t} \mathbb{I}(a_i=a)\boldsymbol{x}_i \boldsymbol{x}_i^\top +\frac{1}{t} \omega \boldsymbol{I}_d \right\}^{-1}\left \{\frac{1}{t}\sum_{i=1}^t \mathbb{I}(a_i=a)\boldsymbol{x}_i r_i\right\}.
\end{eqnarray*}
We are interested in the quantity
\begin{eqnarray} \label{betahat_beta}
\widehat{\boldsymbol{\beta}}_t(a)- \boldsymbol{\beta}(a)  = \left \{ \frac{1}{t} \sum_{i=1}^{t} \mathbb{I}(a_i=a)\boldsymbol{x}_i \boldsymbol{x}_i^\top +\frac{1}{t} \omega \boldsymbol{I}_d \right\}^{-1}\left \{\frac{1}{t}\sum_{i=1}^t \mathbb{I}(a_i=a)\boldsymbol{x}_i r_i\right\}- \boldsymbol{\beta}(a). 
\end{eqnarray}
Note that  $ \boldsymbol{\beta}(a)$ can be written as
\begin{equation*}
    \left \{ \frac{1}{t} \sum_{i=1}^{t} \mathbb{I}(a_i=a)\boldsymbol{x}_i \boldsymbol{x}_i^\top +\frac{1}{t} \omega \boldsymbol{I}_d\right\}^{-1}\left \{\frac{1}{t}\sum_{i=1}^t \mathbb{I}(a_i=a)\boldsymbol{x}_i \boldsymbol{x}_i^\top+\frac{1}{t} \omega \boldsymbol{I}_d \right\}  \boldsymbol{\beta}(a) , 
    \end{equation*}
and since $r_i = \boldsymbol{x}_i^\top  \boldsymbol{\beta}(a) +e_i$, we can write \eqref{betahat_beta} as 
\begin{eqnarray}\label{decompose_beta_pf}
\widehat{\boldsymbol{\beta}}_t(a)- \boldsymbol{\beta}(a) =&&\underbrace{ \left \{ \frac{1}{t} \sum_{i=1}^{t} \mathbb{I}(a_i=a)\boldsymbol{x}_i \boldsymbol{x}_i^\top +\frac{1}{t} \omega \boldsymbol{I}_d \right\}^{-1}\left \{\frac{1}{t}\sum_{i=1}^t \mathbb{I}(a_i=a)\boldsymbol{x}_i  e_i\right\}}_{\eta_3}\\\nonumber
&&-\underbrace{ \left \{ \frac{1}{t} \sum_{i=1}^{t} \mathbb{I}(a_i=a)\boldsymbol{x}_i \boldsymbol{x}_i^\top +\frac{1}{t} \omega \boldsymbol{I}_d \right\}^{-1} \frac{ \omega }{t} \boldsymbol{\beta}(a) }_{\eta_4}.
\end{eqnarray}
Our goal is to find a lower bound of $\prob ( \|\widehat{\boldsymbol{\beta}}_t(a)- \boldsymbol{\beta}(a) \|_{1} \leq h  )$ for any $h>0$. Notice that by the triangle inequality we have $ \|\widehat{\boldsymbol{\beta}}_t(a)- \boldsymbol{\beta}(a)\|_{1} \leq \left\|\eta_3\right\|_{1}+ \left\|\eta_4\right\|_{1}$, thus we can find the lower bound using the inequality as
\begin{eqnarray} \label{eta1}
\prob \left( \left\|\widehat{\boldsymbol{\beta}}_t(a)- \boldsymbol{\beta}(a)\right\|_{1} \leq h \right) \geq \prob \left(\left\|\eta_3\right\|_{1}+ \left\|\eta_4\right\|_{1}\leq h \right).\end{eqnarray}

\smallskip \noindent \textbf{Step 2:}  We focus on bounding $\eta_4$ first. 
By the relationship between eigenvalues and the $L_2$ norm of symmetric matrix, we have $\|\boldsymbol{M}^{-1}\|_2 = \lambda_{\max}(\boldsymbol{M}^{-1}) = \{\lambda_{\min}(\boldsymbol{M}) \}^{-1}$
for any invertible matrix $\boldsymbol{M}$. Thus we can obtain that 
 
\begin{equation*}
\begin{aligned}
\left\|\left \{ \frac{1}{t} \sum_{i=1}^{t} \mathbb{I}(a_i=a)\boldsymbol{x}_i \boldsymbol{x}_i^\top +\frac{1}{t}\omega \boldsymbol{I}_d \right\}^{-1}\right\|_2 & =\left\{\lambda_{\min}\left( \frac{1}{t} \sum_{i=1}^{t} \mathbb{I}(a_i=a)\boldsymbol{x}_i \boldsymbol{x}_i^\top +\frac{1}{t}\omega \boldsymbol{I}_d \right) \right\}^{-1}\\
& = \frac{1}{\lambda_{\min}\left( \frac{1}{t} \sum_{i=1}^{t} \mathbb{I}(a_i=a)\boldsymbol{x}_i \boldsymbol{x}_i^\top  \right)+\frac{1}{t}\omega }\\
& \leq \frac{1}{p_t \lambda_{\min}\left(  \boldsymbol{\Sigma} \right)+\frac{1}{t}\omega  }\leq \frac{1}{p_t \lambda+\frac{1}{t}\omega },
\end{aligned}
\end{equation*}
 which leads to 
\begin{equation}\label{l2norm}
\|\eta_4\|_2  \leq  \left\|\left \{ \frac{1}{t} \sum_{i=1}^{t} \mathbb{I}(a_i=a)\boldsymbol{x}_i \boldsymbol{x}_i^\top +\frac{1}{t}\omega \boldsymbol{I}_d \right\}^{-1}\right\|_2\left\|\frac{ \omega }{t} \boldsymbol{\beta}(a)\right\|_2 \leq \frac{ \omega }{tp_t \lambda+\omega } \left\| \boldsymbol{\beta}(a)\right\|_2.
 \end{equation}
By Cauchy-Schwartz inequality, we further have bound of the $L_1$ norm of $\eta_4$ as
\begin{equation}\label{pf_thm2_part2}
\|\eta_4\|_1 \leq \sqrt{d}\|\eta_4\|_2 \leq \frac{ \omega \sqrt{d}}{tp_t \lambda+\omega }\left\| \boldsymbol{\beta}(a)\right\|_2\leq \frac{ \sqrt{d}}{1 +tp_t \lambda/ \omega}\left\| \boldsymbol{\beta}(a)\right\|_2\leq \sqrt{d} \left\| \boldsymbol{\beta}(a)\right\|_2.
\end{equation}

\smallskip \noindent \textbf{Step 3:} Lastly, using the results in \eqref{pf_thm2_part2},  we have
\begin{eqnarray} \label{eta2}
\prob \left(\left\|\eta_3\right\|_{1} + \left\|\eta_4\right\|_{1}\leq h  \right) \geq   \prob \left(\left\|\eta_3\right\|_{1} \leq h   - \sqrt{d} \left\| \boldsymbol{\beta}(a)\right\|_2\right) .\end{eqnarray}
By the definition of $\eta_3$ and Lemma 2 in \cite{chen2020statistical}, for any constant $c>0$, we have 
\begin{eqnarray*} 
\prob\left(\left\|\eta_3\right\|_{1}  > c\right) \leq 2 d \exp \left\{-\frac{t \left( \frac{p_t\lambda}{2}\right)^{2} c^{2}}{2 d^{2} \sigma^{2} L_{\boldsymbol{x}}^{2}}\right\}=2 d \exp \left\{-\frac{tp_t^{2}\lambda^{2} c^{2}}{8d^{2} \sigma^{2} L_{\boldsymbol{x}}^{2}}\right\}.
 \end{eqnarray*}
Therefore, from \eqref{eta1} and  \eqref{eta2}, taking $c=h   - \sqrt{d} \left\| \boldsymbol{\beta}(a)\right\|_2$, we have that under event $E_{t}$,
\begin{eqnarray} \label{proof_t1_e2}
\begin{aligned}
\prob \left( \left\|\widehat{\boldsymbol{\beta}}_t(a)- \boldsymbol{\beta}(a)\right\|_{1} >  h \right) & \leq 2 d \exp \left\{-\frac{tp_t^{2}\lambda^{2} \left(h   - \sqrt{d}\left\| \boldsymbol{\beta}(a)\right\|_2\right)^{2}}{8d^{2} \sigma^{2} L_{\boldsymbol{x}}^{2}}\right\}.\\
\end{aligned}
\end{eqnarray}
Based on the above results, it is immediate that the online ridge estimator $\widehat{\boldsymbol{\beta}}_t(a)$ is consistent to $\boldsymbol{\beta}(a)$ if $tp_t^{2}\rightarrow \infty$ as $t\rightarrow \infty$. 
The proof is hence completed.

\subsection{Proof of Corollary 1 }
Since $ \widehat{\mu}_t(\boldsymbol{x}_t,a) - \mu(\boldsymbol{x}_t,a)  = \boldsymbol{x}_t^{\top} \left( \widehat{\boldsymbol{\beta}}_t(a)-\boldsymbol{\beta}(a) \right)$, by Holder's inequality, we have
\begin{equation*}
\left| \widehat{\mu}_t(\boldsymbol{x}_t,a) - \mu(\boldsymbol{x}_t,a) \right| \leq   \left\| \boldsymbol{x}_t \right\|_{\infty} \left\|\widehat{\boldsymbol{\beta}}_t(a)-\boldsymbol{\beta}(a)\right\|_{1}  \leq L_{\boldsymbol{x}}  \left\|\widehat{\boldsymbol{\beta}}_t(a)-\boldsymbol{\beta}(a)\right\|_{1}, 
\end{equation*}
which follows
\begin{equation*}
\prob \left\{  \left| \widehat{\mu}_t(\boldsymbol{x}_t,a) - \mu(\boldsymbol{x}_t,a) \right| > \xi \right \} \leq \prob \left\{   L_{\boldsymbol{x}}  \left\|\widehat{\boldsymbol{\beta}}_t(a)-\boldsymbol{\beta}(a)\right\|_{1} > \xi \right \}  = \prob \left\{   \left\|\widehat{\boldsymbol{\beta}}_t(a)-\boldsymbol{\beta}(a)\right\|_{1} > \xi /  L_{\boldsymbol{x}} \right \}.
\end{equation*}
By Lemma 4.1, we further have
\begin{equation*}
\begin{aligned}
\prob \left\{  \left| \widehat{\mu}_t(\boldsymbol{x}_t,a) - \mu(\boldsymbol{x}_t,a) \right| > \xi \right \} 
& \leq \prob \left\{   \left\|\widehat{\boldsymbol{\beta}}_t(a)-\boldsymbol{\beta}(a)\right\|_{1} >  \frac{\xi}{L_{\boldsymbol{x}}}  \right \} \\
& \leq 2 d \exp \left\{-\frac{tp_t^{2}\lambda^{2} \left(\frac{\xi}{L_{\boldsymbol{x}}}    - \sqrt{d}\left\| \boldsymbol{\beta}(a)\right\|_2\right)^{2}}{8d^{2} \sigma^{2} L_{\boldsymbol{x}}^{2}}\right\} \\
& = 2 d \exp \left\{-\frac{tp_t^{2}\lambda^{2} \left( \xi     - \sqrt{d} L_{\boldsymbol{x}} \left\| \boldsymbol{\beta}(a)\right\|_2\right)^{2}}{8d^{2} \sigma^{2} L_{\boldsymbol{x}}^{4}}\right\}.
\end{aligned}
\end{equation*}
Note that by the Triangle Inequality, 
\begin{equation*}
\begin{aligned}
\left| \widehat{\mu}_t(\boldsymbol{x}_t,1)-\widehat{\mu}_t(\boldsymbol{x}_t,0) -  \Delta_{\boldsymbol{x}_t} \right| 
&  = \left| \left\{ \widehat{\mu}_t(\boldsymbol{x}_t,1) - \mu(\boldsymbol{x}_t,1)\right\} -  \left\{  \widehat{\mu}_t(\boldsymbol{x}_t,0) - \mu(\boldsymbol{x}_t,0)\right\} \right|   \\
& \leq \left| \widehat{\mu}_t(\boldsymbol{x}_t,1) - \mu(\boldsymbol{x}_t,1)\right| +  \left|  \widehat{\mu}_t(\boldsymbol{x}_t,0) - \mu(\boldsymbol{x}_t,0)\right|,   \\
\end{aligned}
\end{equation*}
thus for $\left| \widehat{\mu}_t(\boldsymbol{x}_t,1)-\widehat{\mu}_t(\boldsymbol{x}_t,0) -  \Delta_{\boldsymbol{x}_t} \right| $, we have
\begin{equation*}
\begin{aligned}
&\prob \left\{  \left| \widehat{\mu}_t(\boldsymbol{x}_t,1)-\widehat{\mu}_t(\boldsymbol{x}_t,0) -  \Delta_{\boldsymbol{x}_t} \right|   > \xi \right \} \\
 &\leq \prob \left\{ \left| \widehat{\mu}_t(\boldsymbol{x}_t,1) - \mu(\boldsymbol{x}_t,1)\right| +  \left|  \widehat{\mu}_t(\boldsymbol{x}_t,0) - \mu(\boldsymbol{x}_t,0)\right| > \xi \right \} \\
& \leq \prob \left\{ \left| \widehat{\mu}_t(\boldsymbol{x}_t,1) - \mu(\boldsymbol{x}_t,1)\right| > \xi/2 \right \} +   \prob \left\{  \left|  \widehat{\mu}_t(\boldsymbol{x}_t,0) - \mu(\boldsymbol{x}_t,0)\right| > \xi/2 \right \} \\
&\leq 2 d \exp \left\{-\frac{tp_t^{2}\lambda^{2} \left( \xi/2     - \sqrt{d} L_{\boldsymbol{x}} \left\| \boldsymbol{\beta}(1)\right\|_2\right)^{2}}{8d^{2} \sigma^{2} L_{\boldsymbol{x}}^{4}}\right\} + 2 d \exp \left\{-\frac{tp_t^{2}\lambda^{2} \left( \xi/2     - \sqrt{d} L_{\boldsymbol{x}} \left\| \boldsymbol{\beta}(0)\right\|_2\right)^{2}}{8d^{2} \sigma^{2} L_{\boldsymbol{x}}^{4}}\right\} \\
& \leq 4 d \exp \left\{-(t-1)p_{t-1}^{2} c_{\xi}\right\},
\end{aligned}
\end{equation*}
with 
\begin{equation*}
c_{\xi} = \frac{\lambda^{2} \left[ \min \left\{  \left( \xi/2     - \sqrt{d} L_{\boldsymbol{x}} \left\| \boldsymbol{\beta}(1)\right\|_2 \right)^{2}, \left( \xi/2     - \sqrt{d} L_{\boldsymbol{x}} \left\| \boldsymbol{\beta}(0)\right\|_2\right)^{2}   \right\} \right]}{8d^{2} \sigma^{2} L_{\boldsymbol{x}}^{4}},
\end{equation*}
consistent with time $t$.

\subsection{Proof of Theorem 1}
The proof of Theorem 1 consists of two main parts to show the probability of exploration under UCB and TS, respectively, by noting the probability of exploration under EG is given by its definition.

\subsubsection{Proof for UCB}
We first show the probability of exploration under UCB. This proof consists of three main steps stated as following:  
\begin{enumerate}
	\item We first rewrite the target probability by its definition and express it as 
	\begin{equation*}
	\prob\left\{\left|\widehat{\mu}_{t -1}(\boldsymbol{x}_t, 1)-\widehat{\mu}_{t -1}(\boldsymbol{x}_t, 0) \right| <  c_t \left| \widehat{\sigma}_{t-1}(\boldsymbol{x}_t, 0)-\widehat{\sigma}_{t-1}(\boldsymbol{x}_t, 1) \right|  \right\}.
	\end{equation*}
	\item Then, we establish the bound for the variance estimation such that  
    \begin{equation*}
    	c_t \left| \widehat{\sigma}_{t-1}(\boldsymbol{x}_t, 0)-\widehat{\sigma}_{t-1}(\boldsymbol{x}_t, 1) \right| \leq \frac{ 2 c_tL_{\boldsymbol{x}}}{\sqrt{(t-1)p_{t-1}\lambda }}.
    \end{equation*}
	\item Lastly, we bound $ \prob\left\{\left|\widehat{\mu}_{t -1}(\boldsymbol{x}_t, 1)-\widehat{\mu}_{t -1}(\boldsymbol{x}_t, 0) \right| < \frac{ 2 c_tL_{\boldsymbol{x}}}{\sqrt{(t-1)p_{t-1}\lambda }}   \right\} $ using the result in  Corollary 1. 
\end{enumerate}

\smallskip \noindent \textbf{Step 1:} We rewrite the target probability by definition and decompose it into two parts. 

Let $ \Delta_{\boldsymbol{x}_t} \equiv \mu(\boldsymbol{x}_t,1) -\mu(\boldsymbol{x}_t,0) $. Based on the definition of the probability of exploration and the form of the estimated optimal policy $ \widehat{\pi}_t(\boldsymbol{x}_t )$, we have
 \begin{eqnarray}\label{thm1_pf_1}
&&\kappa_t(\boldsymbol{x}_t) = \prob\{a_t\not = \widehat{\pi}_t(\boldsymbol{x}_t )\} =  \Mean[ \mathbb{I}\{a_t\not = \widehat{\pi}_t(\boldsymbol{x}_t )\}]\\\nonumber
 = &&\underbrace{\Mean[ \mathbb{I}(a_t = 0)|\widehat{\pi}_t(\boldsymbol{x}_t)=1]\prob\{\widehat{\pi}_t(\boldsymbol{x}_t)=1\}}_{\eta_0} + \underbrace{\Mean[ \mathbb{I}(a_t = 1)|\widehat{\pi}_t(\boldsymbol{x}_t)=0]\prob\{\widehat{\pi}_t(\boldsymbol{x}_t)=0\}}_{\eta_1},
\end{eqnarray} 
where the expectation is taken with respect to the history $\mathcal{H}_{t-1}$ before time point $t$.

 Next, we rewrite $\eta_0$ and $\eta_1$ using  the estimated mean and variance components $\widehat{\mu}_{t -1}(\boldsymbol{x}_t, a)$ and $\widehat{\sigma}_{t-1}(\boldsymbol{x}_t ,a)$, where $a=0,1$. We focus on $\eta_0$ first. 
 
 Given $\widehat{\pi}_t(\boldsymbol{x}_t)=1$, i.e., $ \widehat{\mu}_{t -1}(\boldsymbol{x}_t, 1)-\widehat{\mu}_{t -1}(\boldsymbol{x}_t, 0) >0$, based on the definition of the taken action in Lin-UCB that 
 \begin{eqnarray*}
a_t=\mathbb{I}\left\{\widehat{\mu}_{t -1}(\boldsymbol{x}_t ,1)+c_t \widehat{\sigma}_{t-1}(\boldsymbol{x}_t ,1) > \widehat{\mu}_{t -1}(\boldsymbol{x}_t ,0)+c_t \widehat{\sigma}_{t-1}(\boldsymbol{x}_t ,0)\right\},
 \end{eqnarray*}
  the probability of choosing action 0 rather than action 1 is
  \begin{equation*}\begin{aligned}
&\Mean[ \mathbb{I}(a_t = 0)|\widehat{\pi}_t(\boldsymbol{x}_t)=1] \\
= & \prob\left\{ \widehat{\mu}_{t -1}(\boldsymbol{x}_t, 1)+c_t \widehat{\sigma}_{t-1}(\boldsymbol{x}_t, 1) < \widehat{\mu}_{t -1}(\boldsymbol{x}_t, 0)+c_t \widehat{\sigma}_{t-1}(\boldsymbol{x}_t, 0) |\widehat{\pi}_t(\boldsymbol{x}_t)=1\right\}  \\
= & \prob\left\{ \widehat{\mu}_{t -1}(\boldsymbol{x}_t, 1)- \widehat{\mu}_{t -1}(\boldsymbol{x}_t, 0)< c_t \widehat{\sigma}_{t-1}(\boldsymbol{x}_t, 0)-c_t \widehat{\sigma}_{t-1}(\boldsymbol{x}_t, 1) |\widehat{\mu}_{t -1}(\boldsymbol{x}_t, 1)-\widehat{\mu}_{t -1}(\boldsymbol{x}_t, 0) >0 \right\}  \\
= &\prob\left[ 0<\widehat{\mu}_{t -1}(\boldsymbol{x}_t, 1)-\widehat{\mu}_{t -1}(\boldsymbol{x}_t, 0)  < c_t \left\{ \widehat{\sigma}_{t-1}(\boldsymbol{x}_t, 0)-\widehat{\sigma}_{t-1}(\boldsymbol{x}_t, 1) \right\} \right] /\prob\{\widehat{\pi}_t(\boldsymbol{x}_t)=1\},
\end{aligned} 
  \end{equation*}
  where the second equality is to rearrange the estimated mean and variance components, and the last equality comes from the definition of the conditional probability. Combining this with \eqref{thm1_pf_1}, we have
 \begin{eqnarray*}
  \eta_0=\prob\left[ 0<\widehat{\mu}_{t -1}(\boldsymbol{x}_t, 1)-\widehat{\mu}_{t -1}(\boldsymbol{x}_t, 0)  < c_t \left\{ \widehat{\sigma}_{t-1}(\boldsymbol{x}_t, 0)-\widehat{\sigma}_{t-1}(\boldsymbol{x}_t, 1) \right\} \right]. \end{eqnarray*} 
 
 Similarly, we have 
\begin{eqnarray*}
\eta_1=\prob\left[ c_t \left\{ \widehat{\sigma}_{t-1}(\boldsymbol{x}_t, 0)-\widehat{\sigma}_{t-1}(\boldsymbol{x}_t, 1) \right\}  <\widehat{\mu}_{t -1}(\boldsymbol{x}_t, 1)-\widehat{\mu}_{t -1}(\boldsymbol{x}_t, 0)  < 0 \right].
 \end{eqnarray*} 
Thus combined with Equation \eqref{thm1_pf_1}, we have
\begin{equation} \label{eq:proofPE}
\begin{aligned}
\kappa_t(\boldsymbol{x}_t) =  \eta_0+\eta_1 = & \prob\left[ 0<\widehat{\mu}_{t -1}(\boldsymbol{x}_t, 1)-\widehat{\mu}_{t -1}(\boldsymbol{x}_t, 0)  < c_t \left\{ \widehat{\sigma}_{t-1}(\boldsymbol{x}_t, 0)-\widehat{\sigma}_{t-1}(\boldsymbol{x}_t, 1) \right\} \right] \\
& +\prob\left[ c_t \left\{ \widehat{\sigma}_{t-1}(\boldsymbol{x}_t, 0)-\widehat{\sigma}_{t-1}(\boldsymbol{x}_t, 1) \right\}  <\widehat{\mu}_{t -1}(\boldsymbol{x}_t, 1)-\widehat{\mu}_{t -1}(\boldsymbol{x}_t, 0)  < 0 \right] \\
& = \prob\left\{\left|\widehat{\mu}_{t -1}(\boldsymbol{x}_t, 1)-\widehat{\mu}_{t -1}(\boldsymbol{x}_t, 0) \right| <  c_t \left| \widehat{\sigma}_{t-1}(\boldsymbol{x}_t, 0)-\widehat{\sigma}_{t-1}(\boldsymbol{x}_t, 1) \right|  \right\}.
\end{aligned}
\end{equation}
   
The rest of the proof is aims to bound the probability 
\begin{equation*}
\prob\left\{\left|\widehat{\mu}_{t -1}(\boldsymbol{x}_t, 1)-\widehat{\mu}_{t -1}(\boldsymbol{x}_t, 0) \right| <  c_t \left| \widehat{\sigma}_{t-1}(\boldsymbol{x}_t, 0)-\widehat{\sigma}_{t-1}(\boldsymbol{x}_t, 1) \right|  \right\}.
\end{equation*}

\smallskip \noindent \textbf{Step 2:} Secondly, we bound the variance $\widehat{\sigma}_{t-1}(\boldsymbol{x}_t, 0)$  and $\widehat{\sigma}_{t-1}(\boldsymbol{x}_t, 1)$ . 

We consider the quantity $\widehat{\sigma}_{t-1}(\boldsymbol{x}_t, 0) = \sqrt{\{\boldsymbol{x}_t^\top \{\boldsymbol{D}_{t-1}(0)^\top \boldsymbol{D}_{t-1}(0) +\omega \boldsymbol{I}_d \}^{-1}\boldsymbol{x}_t \}}$ first.
Let $\mathbf{v}$ be any $d\times 1$ vector, then the sample variance under action 0 is given by 
  \begin{equation}\label{thm1_pf_4}
\begin{aligned}
 \boldsymbol{x}_t^\top \{\boldsymbol{D}_{t-1}(0)^\top \boldsymbol{D}_{t-1}(0) +\omega \boldsymbol{I}_d \}^{-1}\boldsymbol{x}_t & =  \|\boldsymbol{x}_t\|_{2}^2 (\frac{\boldsymbol{x}_t}{ \|\boldsymbol{x}_t\|_{2}})^\top \{\boldsymbol{D}_{t-1}(0)^\top \boldsymbol{D}_{t-1}(0) +\omega \boldsymbol{I}_d \}^{-1}(\frac{\boldsymbol{x}_t}{ \|\boldsymbol{x}_t\|_{2}}) \\
& \leq \|\boldsymbol{x}_t\|_{2}^2 \max\limits_{\|\mathbf{v}\|_{2}=1} \mathbf{v}^\top \{\boldsymbol{D}_{t-1}(0)^\top \boldsymbol{D}_{t-1}(0) +\omega \boldsymbol{I}_d \}^{-1}\mathbf{v} \\
& \leq \|\boldsymbol{x}_t\|_{2}^2 \lambda_{\max}\{( \boldsymbol{D}_{t-1}(0)^\top \boldsymbol{D}_{t-1}(0) +\omega \boldsymbol{I}_d )^{-1}\},
\end{aligned}
  \end{equation}
 where the first inequality is to replace $({\boldsymbol{x}_t}/{ \|\boldsymbol{x}_t\|_{2}})^\top$ with any normalized vector, and the second inequality is due to the definition.   According to \eqref{thm1_pf_4}, combined with Assumption 4.1, we can further bound $\widehat{\sigma}_{t-1}(\boldsymbol{x}_t, 0)$ by 
  \begin{equation}\label{thm1_pf_5}
 \|\boldsymbol{x}_t\|_{2}\sqrt{\lambda_{\max}\{( \boldsymbol{D}_{t-1}(0)^\top \boldsymbol{D}_{t-1}(0) +\omega \boldsymbol{I}_d )^{-1}\} }  \leq \frac{L_{\boldsymbol{x}}}{\sqrt{\lambda_{\min}\{\boldsymbol{D}_{t-1}(0)^\top \boldsymbol{D}_{t-1}(0) +\omega \boldsymbol{I}_d \}}}.
  \end{equation}
It is immediate from \eqref{thm1_pf_4} and \eqref{thm1_pf_5} that  
  \begin{equation}\label{thm1_pf_6}
  \begin{aligned}
0<\widehat{\sigma}_{t-1}(\boldsymbol{x}_t, 0)\leq \frac{L_{\boldsymbol{x}}}{\sqrt{\lambda_{\min}\{ \boldsymbol{D}_{t-1}(0)^\top \boldsymbol{D}_{t-1}(0) +\omega \boldsymbol{I}_d \}}}.
\end{aligned}
  \end{equation}
 Note that 
 \begin{equation*}
\lambda_{\min}\{\boldsymbol{D}_{t-1}(0)^\top \boldsymbol{D}_{t-1}(0) +\omega \boldsymbol{I}_d \}  = \lambda_{\min}\{\boldsymbol{D}_{t-1}(0)^\top \boldsymbol{D}_{t-1}(0) \} + \omega ,
 \end{equation*}
combined with the fact that $\boldsymbol{D}_{t-1}(0)^\top \boldsymbol{D}_{t-1}(0) =\sum_{i=1}^{t-1} (1-a_i)\boldsymbol{x}_i \boldsymbol{x}_i^\top $, then  $\lambda_{\min}\{\boldsymbol{D}_{t-1}(0)^\top \boldsymbol{D}_{t-1}(0) +\omega \boldsymbol{I}_d \} $ can be further expressed as
  \begin{equation*}
\begin{aligned}
  \lambda_{\min}\{\boldsymbol{D}_{t-1}(0)^\top \boldsymbol{D}_{t-1}(0) +\omega \boldsymbol{I}_d \} & = 
(t-1)\lambda_{\min}\left\{\frac{1}{t-1}\sum_{i=1}^{t-1} (1-a_i)\boldsymbol{x}_i \boldsymbol{x}_i^\top \right\} + \omega \\
& >  (t-1)p_{t-1} \lambda_{\min}\left(\boldsymbol{\Sigma}\right)+ \omega  > (t-1)p_{t-1}\lambda + \omega,
\end{aligned}
  \end{equation*}
  where the first inequality is owing to Assumption 4.2 , and the second inequality is owing to Assumption 4.1. 
This together with \eqref{thm1_pf_6} gives the lower and upper bounds of  $\widehat{\sigma}_{t-1}(\boldsymbol{x}_t, 0) $ as 
  \begin{equation}\label{thm1_pf_9_0}
  0<\widehat{\sigma}_{t-1}(\boldsymbol{x}_t, 0) \leq \frac{L_{\boldsymbol{x}}}{\sqrt{(t-1)p_{t-1}\lambda + \omega  }}<  \frac{ L_{\boldsymbol{x}}}{\sqrt{(t-1)p_{t-1}\lambda }}. 
    \end{equation}
    Similarly we have 
      \begin{equation}\label{thm1_pf_9_1}
  0<\widehat{\sigma}_{t-1}(\boldsymbol{x}_t, 1) \leq  \frac{ L_{\boldsymbol{x}}}{\sqrt{(t-1)p_{t-1}\lambda }},
    \end{equation}
which follows that 
\begin{equation*}
 c_t \left| \widehat{\sigma}_{t-1}(\boldsymbol{x}_t, 0)-\widehat{\sigma}_{t-1}(\boldsymbol{x}_t, 1) \right| \leq  c_t  \left(\left| \widehat{\sigma}_{t-1}(\boldsymbol{x}_t, 0)\right|+\left| \widehat{\sigma}_{t-1}(\boldsymbol{x}_t, 1) \right| \right) \leq \frac{ 2 c_tL_{\boldsymbol{x}}}{\sqrt{(t-1)p_{t-1}\lambda }}.
\end{equation*}
Combining \eqref{eq:proofPE} and the above equation, we get the conclusion that 
\begin{equation} \label{eq:proofPE2}
\begin{aligned}
\kappa_t(\boldsymbol{x}_t) & \leq  \prob\left\{\left|\widehat{\mu}_{t -1}(\boldsymbol{x}_t, 1)-\widehat{\mu}_{t -1}(\boldsymbol{x}_t, 0) \right| <  c_t \left| \widehat{\sigma}_{t-1}(\boldsymbol{x}_t, 0)-\widehat{\sigma}_{t-1}(\boldsymbol{x}_t, 1) \right|  \right\} \\
& \leq  \prob\left\{\left|\widehat{\mu}_{t -1}(\boldsymbol{x}_t, 1)-\widehat{\mu}_{t -1}(\boldsymbol{x}_t, 0) \right| < \frac{ 2 c_tL_{\boldsymbol{x}}}{\sqrt{(t-1)p_{t-1}\lambda }}   \right\}. \\
\end{aligned}
\end{equation}

\smallskip \noindent \textbf{Step 3:} Lastly, we aim to bound $ \prob\left\{\left|\widehat{\mu}_{t -1}(\boldsymbol{x}_t, 1)-\widehat{\mu}_{t -1}(\boldsymbol{x}_t, 0) \right| < \frac{ 2 c_tL_{\boldsymbol{x}}}{\sqrt{(t-1)p_{t-1}\lambda }}   \right\} $ using the result in  Corollary 1. 

For any $\xi >0  $, define $E: = \{\left| \widehat{\mu}_t(\boldsymbol{x}_t,1)-\widehat{\mu}_t(\boldsymbol{x}_t,0) -  \Delta_{\boldsymbol{x}_t} \right|   \leq  \xi\}$, which satisfies $\prob \left\{ E \right \} \geq 1 - 4 d \exp \left\{-tp_{t}^{2} c_{\xi}\right\}$ by  Corollary 1. Then on the Event $E$, we have 
\begin{equation*}
\begin{aligned}
\left|\widehat{\mu}_{t -1}(\boldsymbol{x}_t, 1)-\widehat{\mu}_{t -1}(\boldsymbol{x}_t, 0) \right|
& = \left| \Delta_{\boldsymbol{x}_t} + \left\{ \widehat{\mu}_{t -1}(\boldsymbol{x}_t, 1)-\widehat{\mu}_{t -1}(\boldsymbol{x}_t, 0) -  \Delta_{\boldsymbol{x}_t}  \right\} \right|\\
& \geq \left|  \Delta_{\boldsymbol{x}_t} \right| - \left| \widehat{\mu}_t(\boldsymbol{x}_t,1)-\widehat{\mu}_t(\boldsymbol{x}_t,0) -  \Delta_{\boldsymbol{x}_t} \right|   \geq \left|  \Delta_{\boldsymbol{x}_t} \right| - \xi.
\end{aligned}
\end{equation*}
Thus for the probability $ \prob\left\{\left|\widehat{\mu}_{t -1}(\boldsymbol{x}_t, 1)-\widehat{\mu}_{t -1}(\boldsymbol{x}_t, 0) \right| < \frac{ 2 c_tL_{\boldsymbol{x}}}{\sqrt{(t-1)p_{t-1}\lambda }}   \right\} $, we have
\begin{equation} \label{eq:proofPE3}
\begin{aligned}
\kappa_t(\boldsymbol{x}_t)& \leq  \prob\left\{\left|\widehat{\mu}_{t -1}(\boldsymbol{x}_t, 1)-\widehat{\mu}_{t -1}(\boldsymbol{x}_t, 0) \right| < \frac{ 2 c_tL_{\boldsymbol{x}}}{\sqrt{(t-1)p_{t-1}\lambda }}   \right\} \\
&  \leq  
\prob \left\{ 
\left|\widehat{\mu}_{t -1}(\boldsymbol{x}_t, 1)-\widehat{\mu}_{t -1}(\boldsymbol{x}_t, 0) \right| < \frac{ 2 c_t L_{\boldsymbol{x}} }{\sqrt{(t-1)p_{t-1}\lambda }}  \bigm| E
 \right\}
+ \prob \left\{ E^{c} \right\} \\
& \leq  \prob\left\{  \left|  \Delta_{\boldsymbol{x}_t} \right| - \xi  < \frac{ 2 c_tL_{\boldsymbol{x}}}{\sqrt{(t-1)p_{t-1}\lambda }} \right\}+  4 d \exp \left\{-(t-1)p_{t-1}^{2} c_{\xi}\right\} \\
&  =   \prob\left\{  \left|  \Delta_{\boldsymbol{x}_t} \right|   < \frac{ 2 c_tL_{\boldsymbol{x}}}{\sqrt{(t-1)p_{t-1}\lambda }} + \xi  \right\}+  4 d \exp \left\{-(t-1)p_{t-1}^{2} c_{\xi}\right\}. 
\end{aligned}
\end{equation}
Sine $tp_t \rightarrow \infty$ as $t \rightarrow \infty$, for any constant  $\delta > \xi$, there exist large enough $t$ satisfying $\frac{ 2 c_tL_{\boldsymbol{x}}}{\sqrt{(t-1)p_{t-1}\lambda }}\leq \delta -\xi  $. Then by Assumption 4.3, there exists some constant $\gamma$ such that 
\begin{equation*}
\prob\left\{  \left|  \Delta_{\boldsymbol{x}_t} \right|   < \frac{ 2 c_tL_{\boldsymbol{x}}}{\sqrt{(t-1)p_{t-1}\lambda }} + \xi \right\} = \mathcal{O}\left\{  \left( \frac{ 2 c_tL_{\boldsymbol{x}}}{\sqrt{(t-1)p_{t-1}\lambda }} + \xi  \right)^\gamma \right\},
\end{equation*}
i.e., there exists some constant $C$ such that
\begin{equation*}
\prob\left\{  \left|  \Delta_{\boldsymbol{x}_t} \right|   < \frac{ 2 c_tL_{\boldsymbol{x}}}{\sqrt{(t-1)p_{t-1}\lambda }} + \xi \right\} = C \left( \frac{ 2 c_tL_{\boldsymbol{x}}}{\sqrt{(t-1)p_{t-1}\lambda }} + \xi  \right)^\gamma. 
\end{equation*}
Therefore, combined with the Equation \eqref{eq:proofPE3}, we have 
\begin{equation*}
\begin{aligned}
\kappa_t(\boldsymbol{x}_t)& \leq   C \left( \frac{ 2 c_tL_{\boldsymbol{x}}}{\sqrt{(t-1)p_{t-1}\lambda }} + \xi  \right)^\gamma +  4 d \exp \left\{-(t-1)p_{t-1}^{2} c_{\xi}\right\}. 
\end{aligned}
\end{equation*}

The proof is hence completed.

\subsubsection{Proof for  TS}
We next show the probability of exploration under TS consisting of three main steps:  
\begin{enumerate}
	\item 	We firstly define an event  $E: = \{\left| \widehat{\mu}_t(\boldsymbol{x}_t,1)-\widehat{\mu}_t(\boldsymbol{x}_t,0) -  \Delta_{\boldsymbol{x}_t} \right|   \leq  \xi\}$ for any $0< \xi < \left| \Delta_{\boldsymbol{x}_t} \right| /2$,  where the estimated difference between mean functions is close to the true difference. And we have $\prob \left\{ E \right \} \geq 1 - 4 d \exp \left\{-tp_{t}^{2} c_{\xi}\right\}$ by  Corollary 1.
	\item Next, we bound the probability of exploration on the event $E$.
	\item Lastly, we combine the results in the previous two steps to get the unconditioned probability of exploration .
\end{enumerate}

\smallskip \noindent \textbf{Step 1:} For any $0< \xi < \left| \Delta_{\boldsymbol{x}_t} \right| /2$,  define $E: = \{\left| \widehat{\mu}_t(\boldsymbol{x}_t,1)-\widehat{\mu}_t(\boldsymbol{x}_t,0) -  \Delta_{\boldsymbol{x}_t} \right|   \leq  \xi\}$, which satisfies $\prob \left\{ E \right \} \geq 1 - 4 d \exp \left\{-(t-1)p_{t-1}^{2} c_{\xi}\right\}$ by  Corollary 1. Then for the probability $\prob\{a_t\not = \widehat{\pi}_t(\boldsymbol{x}_t )\} $, we have
\begin{equation} \label{thm1_pf_ts2}
\prob\{a_t\not = \widehat{\pi}_t(\boldsymbol{x}_t )\}   \leq \prob\{a_t\not = \widehat{\pi}_t(\boldsymbol{x}_t ) |E \}  + \prob\{E^{c} \}  \leq \prob\{a_t\not = \widehat{\pi}_t(\boldsymbol{x}_t ) |E \}+  4 d \exp \left\{-(t-1)p_{t-1}^{2} c_{\xi}\right\}.
\end{equation}
Without loss of generality, we assume $\Delta_{\boldsymbol{x}_t} >0$, then  $E: = \{0<  \Delta_{\boldsymbol{x}_t}  - \xi  \leq \widehat{\mu}_t(\boldsymbol{x}_t,1)-\widehat{\mu}_t(\boldsymbol{x}_t,0)  \leq  \Delta_{\boldsymbol{x}_t}  + \xi \}$, which implies $\widehat{\pi}_t(\boldsymbol{x}_t)=1$.

Using the law of iterated expectations, based on the definition of the probability of exploration and the form of the estimated optimal policy $ \widehat{\pi}_t(\boldsymbol{x}_t )$, on the event $E$, we have
\begin{equation} \label{eq:proofPETS}
\begin{aligned}
&\prob\{a_t\not = 1 |E\} =  \Mean[ \mathbb{I}\{a_t\not = 1\}|E]=   \Mean\left(\Mean[ \mathbb{I}\{a_t = 0\}| \widehat{\mu}_{t-1}(\boldsymbol{x}_t,1) ,\widehat{\mu}_{t-1}(\boldsymbol{x}_t,0)   ]|E\right).
\end{aligned}
\end{equation}

\smallskip \noindent \textbf{Step 2:}  Next, we focus on deriving the bound of $\Mean[ \mathbb{I}\{a_t = 0\}| \widehat{\mu}_{t-1}(\boldsymbol{x}_t,1) ,\widehat{\mu}_{t-1}(\boldsymbol{x}_t,0)   ]$ on the Event $E$.

Recalling the bandit mechanism of TS, we have $a_t=\mathbb{I}\left\{\boldsymbol{x}_t^\top \boldsymbol{\beta}_t(1) > \boldsymbol{x}_t^\top \boldsymbol{\beta}_t(0) \right\}$, 
where $\boldsymbol{\beta}_t(a)$ is drawn from the posterior distribution of $\boldsymbol{\beta}(a)$ given by 
\begin{equation*}
    \mathcal{N}_d[\widehat{\boldsymbol{\beta}}_{t-1}(a), \rho^2\{\boldsymbol{D}_{t-1}(a)^\top \boldsymbol{D}_{t-1}(a) +\omega \boldsymbol{I}_d\}^{-1}].
\end{equation*} From the posterior distributions and the definitions of $\widehat{\mu}_{t-1}(\boldsymbol{x}_t,a)$ and $\widehat{\sigma}_{t-1}(\boldsymbol{x}_t,a)$, we have 
  \begin{equation*} 
\boldsymbol{x}_t^\top \boldsymbol{\beta}_t(a)    \sim
\mathcal{N}[\boldsymbol{x}_t^\top \widehat{\boldsymbol{\beta}}_{t-1}(a), \rho^2\boldsymbol{x}_t^\top\{\boldsymbol{D}_{t-1}(a)^\top \boldsymbol{D}_{t-1}(a) +\omega \boldsymbol{I}_d\}^{-1}\boldsymbol{x}_t ], 
  \end{equation*}
that is,
    \begin{equation*} 
\boldsymbol{x}_t^\top \boldsymbol{\beta}_t(a)   \sim \mathcal{N}[ \widehat{\mu}_{t-1}(\boldsymbol{x}_t,a), \rho^2\widehat{\sigma}_{t-1}(\boldsymbol{x}_t,a)^2 ]. 
  \end{equation*}
Notice that $\boldsymbol{x}_t^\top \boldsymbol{\beta}_t(1) $ and $\boldsymbol{x}_t^\top \boldsymbol{\beta}_t(0) $ are drawn independently, thus, 
  \begin{equation}\label{thm1_pf_ts1}
\boldsymbol{x}_t^\top \boldsymbol{\beta}_t(1) -\boldsymbol{x}_t^\top \boldsymbol{\beta}_t(0)    \sim \mathcal{N}[ \widehat{\mu}_{t-1}(\boldsymbol{x}_t,1) - \widehat{\mu}_{t-1}(\boldsymbol{x}_t,0), \rho^2\{\widehat{\sigma}_{t-1}(\boldsymbol{x}_t,1)^2+\widehat{\sigma}_{t-1}(\boldsymbol{x}_t,0)^2\} ]. 
  \end{equation}

Recall $a_t=\mathbb{I}\left\{\boldsymbol{x}_t^\top \boldsymbol{\beta}_t(1) > \boldsymbol{x}_t^\top \boldsymbol{\beta} _t(0) \right\}$ in TS, based on the posterior distribution in \eqref{thm1_pf_ts1}. Therefore, on the Event $E$  we have
  \begin{equation} \label{thm1_pf_ts3}
  \begin{aligned}
  & \Mean[ \mathbb{I}\{a_t = 0\}| \widehat{\mu}_{t-1}(\boldsymbol{x}_t,1) ,\widehat{\mu}_{t-1}(\boldsymbol{x}_t,0)   ] \\
& =\prob\left\{ \boldsymbol{x}_t^\top \boldsymbol{\beta}_t(1) -\boldsymbol{x}_t^\top \boldsymbol{\beta}_t(0)<0 |\widehat{\mu}_{t-1}(\boldsymbol{x}_t,1) ,\widehat{\mu}_{t-1}(\boldsymbol{x}_t,0)\right\}\\
 =&\Phi[- \{\widehat{\mu}_{t-1}(\boldsymbol{x}_t,1) - \widehat{\mu}_{t-1}(\boldsymbol{x}_t,0)\}/\sqrt{\rho^2\{\widehat{\sigma}_{t-1}(\boldsymbol{x}_t,1)^2+\widehat{\sigma}_{t-1}(\boldsymbol{x}_t,0)^2\}}] \\
  =&1 - \Phi[ \{\widehat{\mu}_{t-1}(\boldsymbol{x}_t,1) - \widehat{\mu}_{t-1}(\boldsymbol{x}_t,0)\}/\sqrt{\rho^2\{\widehat{\sigma}_{t-1}(\boldsymbol{x}_t,1)^2+\widehat{\sigma}_{t-1}(\boldsymbol{x}_t,0)^2\}}], 
 \end{aligned} 
  \end{equation}
   where $\Phi(\cdot)$ is the cumulative distribution function of the standard normal distribution. 
   Denote $\widehat{z}_t\equiv   \{\widehat{\mu}_{t-1}(\boldsymbol{x}_t,1) - \widehat{\mu}_{t-1}(\boldsymbol{x}_t,0)\}/\sqrt{\rho^2\{\widehat{\sigma}_{t-1}(\boldsymbol{x}_t,1)^2+\widehat{\sigma}_{t-1}(\boldsymbol{x}_t,0)^2\}} >0$, since $\widehat{\pi}_t(\boldsymbol{x}_t)= \mathbb{I}\left\{\widehat{\mu}_{t -1}(\boldsymbol{x},1)> \widehat{\mu}_{t -1}(\boldsymbol{x},0)\right\}=1$, i.e., $ \widehat{\mu}_{t -1}(\boldsymbol{x}_t, 1)-\widehat{\mu}_{t -1}(\boldsymbol{x}_t, 0) >0$. By applying the tail bound established for the normal distribution in Section 7.1 of \cite{feller2008introduction}, we have \eqref{thm1_pf_ts3} can be bounded as
  \begin{equation*}\begin{aligned}
 \Mean[ \mathbb{I}\{a_t = 0\}| \widehat{\mu}_{t-1}(\boldsymbol{x}_t,1) ,\widehat{\mu}_{t-1}(\boldsymbol{x}_t,0)   ] \leq  \exp(-\widehat{z}_t^2/2).
 \end{aligned} 
  \end{equation*}
This yields that on the Event $E$, 
  \begin{equation*} 
  \begin{aligned}
&  \Mean[ \mathbb{I}\{a_t = 0\}| \widehat{\mu}_{t-1}(\boldsymbol{x}_t,1) ,\widehat{\mu}_{t-1}(\boldsymbol{x}_t,0)   ] \\
&\leq  \Mean\left\{    \exp(-\widehat{z}_t^2/2)\right\} = \Mean\left\{    \exp\left(- \frac{\{\widehat{\mu}_{t-1}(\boldsymbol{x}_t,1) - \widehat{\mu}_{t-1}(\boldsymbol{x}_t,0)\}^2}{ 2\rho^2\{\widehat{\sigma}_{t-1}(\boldsymbol{x}_t,1)^2+\widehat{\sigma}_{t-1}(\boldsymbol{x}_t,0)^2\}}\right)\right\}.
 \end{aligned} 
  \end{equation*}
Using similar arguments in proving \eqref{thm1_pf_9_0}that $  \widehat{\sigma}_{t-1}(\boldsymbol{x}_t, 0)\leq   \frac{ L_{\boldsymbol{x}}}{\sqrt{(t-1)p_{t-1}\lambda }} $, we have  
  \begin{equation*}
\widehat{\sigma}_{t-1}(\boldsymbol{x}_t,1)^2+\widehat{\sigma}_{t-1}(\boldsymbol{x}_t,0)^2\leq \frac{ 2L_{\boldsymbol{x}}^2 }{{ {(t-1)p_{t-1} \lambda}}}.
  \end{equation*} 
  Therefore, combining the above two equations leads to
  \begin{equation*}
 \Mean[ \mathbb{I}\{a_t = 0\}| \widehat{\mu}_{t-1}(\boldsymbol{x}_t,1) ,\widehat{\mu}_{t-1}(\boldsymbol{x}_t,0)   ] \leq   \Mean\left\{    \exp\left(- \frac{\{\widehat{\mu}_{t-1}(\boldsymbol{x}_t,1) - \widehat{\mu}_{t-1}(\boldsymbol{x}_t,0)\}^2 { {(t-1)p_{t-1} \lambda}}}{4\rho^2 L_{\boldsymbol{x}}^2 }\right)\right\},
  \end{equation*}
  where the expectation is taken with respect to history $\mathcal{H}_{t-1}$.

 Note that on the Event $E$, we have 
\begin{equation*}
\{\widehat{\mu}_{t-1}(\boldsymbol{x}_t,1) - \widehat{\mu}_{t-1}(\boldsymbol{x}_t,0)\}^2 \geq \left(\left| \Delta_{\boldsymbol{x}_t} \right| - \xi   \right)^2,
\end{equation*}
which follows that on the Event $E$, 
\begin{equation} \label{thm1_pf_ts4}
\begin{aligned}
\Mean[ \mathbb{I}\{a_t = 0\}| \widehat{\mu}_{t-1}(\boldsymbol{x}_t,1) ,\widehat{\mu}_{t-1}(\boldsymbol{x}_t,0)   ]  &  \leq   \Mean \left\{    \exp\left(- \frac{\left(\left| \Delta_{\boldsymbol{x}_t} \right| - \xi   \right)^2  { {(t-1)p_{t-1} \lambda}} }{  4\rho^2 L_{\boldsymbol{x}}^2 }\right)\right\} \\
 &  \leq      \exp\left(- \frac{\left(\left| \Delta_{\boldsymbol{x}_t} \right| - \xi   \right)^2  { {(t-1)p_{t-1} \lambda}}}{ 4\rho^2 L_{\boldsymbol{x}}^2 }\right).
\end{aligned}
\end{equation}

\smallskip \noindent \textbf{Step 3:} Combined with Equation \eqref{thm1_pf_ts2}, Equation \eqref{eq:proofPETS} and Equation \eqref{thm1_pf_ts4},  we have
\begin{equation*}
\begin{aligned}
\prob\{a_t\not = \widehat{\pi}_t(\boldsymbol{x}_t )\}  & \overset{\eqref{thm1_pf_ts2}}{\leq} \prob\{a_t\not = \widehat{\pi}_t(\boldsymbol{x}_t ) |E \}+  4 d \exp \left\{-(t-1)p_{t-1}^{2} c_{\xi}\right\} \\
 & \overset{\eqref{eq:proofPETS}}{\leq}  \Mean\left(\Mean[ \mathbb{I}\{a_t = 0\}| \widehat{\mu}_{t-1}(\boldsymbol{x}_t,1) ,\widehat{\mu}_{t-1}(\boldsymbol{x}_t,0)   ]|E\right)+  4 d \exp \left\{-(t-1)p_{t-1}^{2} c_{\xi}\right\} \\
  & \overset{\eqref{thm1_pf_ts4}}{\leq}   \exp\left(- \frac{\left(\left| \Delta_{\boldsymbol{x}_t} \right| - \xi   \right)^2  { {(t-1)p_{t-1} \lambda}} }{  4\rho^2 L_{\boldsymbol{x}}^2 }\right)+  4 d \exp \left\{-(t-1)p_{t-1}^{2} c_{\xi}\right\}.
\end{aligned} 
\end{equation*}
The proof is hence completed.

\subsection{Proof of Theorem 2}
We detail the proof of Theorem 2 in this section. 
Using the similar arguments in \eqref{decompose_beta_pf} in the proof of Lemma 4.1, we can rewrite $\sqrt{t}\{\widehat{\boldsymbol{\beta}}_t(a)- \boldsymbol{\beta}(a)\}$ as
\begin{eqnarray*} 
\sqrt{t}\{\widehat{\boldsymbol{\beta}}_t(a)- \boldsymbol{\beta}(a)\} = &&\underbrace{\left \{ \frac{1}{t} \sum_{i=1}^{t} \mathbb{I}(a_i=a)\boldsymbol{x}_i \boldsymbol{x}_i^\top +\frac{1}{t} \omega \boldsymbol{I}_d \right\}^{-1}}_{\boldsymbol{\xi}} \underbrace{\left \{\frac{1}{\sqrt{t}}\sum_{i=1}^{t} \mathbb{I}(a_i=a)\boldsymbol{x}_i  e_i \right\}}_{\boldsymbol{\eta}_1}\\
&&-\underbrace{\left \{ \frac{1}{t} \sum_{i=1}^{t} \mathbb{I}(a_i=a)\boldsymbol{x}_i \boldsymbol{x}_i^\top +\frac{1}{t}\omega \boldsymbol{I}_d \right\}^{-1} \frac{\omega}{\sqrt{t}} \boldsymbol{\beta}(a)}_{\boldsymbol{\eta}_2}.
\end{eqnarray*} 
Our goal is to prove that $\sqrt{t}\{\widehat{\boldsymbol{\beta}}_t(a)- \boldsymbol{\beta}(a)\}$ is asymptotically normal. The proof is to generalize Theorem 3.1 in \cite{chen2020statistical} by considering commonly used bandit algorithms, including UCB, TS, and EG here. 
We complete the proof in the following four steps:
\begin{itemize}
	\item Step 1: Prove that $\boldsymbol{\eta}_1 =  ({1}/{\sqrt{t}})\sum_{i=1}^t \mathbb{I}(a_i=a)\boldsymbol{x}_i  e_i  \stackrel{D}{\longrightarrow} \mathcal{N}_{d}\left(\boldsymbol{0}_d, G_{a}\right)$, where $G_{a}$ is the variance matrix to be spesified shortly.
	\item Step 2: Prove that $\boldsymbol{\xi} = \left \{ (1/t) \sum_{i=1}^{t} \mathbb{I}(a_i=a)\boldsymbol{x}_i \boldsymbol{x}_i^\top +(\omega/t)\boldsymbol{I}_d \right\}^{-1} \stackrel{p}{\longrightarrow} \sigma_a^2 G_a^{-1}$, where $\sigma^2_a = \Mean(e_t^2|a_t=a)$ for $a=0,1$.
	\item Step 3: Prove that $\boldsymbol{\eta}_2 = \left \{ (1/t) \sum_{i=1}^{t} \mathbb{I}(a_i=a)\boldsymbol{x}_i \boldsymbol{x}_i^\top +(\omega/t)\boldsymbol{I}_d \right\}^{-1} ({\omega}/{\sqrt{t}}) \boldsymbol{\beta}(a) \overset{p}{\longrightarrow} \boldsymbol{0}_d$.
	\item Step 4: Combine above results in steps 1-3 using Slutsky's theorem.
\end{itemize}

\smallskip \noindent \textbf{Step 1}: We first focus on proving that $\boldsymbol{\eta}_1 =({1}/{\sqrt{t}}) \sum_{i=1}^t \mathbb{I}(a_i=a)\boldsymbol{x}_i  e_i  \stackrel{D}{\longrightarrow} \mathcal{N}_{d}\left(\boldsymbol{0}_d, G_{a}\right)$. Using Cramer-Wold device, it suffices to show that for any $\boldsymbol{v} \in \mathbb{R}^{d}$,
\begin{eqnarray*}
\boldsymbol{\eta}_1(\boldsymbol{v}) \equiv \frac{1}{\sqrt{t}}\sum_{i=1}^t \mathbb{I}(a_i=a) \boldsymbol{v}^{\top}\boldsymbol{x}_i  e_i    \stackrel{D}{\longrightarrow}  \mathcal{N}_d\left(0, \boldsymbol{v}^{\top} G_{a}\boldsymbol{v}\right).
\end{eqnarray*}
Note that $\Mean \left(\mathbb{I}(a_i=a) \boldsymbol{v}^{\top}\boldsymbol{x}_i  e_i  \mid \mathcal{H}_{i-1}\right) =\Mean \left(\mathbb{I}(a_i=a) \boldsymbol{v}^{\top}\boldsymbol{x}_i \mid \mathcal{H}_{i-1}\right) \Mean \left(e_i  \mid \mathcal{H}_{i-1},a_i=a\right)  =0$, we have that $\mathbb{I}(a_i=a) \boldsymbol{v}^{\top}\boldsymbol{x}_i  e_i$ is a Martingale difference sequence. We next show the asymptotic normality of $\boldsymbol{\eta}_1(\boldsymbol{v})$ using Martingale central limit theorem, by the following two parts: i) check the conditional Lindeberg condition; ii)  derivative the limit of the conditional variance.

\textbf{Firstly, we check the conditional Lindeberg condition.} For any $\delta>0$, denote
\begin{eqnarray} \label{psi}
\psi = \sum_{i=1}^t \mathbb{E}\left[\frac{1}{t} \mathbb{I}(a_i=a)\left( \boldsymbol{v}^{\top}  \boldsymbol{x}_i\right)^{2}  e_i^{2} \mathbb{I}\left\{\left|\frac{1}{\sqrt{t}}\mathbb{I}(a_i=a) \boldsymbol{v}^{\top}  \boldsymbol{x}_i  e_i\right|>\delta \right\} \mid \mathcal{H}_{i-1}\right].
\end{eqnarray}
Notice that $\left( \boldsymbol{v}^{\top}  \boldsymbol{x}_i\right)^{2} \leq \|\boldsymbol{v}\|_{2}^{2} L_{\boldsymbol{x}}^{2}d$, we have 
\begin{equation*}
\mathbb{I}\left\{\left|\frac{1}{\sqrt{t}}\mathbb{I}(a_i=a) \boldsymbol{v}^{\top}  \boldsymbol{x}_i  e_i\right|>\delta \right\} \leq \mathbb{I}\left\{\mathbb{I}(a_i=a)e_i^2 \|\boldsymbol{v}\|_{2}^{2} L_{\boldsymbol{x}}^{2}d >t \delta^2\right\}  = \mathbb{I}\left\{\mathbb{I}(a_i=a) e_i^{2}>\frac{ t \delta^{2}}{\|\boldsymbol{v}\|_{2}^{2} L_{\boldsymbol{x}}^{2}d }\right\}.
\end{equation*}
Combining this with \eqref{psi}, we obtain that 
\begin{eqnarray}\label{psi_v2}
\psi \leq \frac{\|\boldsymbol{v}\|_{2}^{2} L_{\boldsymbol{x}}^{2} d}{t} \sum_{i=1}^t \mathbb{E}\left(\mathbb{I}(a_i=a) e_i^{2} \mathbb{I}\left\{\mathbb{I}(a_i=a) e_i^{2}>\frac{ t \delta^{2}}{\|\boldsymbol{v}\|_{2}^{2} L_{\boldsymbol{x}}^{2}d }\right\}\mid \mathcal{H}_{i-1} \right),
\end{eqnarray}
where the right hand side equals
\begin{eqnarray*}
\frac{\|\boldsymbol{v}\|_{2}^{2} L_{\boldsymbol{x}}^{2}d }{t} \sum_{i=1}^t \mathbb{E}\left(\mathbb{I}(a_i=a)  \mid \mathcal{H}_{i-1}\right) \mathbb{E}\left(e_i^{2} \mathbb{I}\left\{\mathbb{I}(a_i=a) e_i^{2}>\frac{\delta^{2} t}{\|\boldsymbol{v}\|_{2}^{2} L_{\boldsymbol{x}}^{2} d}\right\} \mid  \mathcal{H}_{i-1}\right).
\end{eqnarray*}
Then, we can further write \eqref{psi_v2} as
\begin{eqnarray}\label{psi_v3}
\psi \leq \frac{\|\boldsymbol{v}\|_{2}^{2} L_{\boldsymbol{x}}^{2}d }{t} \sum_{i=1}^t \mathbb{E}\left(e_{i(a)}^{2} \mathbb{I}\left\{ e_{i(a)}^{2}>\frac{\delta^{2} t}{\|\boldsymbol{v}\|_{2}^{2} L_{\boldsymbol{x}}^{2} d}\right\} \mid  \mathcal{H}_{i-1}\right).
\end{eqnarray}
where $e_{i(a)}=e_i$ when $a_i = a$ and 0 otherwise. Since $e_i$ conditioned on $a_i$ are $i.i.d.$, $\forall i$, we have the right hand side of the above inequality equals
\begin{eqnarray*}
 \|\boldsymbol{v}\|_{2}^{2} L_{\boldsymbol{x}}^{2}d  \mathbb{E}\left(e^{2} \mathbb{I}\left\{ e^{2}>\frac{t\delta^{2} }{\|\boldsymbol{v}\|_{2}^{2} L_{\boldsymbol{x}}^{2} d}\right\} \right),
\end{eqnarray*}
where $e$ is the random variable given by $e_i|\mathcal{H}_{i-1}$.  
Note that $e^{2} \mathbb{I}\left\{ e^{2}>{t\delta^{2} }/({\|\boldsymbol{v}\|_{2}^{2} L_{\boldsymbol{x}}^{2} d})\right\}$ is dominated by $e^{2}$ with $\Mean e^{2} < \infty$ and converges to 0, as $t \rightarrow \infty$. Then, by Dominated Convergence Theorem, the results in \eqref{psi_v3} can be further bounded by 
\begin{equation*}
 \psi \leq \frac{\|\boldsymbol{v}\|_{2}^{2} L_{\boldsymbol{x}}^{2}d }{t} \sum_{i=1}^t \mathbb{E}\left(e^{2} \mathbb{I}\left\{ e^{2}>\frac{t\delta^{2} }{\|\boldsymbol{v}\|_{2}^{2} L_{\boldsymbol{x}}^{2} d}\right\} \right) \rightarrow 0,  \text{  as } t \rightarrow \infty.  
\end{equation*}
Therefore, conditional Lindeberg condition holds.

\textbf{Secondly, we derive the limit of the conditional variance.} Notice that 
\begin{eqnarray*}
\frac{1}{t}\sum_{i=1}^{t} \mathbb{E}\left[ \mathbb{I}(a_i=a)\left( \boldsymbol{v}^{\top}  \boldsymbol{x}_i\right)^{2}  e_i^{2}\mid \mathcal{H}_{i-1}\right] 
&& = \frac{1}{t}\sum_{i=1}^{t}\mathbb{E}\left\{ \mathbb{E}\left[ \mathbb{I}(a_i=a)\left( \boldsymbol{v}^{\top}  \boldsymbol{x}_i\right)^{2}  e_i^{2}\mid a_i,\boldsymbol{x}_i \right] \mid \mathcal{H}_{i-1} \right\}\\
&& = \frac{1}{t}\sum_{i=1}^{t}\mathbb{E}\left\{  \mathbb{I}(a_i=a)\left( \boldsymbol{v}^{\top}  \boldsymbol{x}_i\right)^{2}  \mathbb{E}\left[e_i^{2}\mid a_i=a,\boldsymbol{x}_i \right] \mid \mathcal{H}_{i-1} \right\}.
\end{eqnarray*}  
Since $e_t$ is independent of $\mathcal{H}_{i-1}$ and $\boldsymbol{x}_i$ given $a_t$, and $ \mathbb{E}\left[e_i^{2}\mid  a_i=a \right] =\sigma_a^2 $, we have
\begin{eqnarray*}
\frac{1}{t}\sum_{i=1}^{t}\mathbb{E}\left[ \mathbb{I}(a_i=a)\left( \boldsymbol{v}^{\top}  \boldsymbol{x}_i\right)^{2}  e_i^{2}\mid \mathcal{H}_{i-1}\right] 
&&= \frac{1}{t}\sum_{i=1}^{t}\mathbb{E}\left[ \mathbb{I}(a_i=a)\left( \boldsymbol{v}^{\top}  \boldsymbol{x}_i\right)^{2}  \mathbb{E}\left[e_i^{2}\mid  a_i=a \right]  \mid \mathcal{H}_{i-1} \right]  \\
&&= \frac{1}{t}\sum_{i=1}^{t}\mathbb{E}\left[ \mathbb{I}(a_i=a) \boldsymbol{v}^{\top}  \boldsymbol{x}_i \boldsymbol{x}_i^{\top}  \boldsymbol{v}\sigma_a^2  \mid \mathcal{H}_{i-1}\right]\\
&&= \frac{1}{t}\sum_{i=1}^{t}\sigma_a^2 \mathbb{E}\left[ \mathbb{I}(a_i=a) \boldsymbol{v}^{\top}  \boldsymbol{x}_i \boldsymbol{x}_i^{\top}  \boldsymbol{v} \mid \mathcal{H}_{i-1}\right],
\end{eqnarray*}  
where 
\begin{eqnarray*}
 \mathbb{E}\left[ \mathbb{I}(a_i=a) \boldsymbol{v}^{\top}  \boldsymbol{x}_i \boldsymbol{x}_i^{\top}  \boldsymbol{v}  \mid \mathcal{H}_{i-1}\right] 
 =&& \mathbb{E}\{ \mathbb{E}\left[ \mathbb{I}(a_i\neq \pi^{*}( \boldsymbol{x}_i))\mathbb{I}(\pi^{*}( \boldsymbol{x}_i)\neq a) \boldsymbol{v}^{\top}  \boldsymbol{x}_i \boldsymbol{x}_i^{\top}  \boldsymbol{v} \mid \boldsymbol{x}_i \right] \mid \mathcal{H}_{i-1}\} \\ \nonumber
 + &&  \mathbb{E}\{  \mathbb{E}\left[ \mathbb{I}(a_i=\pi^{*}( \boldsymbol{x}_i))\mathbb{I}(\pi^{*}( \boldsymbol{x}_i)=a) \boldsymbol{v}^{\top}  \boldsymbol{x}_i \boldsymbol{x}_i^{\top}  \boldsymbol{v}  \mid \boldsymbol{x}_i \right] \mid \mathcal{H}_{i-1}\} \\ \nonumber
 =&& \mathbb{E}\{ \mathbb{E}\left[ \mathbb{I}(a_i\neq \pi^{*}( \boldsymbol{x}_i)) \mid \boldsymbol{x}_i,\mathcal{H}_{i-1} \right] \mathbb{I}(\pi^{*}( \boldsymbol{x}_i)\neq a) \boldsymbol{v}^{\top}  \boldsymbol{x}_i \boldsymbol{x}_i^{\top}  \boldsymbol{v}\} \\ \nonumber
 + &&  \mathbb{E}\{  \mathbb{E}\left[ \mathbb{I}(a_i=\pi^{*}( \boldsymbol{x}_i)) \mid \boldsymbol{x}_i,\mathcal{H}_{i-1} \right] \mathbb{I}(\pi^{*}( \boldsymbol{x}_i)=a) \boldsymbol{v}^{\top}  \boldsymbol{x}_i \boldsymbol{x}_i^{\top}  \boldsymbol{v}\} .
 \end{eqnarray*}
Here, the first equation comes from iteration expectation over $\boldsymbol{x}_{i}$ and the fact that $\mathbb{I}(\pi^{*}( \boldsymbol{x}_i)\neq a) \boldsymbol{v}^{\top}  \boldsymbol{x}_i \boldsymbol{x}_i^{\top}  \boldsymbol{v}$ is a constant given $ \boldsymbol{x}_i $ and $\mathcal{H}_{i-1}$.
 \begin{eqnarray*}
\begin{aligned}
\mathbb{I}(a_i= a) & =\mathbb{I}(a_i= a\neq \pi^{*}( \boldsymbol{x}_i))+ \mathbb{I}(a_i= a = \pi^{*}( \boldsymbol{x}_i))\\
& =\mathbb{I}(a_i\neq \pi^{*}( \boldsymbol{x}_i))\mathbb{I}(\pi^{*}( \boldsymbol{x}_i)\neq a)+ \mathbb{I}(a_i=\pi^{*}( \boldsymbol{x}_i))\mathbb{I}(\pi^{*}( \boldsymbol{x}_i)=a), 
\end{aligned}
\end{eqnarray*}
and the second equation is owing to the fact that $\mathbb{I}(\pi^{*}( \boldsymbol{x}_i)=a) \boldsymbol{v}^{\top}  \boldsymbol{x}_i \boldsymbol{x}_i^{\top}  \boldsymbol{v}$ is a constant given  $\boldsymbol{x}_{i}$ and independent of $\mathcal{H}_{i-1}$. 
Define 
\begin{equation} \label{eq:nudef}
\nu_{i}\left(\boldsymbol{x}_{i},\mathcal{H}_{i-1}\right)
 \equiv \operatorname{Pr}\left\{a_{i} \neq \pi^{*}( \boldsymbol{x}_i) \right |\boldsymbol{x}_i,\mathcal{H}_{i-1} \}=\mathbb{E}\left[\mathbb{I}\left\{a_{i} \neq \pi^{*}( \boldsymbol{x}_i) \right\}|\boldsymbol{x}_i,\mathcal{H}_{i-1}\right],
\end{equation}
then the conditional variance can be expressed as 
 \begin{eqnarray}\label{mclt_varv}
\begin{aligned}
   \frac{1}{t}\sum_{i=1}^{t}\mathbb{E}\left[ \mathbb{I}(a_i=a) \boldsymbol{v}^{\top}  \boldsymbol{x}_i \boldsymbol{x}_i^{\top}  \boldsymbol{v}  \mid \mathcal{H}_{i-1}\right]
 & =  \frac{1}{t}\sum_{i=1}^{t} \mathbb{E}\{\nu_{i}\left(\boldsymbol{x}_{i},\mathcal{H}_{i-1}\right) \mathbb{I}(\pi^{*}( \boldsymbol{x}_i)\neq a) \boldsymbol{v}^{\top}  \boldsymbol{x}_i \boldsymbol{x}_i^{\top}  \boldsymbol{v}\} \\ \nonumber
 & +    \frac{1}{t}\sum_{i=1}^{t} \mathbb{E}\left[  \{1-\nu_{i}\left(\boldsymbol{x}_{i},\mathcal{H}_{i-1}\right) \} \mathbb{I}(\pi^{*}( \boldsymbol{x}_i)=a) \boldsymbol{v}^{\top}  \boldsymbol{x}_i \boldsymbol{x}_i^{\top}  \boldsymbol{v}  \right],
\end{aligned}
\end{eqnarray}
which can be expressed as
\begin{eqnarray}
\begin{aligned}
  &  \frac{1}{t}\sum_{i=1}^{t}\int \nu_{i}\left(\boldsymbol{x},\mathcal{H}_{i-1}\right)\mathbb{I}\{\boldsymbol{x}^{\top} \boldsymbol{\beta}(a) < \boldsymbol{x}^{\top}\boldsymbol{\beta}(1-a) \}  \boldsymbol{v}^{\top}  \boldsymbol{x} \boldsymbol{x}^{\top}  \boldsymbol{v}  d P_{\mathcal{X}} \\\nonumber
& +    \frac{1}{t}\sum_{i=1}^{t}\int \{1-\nu_{i}\left(\boldsymbol{x},\mathcal{H}_{i-1}\right) \}  \mathbb{I}\left\{\boldsymbol{x}^{\top} \boldsymbol{\beta}(a) \geq \boldsymbol{x}^{\top}\boldsymbol{\beta}(1-a)\right\} \boldsymbol{v}^{\top}  \boldsymbol{x} \boldsymbol{x}^{\top}  \boldsymbol{v}  d P_{\mathcal{X}} \\ 
  &  =\int \frac{1}{t}\sum_{i=1}^{t}\nu_{i}\left(\boldsymbol{x},\mathcal{H}_{i-1}\right)\mathbb{I}\{\boldsymbol{x}^{\top} \boldsymbol{\beta}(a) < \boldsymbol{x}^{\top}\boldsymbol{\beta}(1-a) \}  \boldsymbol{v}^{\top}  \boldsymbol{x} \boldsymbol{x}^{\top}  \boldsymbol{v}  d P_{\mathcal{X}} \\\nonumber
& +   \int \{1- \frac{1}{t}\sum_{i=1}^{t}\nu_{i}\left(\boldsymbol{x},\mathcal{H}_{i-1}\right) \}   \mathbb{I}\left\{\boldsymbol{x}^{\top} \boldsymbol{\beta}(a) \geq \boldsymbol{x}^{\top}\boldsymbol{\beta}(1-a)\right\} \boldsymbol{v}^{\top}  \boldsymbol{x} \boldsymbol{x}^{\top}  \boldsymbol{v}  d P_{\mathcal{X}}.
\end{aligned}
\end{eqnarray}
Since $\lim_{i\rightarrow \infty}\prob\{a_i \neq \pi^{*}(\boldsymbol{x}) \} = \kappa_{\infty}(\boldsymbol{x})$, we have for any $\epsilon >0 $, there exist a constant $t_0 >0$ such that $\left|\prob\{a_i \neq \pi^{*}(\boldsymbol{x}) \} - \kappa_{\infty}(\boldsymbol{x})\right| < \epsilon$ for all $i \geq t_0$. Therefore,  for the expectation of $\frac{1}{t}\sum_{i=1}^{t}\nu_{i}\left(\boldsymbol{x},\mathcal{H}_{i-1}\right)$ over the history, we have
\begin{equation*}
 \Mean \left[\frac{1}{t}\sum_{i=1}^{t}\nu_{i}\left(\boldsymbol{x},\mathcal{H}_{i-1}\right)\right] = \frac{1}{t}\sum_{i=1}^{t}\Mean \left[\nu_{i}\left(\boldsymbol{x},\mathcal{H}_{i-1}\right)\right] = \frac{1}{t}\sum_{i=1}^{t}\prob\{a_i \neq \pi^{*}(\boldsymbol{x}) \}.
\end{equation*}
It follows immediately that 
\begin{equation*}
\begin{aligned}
 &\Mean \left[\frac{1}{t}\sum_{i=1}^{t}\nu_{i}\left(\boldsymbol{x},\mathcal{H}_{i-1}\right)\right] -\kappa_{\infty}(\boldsymbol{x})\\
 &  =  \frac{1}{t}\sum_{i=1}^{t_0}\left[\prob\{a_i \neq \pi^{*}(\boldsymbol{x}) \} -\kappa_{\infty}(\boldsymbol{x})\right]+\frac{1}{t}\sum_{i=t_0}^{t}\left[\prob\{a_i \neq \pi^{*}(\boldsymbol{x}) \} -\kappa_{\infty}(\boldsymbol{x})\right].
\end{aligned}
\end{equation*}
Therefore, by the triangle inequality, we have
\begin{equation*}
\begin{aligned}
 &\left|\Mean \left[\frac{1}{t}\sum_{i=1}^{t}\nu_{i}\left(\boldsymbol{x},\mathcal{H}_{i-1}\right)\right] -\kappa_{\infty}(\boldsymbol{x})\right| \\
 &  =  \frac{1}{t}\sum_{i=1}^{t_0}\left|\prob\{a_i \neq \pi^{*}(\boldsymbol{x}) \} -\kappa_{\infty}(\boldsymbol{x})\right|+\frac{1}{t}\sum_{i=t_0}^{t}\left|\prob\{a_i \neq \pi^{*}(\boldsymbol{x}) \} -\kappa_{\infty}(\boldsymbol{x})\right|\\
 & < \frac{1}{t}\sum_{i=1}^{t_0}\left[\left|\prob\{a_i \neq \pi^{*}(\boldsymbol{x}) \} \right| +\left|\kappa_{\infty}(\boldsymbol{x})\right|\right]+\frac{1}{t}\sum_{i=t_0}^{t}\epsilon \\
 & \leq \frac{1}{t}\sum_{i=1}^{t_0}2+\frac{1}{t}\sum_{i=t_0}^{t}\epsilon = \frac{2t_0 }{t} + \frac{t-t_0}{t}\epsilon. \\
\end{aligned}
\end{equation*}
Since the above equation holds for any $\epsilon>0$ and $0<\frac{t-t_0}{t}<1$, we have
\begin{equation*}
\left|\Mean \left[\frac{1}{t}\sum_{i=1}^{t}\nu_{i}\left(\boldsymbol{x},\mathcal{H}_{i-1}\right)\right] -\kappa_{\infty}(\boldsymbol{x})\right| \leq  \frac{2t_0 }{t},
\end{equation*}
which goes to zero as $t \rightarrow \infty$. Thus, 
\begin{equation}\label{proof:expectation}
\Mean \left[\frac{1}{t}\sum_{i=1}^{t}\nu_{i}\left(\boldsymbol{x},\mathcal{H}_{i-1}\right)\right] =\kappa_{\infty}(\boldsymbol{x}) + o_p(1).
\end{equation}
Next, we consider the variance of $\frac{1}{t}\sum_{i=1}^{t}\nu_{i}\left(\boldsymbol{x},\mathcal{H}_{i-1}\right)$. Denote $\Mean \left[\frac{1}{t}\sum_{i=1}^{t}\nu_{i}\left(\boldsymbol{x},\mathcal{H}_{i-1}\right)\right] = \mu_{\nu}(\boldsymbol{x})$, we have $\mu_{\nu}(\boldsymbol{x})=\kappa_{\infty}(\boldsymbol{x}) + o_p(1)$. Notice that $\nu_{i}\left(\boldsymbol{x}_{i},\mathcal{H}_{i-1}\right)
 \equiv \operatorname{Pr}\left\{a_{i} \neq \pi^{*}( \boldsymbol{x}_i) \right |\boldsymbol{x}_i,\mathcal{H}_{i-1} \} \in [0,1]$, by Lemma \ref{lemma:bounded}, we have 
\begin{equation*}
\begin{aligned}
 \operatorname{Var}\left[\frac{1}{t}\sum_{i=1}^{t}\nu_{i}\left(\boldsymbol{x},\mathcal{H}_{i-1}\right)\right]\leq \mu_{\nu}(\boldsymbol{x}) -  \mu_{\nu}(\boldsymbol{x})^2 = \kappa_{\infty}(\boldsymbol{x}) \left\{ 1-\kappa_{\infty}(\boldsymbol{x}) \right\} + o_p(1),
\end{aligned}
\end{equation*}
which goes to zero as $t\rightarrow \infty$. Combined with Equation \eqref{proof:expectation}, it follows immediately that as $t$ goes to $\infty$, we have
\begin{equation*}
\frac{1}{t}\sum_{i=1}^{t}\nu_{i}\left(\boldsymbol{x},\mathcal{H}_{i-1}\right) \rightarrow \kappa_{\infty}(\boldsymbol{x}).
\end{equation*}
Therefore, as $t$ goes to $\infty$, we have  $\frac{1}{t}\sum_{i=1}^{t}\mathbb{E}\left[ \mathbb{I}(a_i=a) \boldsymbol{v}^{\top}  \boldsymbol{x}_i \boldsymbol{x}_i^{\top}  \boldsymbol{v}  \mid \mathcal{H}_{i-1}\right]$ converges to

\begin{eqnarray} \label{eq:C39}
   &&\int \kappa_{\infty} (\boldsymbol{x}) \mathbb{I}\{\boldsymbol{x}^{\top}  \boldsymbol{\beta}(a) < \boldsymbol{x}^{\top}\boldsymbol{\beta}(1-a) \}  \boldsymbol{v}^{\top}  \boldsymbol{x} \boldsymbol{x}^{\top}  \boldsymbol{v}  d P_{\mathcal{X}} \\ \nonumber
   &&+\int \{1- \kappa_{\infty} (\boldsymbol{x}) \}\mathbb{I}\left\{\boldsymbol{x}^{\top}  \boldsymbol{\beta}(a) \geq \boldsymbol{x}^{\top}\boldsymbol{\beta}(1-a)\right\} \boldsymbol{v}^{\top}  \boldsymbol{x} \boldsymbol{x}^{\top}  \boldsymbol{v}  d P_{\mathcal{X}} .
\end{eqnarray}
Thus, following the similar arguments in S1.2 in \cite{chen2020statistical}, we have
\begin{equation*}
\boldsymbol{\eta}_1 (\boldsymbol{v})=  \frac{1}{\sqrt{t}}\sum_{i=1}^t \mathbb{I}(a_i=a) \boldsymbol{v}^{\top}\boldsymbol{x}_i  e_i    \stackrel{D}{\longrightarrow}  \mathcal{N}_d\left(0, \boldsymbol{v}^{\top} G_{a}\boldsymbol{v}\right),    
\end{equation*}
where
\begin{equation*}\begin{aligned}
 \boldsymbol{v}^{\top} G_{a}\boldsymbol{v} = \sigma_a^2  &\left\{  \int  \kappa_{\infty} (\boldsymbol{x})  \mathbb{I}\{\boldsymbol{x}^{\top}  \boldsymbol{\beta}(a) < \boldsymbol{x}^{\top}\boldsymbol{\beta}(1-a) \}  \boldsymbol{v}^{\top}  \boldsymbol{x} \boldsymbol{x}^{\top}  \boldsymbol{v}  d P_{\mathcal{X}} \right. \\
& \left.  + \int \{1- \kappa_{\infty} (\boldsymbol{x}) \} \mathbb{I}\left\{\boldsymbol{x}^{\top}  \boldsymbol{\beta}(a) \geq \boldsymbol{x}^{\top}\boldsymbol{\beta}(1-a)\right\} \boldsymbol{v}^{\top}  \boldsymbol{x} \boldsymbol{x}^{\top}  \boldsymbol{v}  d P_{\mathcal{X}} \right\}.
\end{aligned}\end{equation*}
Finally, by Martingale Central Limit Theorem, we have
\begin{equation*}
\boldsymbol{\eta}_1  =  \frac{1}{\sqrt{t}}\sum_{i=1}^t \mathbb{I}(a_i=a)  \boldsymbol{x}_i  e_i    \stackrel{D}{\longrightarrow}  \mathcal{N}_d\left(0,   G_{a} \right),    
\end{equation*}
where
\begin{equation}\label{mclt_var_eta}\begin{aligned} 
 G_{a}  = \sigma_a^2  & \left\{   \int  \kappa_{\infty} (\boldsymbol{x}) \mathbb{I}\{\boldsymbol{x}^{\top}  \boldsymbol{\beta}(a) < \boldsymbol{x}^{\top}\boldsymbol{\beta}(1-a) \}   \boldsymbol{x} \boldsymbol{x}^{\top}   d P_{\mathcal{X}}  \right. \\
&  \left. +\int\{1- \kappa_{\infty} (\boldsymbol{x})\} \mathbb{I}\left\{\boldsymbol{x}^{\top}  \boldsymbol{\beta}(a) \geq \boldsymbol{x}^{\top}\boldsymbol{\beta}(1-a)\right\}   \boldsymbol{x} \boldsymbol{x}^{\top} d P_{\mathcal{X}} \right\}  .
\end{aligned}\end{equation}The first part is thus completed.

\smallskip \noindent \textbf{Step 2:} We next show that $\boldsymbol{\xi} =\left \{ (1/t) \sum_{i=1}^{t} \mathbb{I}(a_i=a)\boldsymbol{x}_i \boldsymbol{x}_i^\top +\frac{1}{t}\omega \boldsymbol{I}_d \right\}^{-1} \stackrel{p}{\longrightarrow} \sigma_a^2 G_a^{-1}$, which is sufficient to find the limit of $\left \{ (1/t) \sum_{i=1}^{t} \mathbb{I}(a_i=a)\boldsymbol{x}_i \boldsymbol{x}_i^\top +\frac{1}{t} \omega\boldsymbol{I}_d \right\}$. By Lemma 6 in \cite{chen2020statistical}, it suffices to show the limit of $ \frac{1}{t} \sum_{i=1}^{t} \mathbb{I}(a_i=a)\boldsymbol{v}^{\top}\boldsymbol{x}_i \boldsymbol{x}_i^\top\boldsymbol{v}$ for any $\boldsymbol{v} \in \mathbb{R}^{d}$.

Since $\prob(| \mathbb{I}(a_i=a)\boldsymbol{v}^{\top}\boldsymbol{x}_i \boldsymbol{x}_i^\top\boldsymbol{v}| > h)\leq \prob(|\boldsymbol{v}^{\top}\boldsymbol{x} \boldsymbol{x}^T\boldsymbol{v}| > h)$ for each $h>0$ and $i\leq 1$, by Theorem 2.19 in \cite{hall2014martingale}, we have
\begin{equation}\label{thm2_step2_cons}
 \frac{1}{t} \sum_{i=1}^{t}\left[ \mathbb{I}(a_i=a)\boldsymbol{v}^{\top}\boldsymbol{x}_i \boldsymbol{x}_i^\top\boldsymbol{v}  - \Mean \left\{ \mathbb{I}(a_i=a)\boldsymbol{v}^{\top}\boldsymbol{x}_i \boldsymbol{x}_i^\top\boldsymbol{v}\mid   \mathcal{H}_{i-1}\right\} \right] \overset{p}{\longrightarrow} 0, \text{  as }t \rightarrow \infty.  
\end{equation}
Recall the results in \eqref{mclt_varv} and \eqref{mclt_var_eta}, we have $ \Mean \left\{ \mathbb{I}(a_i=a)\boldsymbol{v}^{\top}\boldsymbol{x}_i \boldsymbol{x}_i^\top\boldsymbol{v}   \mid   \mathcal{H}_{i-1}\right\} =  {G_{a}}/{\sigma_a^2 }$. Combining this with \eqref{thm2_step2_cons}, 
we have
\begin{equation*}
 \frac{1}{t} \sum_{i=1}^{t} \mathbb{I}(a_i=a)\boldsymbol{v}^{\top}\boldsymbol{x}_i \boldsymbol{x}_i^\top\boldsymbol{v}  \overset{p}{\longrightarrow} \frac{G_{a}}{\sigma_a^2 },  \text{  as }t \rightarrow \infty.  
\end{equation*}
By Lemma 6 in \cite{chen2020statistical} and Continuous Mapping Theorem, we further  have
\begin{equation*}
\boldsymbol{\xi} =\left \{ \frac{1}{t} \sum_{i=1}^{t} \mathbb{I}(a_i=a)\boldsymbol{x}_i \boldsymbol{x}_i^\top +\frac{1}{t}\omega \boldsymbol{I}_d \right\}^{-1} \stackrel{p}{\longrightarrow} \sigma_a^2 G_a^{-1}.
\end{equation*}

\smallskip \noindent \textbf{Step 3:} We focus on proving $\boldsymbol{\eta}_2 = \left \{ (1/t) \sum_{i=1}^{t} \mathbb{I}(a_i=a)\boldsymbol{x}_i \boldsymbol{x}_i^\top +(\omega/t) \boldsymbol{I}_d \right\}^{-1} ({\omega}/{\sqrt{t}}) \boldsymbol{\beta}(a) \overset{p}{\longrightarrow} \boldsymbol{0}_d$ next.
This suffices to show that $\boldsymbol{b}_i^\top\boldsymbol{\eta}_2 \overset{p}{\longrightarrow} 0$ holds for any standard basis $\boldsymbol{b}_i \in \mathbb{R}^d$. Since
\begin{equation*}
\begin{aligned}
\boldsymbol{b}_i^\top\boldsymbol{\eta}_2  & = \left\|\boldsymbol{b}_i^\top \left \{ \frac{1}{t} \sum_{i=1}^{t} \mathbb{I}(a_i=a)\boldsymbol{x}_i \boldsymbol{x}_i^\top +\frac{1}{t}\omega\boldsymbol{I}_d \right\}^{-1} \frac{\omega}{\sqrt{t}} \boldsymbol{\beta}(a)  \right\|_{2} \\
& \leq \frac{\omega}{\sqrt{t}} \left\|\boldsymbol{b}_i^\top\right\|_{2}  \left\|\left \{ \frac{1}{t} \sum_{i=1}^{t} \mathbb{I}(a_i=a)\boldsymbol{x}_i \boldsymbol{x}_i^\top +\frac{1}{t}\omega\boldsymbol{I}_d \right\}^{-1}\right\|_{2}  \left\|  \boldsymbol{\beta}(a)  \right\|_{2},
\end{aligned}
\end{equation*}
and by \eqref{l2norm}, we have
\begin{equation*}
 \boldsymbol{b}_i^\top\boldsymbol{\eta}_2 \leq   \frac{ \omega \left\|  \boldsymbol{\beta}(a)  \right\|_{2}}{\sqrt{tp_t^2} \lambda+\frac{1}{\sqrt{t}}\omega } .
\end{equation*}
Thus, we have $\boldsymbol{b}_i^\top\boldsymbol{\eta}_2 \overset{p}{\longrightarrow} 0$, as $ tp_t^2 \rightarrow \infty$.
 
\smallskip \noindent \textbf{Step 4:} Finally, we combine the above results using Slutsky's theorem, and conclude that
\begin{equation*}
 \sqrt{t}\{\widehat{\boldsymbol{\beta}}_t(a)- \boldsymbol{\beta}(a)\}= \boldsymbol{\xi}\boldsymbol{\eta}_1+\boldsymbol{\eta}_2  \stackrel{D}{\longrightarrow} \mathcal{N}_{d}\left(\boldsymbol{0}_d, \sigma_a^4 G_a^{-1}\right) ,  
\end{equation*}
where $G_{a}$ is defined in \eqref{mclt_var_eta}. Denote the variance term as
\begin{equation*}
\begin{aligned}
\sigma_{\boldsymbol{\beta}(a)}^2  = \sigma_a^2  &\left[  \int   \kappa_{\infty} (\boldsymbol{x})\mathbb{I}\{\boldsymbol{x}^{\top} \boldsymbol{\beta}(a) < \boldsymbol{x}^{\top}\boldsymbol{\beta}(1-a) \}   \boldsymbol{x} \boldsymbol{x}^{\top}   d P_{\mathcal{X}} \right. \\
& \left.  + \int \{1-  \kappa_{\infty} (\boldsymbol{x})\} \mathbb{I}\left\{\boldsymbol{x}^{\top} \boldsymbol{\beta}(a) \geq \boldsymbol{x}^{\top}\boldsymbol{\beta}(1-a)\right\}   \boldsymbol{x} \boldsymbol{x}^{\top}   d P_{\mathcal{X}} \right]^{-1},
\end{aligned}
 \end{equation*}
with $\sigma^2_a = \Mean(e_t^2|a_t=a)$ denoting the conditional variance of $e_t$ given $a_t = a$, for $a=0,1$, we have
\begin{equation*}
\sqrt{t}\left\{\widehat{\boldsymbol{\beta}}_t(a)-\boldsymbol{\beta}(a)\right\}  \stackrel{D}{\longrightarrow}  \mathcal{N}_d\{ \boldsymbol{0}_d, \sigma_{\boldsymbol{\beta}(a)}^2\}.
\end{equation*}  
The proof is hence completed.

\subsection{Proof of Theorem 3}

Finally, we prove the asymptotic normality of the proposed value estimator under DREAM in Theorem 3 in this section. The proof consists of four steps. In step 1, we aim to show
 \begin{equation*} 
\widehat{V}_T =  \widetilde{V}_T  +o_p(T^{-1/2}),
\end{equation*} 
where 
 \begin{equation*}
\widetilde{V}_T=  \frac{1}{T}\sum_{t=1}^T \frac{\mathbb{I}\{a_t=\widehat{\pi}_t(\boldsymbol{x}_t )\} }{ 1-\kappa_{t}(\boldsymbol{x}_t)} \Big[ r_t - {\mu} \{\boldsymbol{x}_t ,\widehat{\pi}_t(\boldsymbol{x}_t )\}\Big] +  {\mu} \{\boldsymbol{x}_t ,\widehat{\pi}_t(\boldsymbol{x}_t ) \}.
\end{equation*} 
Next,  in Step 2, we establish
 \begin{equation*}
\widetilde{V}_T =  \overline{V}_T +o_p(T^{-1/2}),
\end{equation*}
where 
 \begin{equation*}
\overline{V}_T=  \frac{1}{T}\sum_{t=1}^T \frac{\mathbb{I}\{a_t={\pi}^*(\boldsymbol{x}_t )\} }{ \prob\{a_t={\pi}^*(\boldsymbol{x}_t)\}} \Big[ r_t - {\mu} \{\boldsymbol{x}_t ,{\pi}^*(\boldsymbol{x}_t )\}\Big] +  {\mu} \{\boldsymbol{x}_t ,{\pi}^*(\boldsymbol{x}_t ) \}.
\end{equation*} 
The above two steps yields that
 \begin{equation}\label{thm4_p1_mid2}
\widehat{V}_T  = \overline{V}_T + o_p(T^{-1/2}).
\end{equation} 
Then, in Step 3, based on \eqref{thm4_p1_mid2} and Martingale  Central Limit Theorem, we show
 \begin{equation*}
\sqrt{T}(\widehat{V}_T -V^*) \stackrel{D}{\longrightarrow}  \mathcal{N}\left(0,  \sigma_{DR}^2 \right),
\end{equation*}
with 
\begin{equation*}
\sigma_{DR}^2 = \int_{\boldsymbol{x}} \frac{ \pi^*(\boldsymbol{x} ) \sigma_{1}^2 +\{1-\pi^*(\boldsymbol{x} )\}\sigma_{0}^2 }{1-\kappa_\infty(\boldsymbol{x})} d {P_{\mathcal{X}}} +\Var \left[  {\mu} \{\boldsymbol{x} ,\pi^*(\boldsymbol{x} ) \} \right] ,
\end{equation*}
where $\sigma^2_a = \Mean(e_t^2|a_t=a)$ for $a=0,1$, and $\kappa_{t}\rightarrow \kappa_{\infty}$ as $t\rightarrow \infty$.

Lastly, in Step 4, we show the variance estimator in Equation (5) in the main paper is a consistent estimator of $\sigma_{DR}^2$. The proof for Theorem 3 is thus completed.

\smallskip \noindent \textbf{Step 1:} We first show $\widehat{V}_T =  \widetilde{V}_T  +o_p(T^{-1/2})$. To this end, define a middle term as
\begin{equation*}
\widetilde{\phi}_T=  \frac{1}{T}\sum_{t=1}^T\frac{\mathbb{I}\{a_t=\widehat{\pi}_t(\boldsymbol{x}_t )\} }{ 1-\kappa_{t}(\boldsymbol{x}_t)} \Big[ r_t - \widehat{\mu}_{t-1} \{\boldsymbol{x}_t ,\widehat{\pi}_t(\boldsymbol{x}_t )\}\Big] +  \widehat{\mu}_{t-1} \{\boldsymbol{x}_t ,\widehat{\pi}_t(\boldsymbol{x}_t ) \}.
\end{equation*}
Thus, it suffices to show $\widehat{V}_T =  \widetilde{\phi}_T  +o_p(T^{-1/2})$ and $\widetilde{\phi}_T =  \widetilde{V}_T  +o_p(T^{-1/2})$. 

Firstly, we have 
\begin{eqnarray}\label{mid_step1}
\widehat{V}_T - \widetilde{\phi}_T&&=  \frac{1}{T}\sum_{t=1}^T \left[\frac{\mathbb{I}\{a_t=\widehat{\pi}_t(\boldsymbol{x}_t )\}}{1-\widehat{\kappa}_{t}(\boldsymbol{x}_t )}-\frac{\mathbb{I}\{a_t=\widehat{\pi}_t(\boldsymbol{x}_t )\}}{1-\kappa_{t}(\boldsymbol{x}_t)}\right] \Big[ r_t - \widehat{\mu}_{t-1} \{\boldsymbol{x}_t ,\widehat{\pi}_t(\boldsymbol{x}_t )\}\Big] \\\nonumber
&&=  \frac{1}{T}\sum_{t=1}^T \left\{\widehat{\kappa}_{t}(\boldsymbol{x}_t) -\kappa_{t}(\boldsymbol{x}_t)\right\}\left[\frac{\mathbb{I}\{a_t=\widehat{\pi}_t(\boldsymbol{x}_t )\}\Big[ r_t - \widehat{\mu}_{t-1} \{\boldsymbol{x}_t ,\widehat{\pi}_t(\boldsymbol{x}_t )\}\Big] }{\{1-\widehat{\kappa}_{t}(\boldsymbol{x}_t)\}\{1-\kappa_{t}(\boldsymbol{x}_t)\}}\right] .
\end{eqnarray}
 We can further decompose \eqref{mid_step1} by
\begin{eqnarray}\nonumber
\widehat{V}_T - \widetilde{\phi}_T=  \frac{1}{T}\sum_{t=1}^T &&\left\{\widehat{\kappa}_{t}(\boldsymbol{x}_t) -\kappa_{t}(\boldsymbol{x}_t)\right\}\\\nonumber
&&\left[\frac{\mathbb{I}\{a_t=\widehat{\pi}_t(\boldsymbol{x}_t )\}\Big[ r_t - {\mu}  \{\boldsymbol{x}_t ,\widehat{\pi}_t(\boldsymbol{x}_t )\}+{\mu}  \{\boldsymbol{x}_t ,\widehat{\pi}_t(\boldsymbol{x}_t )\} -\widehat{\mu}_{t-1} \{\boldsymbol{x}_t ,\widehat{\pi}_t(\boldsymbol{x}_t )\}\Big] }{\{1-\widehat{\kappa}_{t}(\boldsymbol{x}_t)\}\{1-\kappa_{t}(\boldsymbol{x}_t)\}}\right] \\\label{mid_step2}
=  \frac{1}{T}\sum_{t=1}^T &&\left\{\widehat{\kappa}_{t}(\boldsymbol{x}_t) -\kappa_{t}(\boldsymbol{x}_t)\right\}\left[\frac{\mathbb{I}\{a_t=\widehat{\pi}_t(\boldsymbol{x}_t )\}\Big[ r_t - {\mu}  \{\boldsymbol{x}_t ,\widehat{\pi}_t(\boldsymbol{x}_t )\}\Big]}{\{1-\widehat{\kappa}_{t}(\boldsymbol{x}_t)\}\{1-\kappa_{t}(\boldsymbol{x}_t)\}}\right] \\\nonumber
+ \frac{1}{T}\sum_{t=1}^T &&\left\{\widehat{\kappa}_{t}(\boldsymbol{x}_t) -\kappa_{t}(\boldsymbol{x}_t)\right\}\left[\frac{\mathbb{I}\{a_t=\widehat{\pi}_t(\boldsymbol{x}_t )\}\Big[  {\mu}  \{\boldsymbol{x}_t ,\widehat{\pi}_t(\boldsymbol{x}_t )\} -\widehat{\mu}_{t-1} \{\boldsymbol{x}_t ,\widehat{\pi}_t(\boldsymbol{x}_t )\}\Big]}{\{1-\widehat{\kappa}_{t}(\boldsymbol{x}_t)\}\{1-\kappa_{t}(\boldsymbol{x}_t)\}}\right] .
\end{eqnarray}

We first show that the first term in \eqref{mid_step2} is $o_p(T^{-1/2})$. 
Define a class of function 
\begin{eqnarray*}
\mathcal{F}_\kappa(\boldsymbol{x}, a, r) =  \Bigg\{ \left\{\widehat{\kappa}(\boldsymbol{x}) -\kappa(\boldsymbol{x})\right\}\left[\frac{\mathbb{I}\{a=\pi(\boldsymbol{x} )\}\Big[ r - {\mu}  \{\boldsymbol{x} ,\pi(\boldsymbol{x} )\}\Big] }{\{1-\widehat{\kappa}(\boldsymbol{x})\}\{1-\kappa(\boldsymbol{x})\}}\right]: \widehat{\kappa}(\cdot), \kappa(\cdot)\in \Lambda, \pi(\cdot)  \in \Pi \Bigg\},
\end{eqnarray*} 
where $\Pi$ and $\Lambda$ are two classes of functions that maps context $\boldsymbol{x}\in \mathcal{X}$ to a probability.
\\
Define the supremum of the empirical process indexed by $ \mathcal{F}_\kappa $ as
\begin{eqnarray} \label{eq:defG}
||\mathbb{G}_n||_{\mathcal{F}} \equiv &&\sup_{\pi \in \Pi} \frac{1}{\sqrt{T}}\sum_{t=1}^{T} \left[  \mathcal{F}_\kappa (\boldsymbol{x}_t, a_t, r_t) - \mathbb{E}\{  \mathcal{F}_\kappa (\boldsymbol{x}_t, a_t, r_t) | \mathcal{H}_{t-1}\} \right].
\end{eqnarray}
Notice that 
\begin{equation*}
\begin{aligned}
& \mathbb{E}\{  \mathcal{F}_\kappa (\boldsymbol{x}_t, a_t, r_t) | \mathcal{H}_{t-1}\}\\
=& \mathbb{E}\left(\left\{\widehat{\kappa}(\boldsymbol{x}_t) -\kappa_{t}(\boldsymbol{x}_t)\right\}\left[\frac{\mathbb{I}\{a_t=\widehat{\pi}_t(\boldsymbol{x}_t )\}\Big[ r_t - {\mu}  \{\boldsymbol{x}_t ,\widehat{\pi}_t(\boldsymbol{x}_t )\}\Big] }{\{1-\widehat{\kappa}(\boldsymbol{x}_t)\}\{1-\kappa_{t}(\boldsymbol{x}_t)\}}\right]\mid \mathcal{H}_{t-1}\right)\\
= & \mathbb{E}\left(\left\{\widehat{\kappa}(\boldsymbol{x}_t) -\kappa_{t}(\boldsymbol{x}_t)\right\}\left[\frac{\mathbb{I}\{a_t=\widehat{\pi}_t(\boldsymbol{x}_t )\} e_t}{\{1-\widehat{\kappa}(\boldsymbol{x}_t)\}\{1-\kappa_{t}(\boldsymbol{x}_t)\}}\right]\mid \mathcal{H}_{t-1}\right),
\end{aligned}
\end{equation*}
by the definitions and thus, using the iteration of expectation, we have
\begin{equation*}
\begin{aligned}
& \mathbb{E}\{  \mathcal{F}_\kappa (\boldsymbol{x}_t, a_t, r_t) | \mathcal{H}_{t-1}\}\\
= & \mathbb{E} \left\{ \mathbb{E}\left(\left\{\widehat{\kappa}(\boldsymbol{x}_t) -\kappa_{t}(\boldsymbol{x}_t)\right\}\left[\frac{\mathbb{I}\{a_t=\widehat{\pi}_t(\boldsymbol{x}_t )\} e_t }{\{1-\widehat{\kappa}(\boldsymbol{x}_t)\}\{1-\kappa_{t}(\boldsymbol{x}_t)\}}\right]\mid a_t,  \boldsymbol{x}_t\right)\mid \mathcal{H}_{t-1} \right\}\\
= & \mathbb{E} \left\{ \left\{\widehat{\kappa}(\boldsymbol{x}_t) -\kappa_{t}(\boldsymbol{x}_t)\right\}\left[\frac{\mathbb{E}\{\mathbb{I}\{a_t=\widehat{\pi}_t(\boldsymbol{x}_t )\}\mid \boldsymbol{x}_t\}}{\{1-\widehat{\kappa}(\boldsymbol{x}_t)\}\{1-\kappa_{t}(\boldsymbol{x}_t)\}}\right] \mathbb{E}\left(e_t \mid a_t, \boldsymbol{x}_t\right)\mid \mathcal{H}_{t-1} \right\} = 0,
\end{aligned}
\end{equation*}
where the last equation is due to the definition of the noise $e_t$.
Therefore, Equation \eqref{eq:defG} can be further written as 
\begin{equation*}
||\mathbb{G}_n||_{\mathcal{F}} \equiv  \sup_{\pi \in \Pi} \frac{1}{\sqrt{T}}\sum_{t=1}^{T}  \mathcal{F}_\kappa (\boldsymbol{x}_t, a_t, r_t).
\end{equation*}
Next, we show the second moment is bounded by
\begin{equation*}
\begin{aligned}
& \mathbb{E}\left\{ \left(\left\{\widehat{\kappa}_{t}(\boldsymbol{x}_t) -\kappa_{t}(\boldsymbol{x}_t)\right\}\left[\frac{\mathbb{I}\{a_t=\widehat{\pi}_t(\boldsymbol{x}_t )\}\Big[ r_t - {\mu}  \{\boldsymbol{x}_t ,\widehat{\pi}_t(\boldsymbol{x}_t )\}\Big] }{\{1-\widehat{\kappa}_{t}(\boldsymbol{x}_t)\}\{1-\kappa_{t}(\boldsymbol{x}_t)\}}\right]\right)^2\mid \mathcal{H}_{t-1}\right\}\\
=&  \mathbb{E}\left\{ \left[\frac{ \left\{\widehat{\kappa}_{t}(\boldsymbol{x}_t) -\kappa_{t}(\boldsymbol{x}_t)\right\} }{\{1-\widehat{\kappa}_{t}(\boldsymbol{x}_t)\}\{1-\kappa_{t}(\boldsymbol{x}_t)\}}\right]^2\mathbb{I}\{a_t=\widehat{\pi}_t(\boldsymbol{x}_t )\}\Big[ r_t - {\mu}  \{\boldsymbol{x}_t ,\widehat{\pi}_t(\boldsymbol{x}_t )\}\Big]^2\mid \mathcal{H}_{t-1}\right\}\\
=&  \mathbb{E}\left\{ \left[\frac{ \left\{\widehat{\kappa}_{t}(\boldsymbol{x}_t) -\kappa_{t}(\boldsymbol{x}_t)\right\} }{\{1-\widehat{\kappa}_{t}(\boldsymbol{x}_t)\}\{1-\kappa_{t}(\boldsymbol{x}_t)\}}\right]^2\mathbb{I}\{a_t=\widehat{\pi}_t(\boldsymbol{x}_t )\}e_t^2\mid \mathcal{H}_{t-1}\right\}\end{aligned},
\end{equation*}
by the definitions and thus, using the iteration of expectation, we have
\begin{equation*}
\begin{aligned}
& \mathbb{E}\left\{ \left(\left\{\widehat{\kappa}_{t}(\boldsymbol{x}_t) -\kappa_{t}(\boldsymbol{x}_t)\right\}\left[\frac{\mathbb{I}\{a_t=\widehat{\pi}_t(\boldsymbol{x}_t )\}\Big[ r_t - {\mu}  \{\boldsymbol{x}_t ,\widehat{\pi}_t(\boldsymbol{x}_t )\}\Big] }{\{1-\widehat{\kappa}_{t}(\boldsymbol{x}_t)\}\{1-\kappa_{t}(\boldsymbol{x}_t)\}}\right]\right)^2\mid \mathcal{H}_{t-1}\right\}\\
=&  \mathbb{E} \left(\mathbb{E} \left\{ \left[\frac{ \left\{\widehat{\kappa}_{t}(\boldsymbol{x}_t) -\kappa_{t}(\boldsymbol{x}_t)\right\} }{\{1-\widehat{\kappa}_{t}(\boldsymbol{x}_t)\}\{1-\kappa_{t}(\boldsymbol{x}_t)\}}\right]^2\mathbb{I}\{a_t=\widehat{\pi}_t(\boldsymbol{x}_t )\}e_t^2 \mid a_t,  \boldsymbol{x}_t\right\} \mid \mathcal{H}_{t-1}\right)\\
=&  \mathbb{E} \left( \left[\frac{ \left\{\widehat{\kappa}_{t}(\boldsymbol{x}_t) -\kappa_{t}(\boldsymbol{x}_t)\right\} }{\{1-\widehat{\kappa}_{t}(\boldsymbol{x}_t)\}\{1-\kappa_{t}(\boldsymbol{x}_t)\}}\right]^2\mathbb{I}\{a_t=\widehat{\pi}_t(\boldsymbol{x}_t )\}\mathbb{E} \left\{e_t^2 \mid a_t, \boldsymbol{x}_t\right\} \mid   \mathcal{H}_{t-1}\right)\\
=&  \mathbb{E} \left( \left[\frac{1}{1-\widehat{\kappa}_{t}(\boldsymbol{x}_t)} -\frac{1}{1-\kappa_{t}(\boldsymbol{x}_t)}  \right]^2\mathbb{I}\{a_t=\widehat{\pi}_t(\boldsymbol{x}_t )\} \sigma_{a_t}^2 \mid   \mathcal{H}_{t-1}\right).
\end{aligned}
\end{equation*}
Notice that $\widehat{\kappa}_{t}(\boldsymbol{x}_t) \leq C_1<1$ and  $\kappa_{t}(\boldsymbol{x}_t) \leq C_1<1$ for sure for some constant $C_1<1$ (by definition of a valid bandit algorithm and results of Theorem 1), we have
\begin{equation*}
\left[\frac{1}{1-\widehat{\kappa}_{t}(\boldsymbol{x}_t)} -\frac{1}{1-\kappa_{t}(\boldsymbol{x}_t)}  \right]^2 \leq \left\{\left| \frac{1}{1-\widehat{\kappa}_{t}(\boldsymbol{x}_t)} \right|+\left|\frac{1}{1-\kappa_{t}(\boldsymbol{x}_t)}\right|  \right\}^2 \leq \left(\frac{2}{1-C_1}\right)^2\equiv C_2,
\end{equation*}
where $C_2$ is a bounded constant. 
Thus we have 
\begin{equation*}
\begin{aligned}
 & \mathbb{E} \left( \left[\frac{1}{1-\widehat{\kappa}_{t}(\boldsymbol{x}_t)} -\frac{1}{1-\kappa_{t}(\boldsymbol{x}_t)}  \right]^2\mathbb{I}\{a_t=\widehat{\pi}_t(\boldsymbol{x}_t )\} \sigma_{a_t}^2 \mid   \mathcal{H}_{t-1}\right)\\   
& \leq  \mathbb{E} \left( C_2 \mathbb{I}\{a_t=\widehat{\pi}_t(\boldsymbol{x}_t )\} \sigma_{a_t}^2 \mid   \mathcal{H}_{t-1}\right)\\
& \leq  C_2 \max\{\sigma_{0}^2,\sigma_{1}^2 \}.
\end{aligned}
\end{equation*}
Therefore, for the second moment of the inner term of the first term, we have 
\begin{equation*}
\begin{aligned}
&\sum_{t=1}^T \mathbb{E}\left\{  \left(\frac{1}{\sqrt{T}}\left\{\widehat{\kappa}_{t}(\boldsymbol{x}_t) -\kappa_{t}(\boldsymbol{x}_t)\right\}\left[\frac{\mathbb{I}\{a_t=\widehat{\pi}_t(\boldsymbol{x}_t )\}\Big[ r_t - {\mu}  \{\boldsymbol{x}_t ,\widehat{\pi}_t(\boldsymbol{x}_t )\}\Big] }{\{1-\widehat{\kappa}_{t}(\boldsymbol{x}_t)\}\{1-\kappa_{t}(\boldsymbol{x}_t)\}}\right]\right)^2\mid \mathcal{H}_{t-1}\right\}\\
&\leq \sum_{t=1}^T \frac{ C_2 \max\{\sigma_{0}^2,\sigma_{1}^2 \}}{T }=C_2 \max\{\sigma_{0}^2,\sigma_{1}^2 \} < \infty.
\end{aligned}
\end{equation*}
Therefore, we have 
\begin{equation*}
d_{1}(f) \equiv \left\|\mathbb{E}\left(\mid  \mathcal{F}_\kappa (\boldsymbol{x}_1, a_1, r_1) \| \mathcal{H}_{0}\right)\right\|_{\infty}  <\infty,
\end{equation*}
and
\begin{equation*}
d_{2}(f)\equiv \left\|\mathbb{E}\left(\left(\mathcal{F}_\kappa (\boldsymbol{x}_1, a_1, r_1)\right)^{2} \mid \mathcal{H}_{0}\right)\right\|_{\infty}^{1 / 2}<\infty.
\end{equation*}
It follows from  the maximal inequality developed in Section 4.2 of \cite{dedecker2002maximal} that there exist some constant $K\ge 1$ such that
\begin{eqnarray*}
\Mean \Big[||\mathbb{G}_n||_{\mathcal{F}} \Big] \lesssim   K\left(\sqrt{p} d_{2}(f)+\frac{1}{\sqrt{T}}\left\|\max_{1 \leq t \leq T}\left| \mathcal{F}_\kappa (\boldsymbol{x}_t, a_t, r_t) -\mathbb{E}\left( \mathcal{F}_\kappa (\boldsymbol{x}_t, a_t, r_t)  \mid \mathcal{H}_{t-1}\right)\right|\right\|\right).
\end{eqnarray*}
The above right-hand-side is upper
 bounded by 
\begin{eqnarray*}
O(1)\sqrt{T^{-1/2}},
\end{eqnarray*}
 where $O(1)$ denotes some universal constant. Hence, we have
\begin{eqnarray} \label{eq:resultG}
\Mean \Big[||\mathbb{G}_n||_{\mathcal{F}} \Big] = \mathcal{O}_p(T^{-1/2}).
\end{eqnarray}
Combined with Equation \eqref{eq:defG}, we have
\begin{equation*}
\frac{1}{\sqrt{T}}\sum_{t=1}^{T}  \mathcal{F}_\kappa (\boldsymbol{x}_t, a_t, r_t) = \mathcal{O}_p(T^{-1/2}). 
\end{equation*}
Therefore, for the first term in \eqref{mid_step2}, we have
\begin{equation*}
\frac{1}{T}\sum_{t=1}^T \left\{\widehat{\kappa}_{t}(\boldsymbol{x}_t) -\kappa_{t}(\boldsymbol{x}_t)\right\}\left[\frac{\mathbb{I}\{a_t=\widehat{\pi}_t(\boldsymbol{x}_t )\}\Big[ r_t - {\mu}  \{\boldsymbol{x}_t ,\widehat{\pi}_t(\boldsymbol{x}_t )\}\Big] }{\{1-\widehat{\kappa}_{t}(\boldsymbol{x}_t)\}\{1-\kappa_{t}(\boldsymbol{x}_t)\}}\right]  =\mathcal{O}_p(T^{-1}) =  o_p(T^{-1/2}).
\end{equation*}

Then we consider the second term in \eqref{mid_step2}, where
\begin{eqnarray*}
&&\frac{1}{T}\sum_{t=1}^T \left\{\widehat{\kappa}_{t}(\boldsymbol{x}_t) -\kappa_{t}(\boldsymbol{x}_t)\right\}\left[\frac{\mathbb{I}\{a_t=\widehat{\pi}_t(\boldsymbol{x}_t )\}\Big[  {\mu}  \{\boldsymbol{x}_t ,\widehat{\pi}_t(\boldsymbol{x}_t )\} -\widehat{\mu}_{t-1} \{\boldsymbol{x}_t ,\widehat{\pi}_t(\boldsymbol{x}_t )\}\Big]}{\{1-\widehat{\kappa}_{t}(\boldsymbol{x}_t)\}\{1-\kappa_{t}(\boldsymbol{x}_t)\}}\right] \\
\leq &&\frac{1}{T}\sum_{t=1}^T B_\kappa\left|\widehat{\kappa}_{t}(\boldsymbol{x}_t) -\kappa_{t}(\boldsymbol{x}_t)\right| \left|  {\mu}  \{\boldsymbol{x}_t ,\widehat{\pi}_t(\boldsymbol{x}_t )\} -\widehat{\mu}_{t-1} \{\boldsymbol{x}_t ,\widehat{\pi}_t(\boldsymbol{x}_t )\}\right|,
\end{eqnarray*}
for some $B_\kappa$ as the bound of $\mathbb{I}\{a_t=\widehat{\pi}_t(\boldsymbol{x}_t )\}/[{\{1-\widehat{\kappa}_{t}(\boldsymbol{x}_t)\}\{1-\kappa_{t}(\boldsymbol{x}_t)\}}]$. 
By Cauchy-Schwartz inequality, we have the above term further bounded by 
\begin{eqnarray*}
&&\frac{1}{T}\sum_{t=1}^T B_\kappa\left|\widehat{\kappa}_{t}(\boldsymbol{x}_t) -\kappa_{t}(\boldsymbol{x}_t)\right| \left|  {\mu}  \{\boldsymbol{x}_t ,\widehat{\pi}_t(\boldsymbol{x}_t )\} -\widehat{\mu}_{t-1} \{\boldsymbol{x}_t ,\widehat{\pi}_t(\boldsymbol{x}_t )\}\right|\\
\leq && B_\kappa\sqrt{\frac{1}{T}\sum_{t=1}^T \left|\widehat{\kappa}_{t}(\boldsymbol{x}_t) -\kappa_{t}(\boldsymbol{x}_t)\right|^2 
\frac{1}{T}\sum_{t=1}^T \left|  {\mu}  \{\boldsymbol{x}_t ,\widehat{\pi}_t(\boldsymbol{x}_t )\} -\widehat{\mu}_{t-1} \{\boldsymbol{x}_t ,\widehat{\pi}_t(\boldsymbol{x}_t )\}\right|^2}.
\end{eqnarray*}
Given  Assumption  4.4,
we have the above bounded by $o_p(T^{-1/2})$, and thus the second term in \eqref{mid_step2} is $o_p(T^{-1/2})$.

Therefore, we have $\widehat{V}_T =  \widetilde{\phi}_T  +o_p(T^{-1/2})$ hold. 

Then, we focus on proving 
\begin{eqnarray}\label{mid_step3}
\widetilde{\phi}_T -\widetilde{V}_T = \frac{1}{T}\sum_{t=1}^T\left[\frac{\mathbb{I}\{a_t=\widehat{\pi}_t(\boldsymbol{x}_t )\} }{ 1-\kappa_{t}(\boldsymbol{x}_t)} -1\right]\Big[   {\mu} \{\boldsymbol{x}_t ,\widehat{\pi}_t(\boldsymbol{x}_t )\} - \widehat{\mu}_{t-1} \{\boldsymbol{x}_t ,\widehat{\pi}_t(\boldsymbol{x}_t )\}\Big],
\end{eqnarray}
is $o_p(T^{-1/2})$. Specifically, since
\begin{equation*}
    {\mu} \{\boldsymbol{x}_t ,\widehat{\pi}_t(\boldsymbol{x}_t )\} - \widehat{\mu}_{t-1} \{\boldsymbol{x}_t ,\widehat{\pi}_t(\boldsymbol{x}_t )\} = \mathcal{O}_p(t^{-1/2}),
\end{equation*}
similar as the proof of first term in \eqref{mid_step2}, we could prove that  $\widetilde{\phi}_T -\widetilde{V}_T = o_p(T^{-1/2})$.
Thus $\widehat{V}_T =  \widetilde{V}_T  +o_p(T^{-1/2})$ holds.  The first step is thus completed. 

\noindent\textbf{Step 2:} We next focus on  proving $\widetilde{V}_T =  \overline{V}_T +o_p(T^{-1/2})$. By definition of $\widetilde{V}_T$ and $\overline{V}_T$, we have
 \begin{eqnarray}\label{thm4_step2_main_res}
\sqrt{T}(\widetilde{V}_T - \overline{V}_T)= && \underbrace{\frac{1}{\sqrt{T}}\sum_{t=1}^T\left[\frac{\mathbb{I}\{a_t=\widehat{\pi}_t(\boldsymbol{x}_t )\} }{1-\kappa_{t}(\boldsymbol{x}_t)} -1\right] \Big[ {\mu} \{\boldsymbol{x}_t ,{\pi}^*(\boldsymbol{x}_t )\}-{\mu} \{\boldsymbol{x}_t ,\widehat{\pi}_t(\boldsymbol{x}_t )\}\Big]}_{\eta_5}\\\nonumber &&+\underbrace{\frac{1}{\sqrt{T}}\sum_{t=1}^T\left[\frac{\mathbb{I}\{a_t=\widehat{\pi}_t(\boldsymbol{x}_t )\} }{1-\kappa_{t}(\boldsymbol{x}_t)}- \frac{\mathbb{I}\{a_t={\pi}^*(\boldsymbol{x}_t )\} }{ \prob\{a_t={\pi}^*(\boldsymbol{x}_t)\}} \right]\Big[r_t- {\mu} \{\boldsymbol{x}_t ,\pi^*_t(\boldsymbol{x}_t )\}\Big]}_{\eta_6}.
\end{eqnarray} 
We first show $ \eta_5 =o_p(1)$. Since $\kappa_{t}(\boldsymbol{x}_t) \leq C_2 <1 $ for sure for some constant $0< C_2 <1$ (by definition of a valid bandit algorithm as results for Theorem 1), it suffices to show 
\begin{equation*}
\psi_5 = { {T}^{-1/2} } \sum_{t=1}^T\left|\mu\{\boldsymbol{x}_t, \widehat{\pi}_t(\boldsymbol{x}_t ) \} - \mu\{\boldsymbol{x}_t, \pi^*\} \right|= o_p(1),
\end{equation*}
which is the direct conclusion of Lemma \ref{lemma2}.

Next, we show $ \eta_6 =o_p(1)$. Firstly we can express $\eta_6$ as 
\begin{equation*}
\begin{aligned}
\eta_6  =&\frac{1}{\sqrt{T}}\sum_{t=1}^T\left[\frac{\mathbb{I}\{a_t=\widehat{\pi}_t(\boldsymbol{x}_t )\} }{1-\kappa_{t}(\boldsymbol{x}_t)}- \frac{\mathbb{I}\{a_t={\pi}^*(\boldsymbol{x}_t )\} }{ \prob\{a_t={\pi}^*(\boldsymbol{x}_t)\}} \right]\Big[r_t- {\mu} \{\boldsymbol{x}_t ,\pi^*_t(\boldsymbol{x}_t )\}\Big]\\
=&\frac{1}{\sqrt{T}}\sum_{t=1}^T\left[\frac{\mathbb{I}\{a_t=\widehat{\pi}_t(\boldsymbol{x}_t )\} }{1-\kappa_{t}(\boldsymbol{x}_t)}- \frac{\mathbb{I}\{a_t={\pi}^*(\boldsymbol{x}_t )\} }{ \prob\{a_t={\pi}^*(\boldsymbol{x}_t)\}} \right] e_t \\
=& \frac{1}{\sqrt{T}}\sum_{t=1}^T\left[\frac{1 }{1-\kappa_{t}(\boldsymbol{x}_t)}- \frac{\mathbb{I}\{a_t={\pi}^*(\boldsymbol{x}_t )\} }{ \prob\{a_t={\pi}^*(\boldsymbol{x}_t)\}} \right]e_t  \mathbb{I}\{a_t=\widehat{\pi}_t(\boldsymbol{x}_t )\} \\
& - \frac{1}{\sqrt{T}}\sum_{t=1}^T\frac{\mathbb{I}\{a_t={\pi}^*(\boldsymbol{x}_t )\} }{ \prob\{a_t={\pi}^*(\boldsymbol{x}_t)\}} e_t \mathbb{I}\{a_t\neq \widehat{\pi}_t(\boldsymbol{x}_t )\}\\
 =& \underbrace{\frac{1}{\sqrt{T}}\sum_{t=1}^T\left[\frac{1 }{1-\kappa_{t}(\boldsymbol{x}_t)}- \frac{\mathbb{I}\{a_t={\pi}^*(\boldsymbol{x}_t )\} }{ \prob\{a_t={\pi}^*(\boldsymbol{x}_t)\}} \right]e_t \mathbb{I}\{a_t=\widehat{\pi}_t(\boldsymbol{x}_t )\}}_{\psi_6} \\
& -  \underbrace{\frac{1}{\sqrt{T}}\sum_{t=1}^T\frac{1 }{\prob\{a_t={\pi}^*(\boldsymbol{x}_t)\}} e_t \mathbb{I}\{a_t\neq \widehat{\pi}_t(\boldsymbol{x}_t )\}\mathbb{I}\{a_t={\pi}^*(\boldsymbol{x}_t )\}}_{\psi_7}. \\
\end{aligned}
\end{equation*}

Note that 
\begin{equation*}
\begin{aligned}
\prob\{a_t={\pi}^*(\boldsymbol{x}_t)\} & \geq \prob\{a_t=\widehat{\pi}_{t}(\boldsymbol{x}_t), \widehat{\pi}_{t}(\boldsymbol{x}_t)={\pi}^*(\boldsymbol{x}_t)\}\\
& \geq \prob\{a_t=\widehat{\pi}_{t}(\boldsymbol{x}_t)\}+\prob\{ \widehat{\pi}_{t}(\boldsymbol{x}_t)={\pi}^*(\boldsymbol{x}_t)\}-1\\
& = \prob\{ \widehat{\pi}_{t}(\boldsymbol{x}_t)= {\pi}^*(\boldsymbol{x}_t)\} - \kappa_{t}(\boldsymbol{x}_t),
\end{aligned}
\end{equation*}
where $\kappa_{t}(\boldsymbol{x}_t) = o_p(1)$ by Theorem 1 and $\prob\{ \widehat{\pi}_{t}(\boldsymbol{x}_t)= {\pi}^*(\boldsymbol{x}_t)\} \geq 1-ct^{- \alpha \gamma} $ by Lemma \ref{lemma} as $t \rightarrow \infty$. Thus we have for large enough t, 
\begin{equation} \label{eq:ptbound}
    \prob\{a_t={\pi}^*(\boldsymbol{x}_t)\} > C_1,
\end{equation}
for some constant $C_1>0$.

Then we focus on proving $ \psi_6 = o_p(1)$  here. 
Define a class of function 
\begin{eqnarray*}
\mathcal{F}_\kappa(\boldsymbol{x}, a, r) =  \Bigg\{ \left[\frac{1 }{1-\kappa(\boldsymbol{x})}- \frac{\mathbb{I}\{a={\pi}^*(\boldsymbol{x})\}}{\prob\{a={\pi}^*(\boldsymbol{x})\}} \right]e_t \mathbb{I}\{a=\pi(\boldsymbol{x} )\}: \kappa_{t} \in \Lambda, \pi(\cdot)  \in \Pi  \Bigg\},
\end{eqnarray*} 
where $\Pi$ and $\Lambda$ are two classes of functions that maps context $\boldsymbol{x}\in \mathcal{X}$ to a probability.
\\
Define the supremum of the empirical process indexed by $ \mathcal{F}_\kappa $ as
\begin{eqnarray} \label{eq:defG2}
||\mathbb{G}_n||_{\mathcal{F}} \equiv &&\sup_{\pi \in \Pi} \frac{1}{\sqrt{T}}\sum_{t=1}^{T} \left[  \mathcal{F}_\kappa (\boldsymbol{x}_t, a_t, r_t) - \mathbb{E}\{  \mathcal{F}_\kappa (\boldsymbol{x}_t, a_t, r_t) | \mathcal{H}_{t-1}\} \right].
\end{eqnarray}
Firstly we notice that 
\begin{equation*}
\mathbb{E}\{  \mathcal{F}_\kappa (\boldsymbol{x}_t, a_t, r_t) | \mathcal{H}_{t-1}\}  = \Mean \left( \left[\frac{1 }{1-\kappa_{t}(\boldsymbol{x}_t)}- \frac{\mathbb{I}\{a_t={\pi}^*(\boldsymbol{x}_t )\} }{ \prob\{a_t={\pi}^*(\boldsymbol{x}_t)\}} \right]e_t \mathbb{I}\{a=\pi_t(\boldsymbol{x} )\} \mid \mathcal{H}_{t-1} \right) = 0,
\end{equation*}
since $\Mean \left(e_t|\{a_t, \mathcal{H}_{t-1}\right) = 0$.
\\
Secondly since $\kappa_{t}(\boldsymbol{x}_t) \leq C_2 <1 $ for sure for some constant $0< C_2 <1$ (by definition of a valid bandit algorithm an results for Theorem 1), by the Triangle inequality, we have
\begin{equation*}
  \left|\frac{1 }{1-\kappa_{t}(\boldsymbol{x}_t)}- \frac{\mathbb{I}\{a_t={\pi}^*(\boldsymbol{x}_t )\} }{ \prob\{a_t={\pi}^*(\boldsymbol{x}_t)\}} \right|  \leq    \left|\frac{1 }{1-\kappa_{t}(\boldsymbol{x}_t)}-\right|+\left| \frac{\mathbb{I}\{a_t={\pi}^*(\boldsymbol{x}_t )\} }{ \prob\{a_t={\pi}^*(\boldsymbol{x}_t)\}} \right| \leq \frac{1 }{1-C_2} + \frac{1}{C_1} \triangleq C.
\end{equation*}
Therefore, we have 
\begin{equation*}
\begin{aligned}
& \sum_{t=1}^T \Mean \left\{ \left(\frac{1}{\sqrt{T}}\left[\frac{1 }{1-\kappa_{t}(\boldsymbol{x}_t)}- \frac{\mathbb{I}\{a_t={\pi}^*(\boldsymbol{x}_t )\} }{ \prob\{a_t={\pi}^*(\boldsymbol{x}_t)\}} \right]e_t \mathbb{I}\{a_t=\widehat{\pi}_t(\boldsymbol{x}_t )\}\right)^2 \mid \mathcal{H}_{t-1} \right\} \\
&\leq \sum_{t=1}^T  \Mean\left\{ \left(\frac{1}{\sqrt{T}}\left[\frac{1 }{1-\kappa_{t}(\boldsymbol{x}_t)}- \frac{\mathbb{I}\{a_t={\pi}^*(\boldsymbol{x}_t )\} }{ \prob\{a_t={\pi}^*(\boldsymbol{x}_t)\}} \right]e_t \mathbb{I}\{a_t=\widehat{\pi}_t(\boldsymbol{x}_t )\}\right)^2 \mid \mathcal{H}_{t-1} \right\} \\
&\leq \sum_{t=1}^T \frac{C^2}{T } \Mean \left\{ e_t^2 \mid \mathcal{H}_{t-1} \right\} \leq \sum_{t=1}^T \frac{C^2 }{ T } \max\{\sigma_0^2,\sigma_1^2\} = C^2 \max\{\sigma_0^2,\sigma_1^2\} < \infty.
\end{aligned}
\end{equation*}
Therefore, we have 
\begin{equation*}
d_{1}(f) \equiv \left\|\mathbb{E}\left(\mid  \mathcal{F}_\kappa (\boldsymbol{x}_1, a_1, r_1) \| \mathcal{H}_{0}\right)\right\|_{\infty}  <\infty,
\end{equation*}
and
\begin{equation*}
d_{2}(f)\equiv \left\|\mathbb{E}\left(\left(\mathcal{F}_\kappa (\boldsymbol{x}_1, a_1, r_1)\right)^{2} \mid \mathcal{H}_{0}\right)\right\|_{\infty}^{1 / 2}<\infty.
\end{equation*}
It follows from  the maximal inequality developed in Section 4.2 of \cite{dedecker2002maximal} that there exists some constant $K\ge 1$ such that
\begin{eqnarray*}
\Mean \Big[||\mathbb{G}_n||_{\mathcal{F}} \Big] \lesssim   K\left( d_{2}(f)+\frac{1}{\sqrt{T}}\left\|\max_{1 \leq t \leq T}\left| \mathcal{F}_\kappa (\boldsymbol{x}_t, a_t, r_t) -\mathbb{E}\left( \mathcal{F}_\kappa (\boldsymbol{x}_t, a_t, r_t)  \mid \mathcal{H}_{t-1}\right)\right|\right\|\right)
\end{eqnarray*}
The above right-hand-side is upper
 bounded by 
\begin{eqnarray*}
\mathcal{O}(1)\sqrt{T^{-1/2}},
\end{eqnarray*}
 where $\mathcal{O}(1)$ denotes some universal constant. Hence, we have
\begin{eqnarray} \label{eq:resultG2}
\Mean \Big[||\mathbb{G}_n||_{\mathcal{F}} \Big] = \mathcal{O}_p(T^{-1/2}).
\end{eqnarray}
Combined with Equation \eqref{eq:defG2}, we have
\begin{equation*}
\frac{1}{\sqrt{T}}\sum_{t=1}^{T}  \mathcal{F}_\kappa (\boldsymbol{x}_t, a_t, r_t) = \mathcal{O}_p(T^{-1/2}). 
\end{equation*}
and $ \psi_6=o_p(1)$.

Next, for $ \psi_7$, by triangle inequality, we have 
\begin{equation*}
\begin{aligned}
|\psi_7|  & =  \left|\frac{1}{\sqrt{T}}\sum_{t=1}^T\frac{1 }{\prob\{a_t={\pi}^*(\boldsymbol{x}_t)\}} e_t \mathbb{I}\{a_t\neq \widehat{\pi}_t(\boldsymbol{x}_t )\}\mathbb{I}\{a_t={\pi}^*(\boldsymbol{x}_t )\}\right|\\
& \leq \frac{1}{\sqrt{T}}\sum_{t=1}^T \left|\frac{1 }{\prob\{a_t={\pi}^*(\boldsymbol{x}_t)\}} e_t \mathbb{I}\{a_t\neq \widehat{\pi}_t(\boldsymbol{x}_t )\}\mathbb{I}\{a_t={\pi}^*(\boldsymbol{x}_t )\}\right|\\
& \leq \frac{1}{\sqrt{T}}\sum_{t=1}^T \left|\frac{1}{ \prob\{a_t={\pi}^*(\boldsymbol{x}_t)\}}\right|e_t \left| \mathbb{I}\{a_t\neq \widehat{\pi}_t(\boldsymbol{x}_t )\}\mathbb{I}\{a_t={\pi}^*(\boldsymbol{x}_t )\}\right|\\
& \leq \frac{1}{\sqrt{T}}\sum_{t=1}^T \frac{1}{C_1}e_t.
\end{aligned}
\end{equation*}
Notice that $\frac{1}{\sqrt{T}}\sum_{t=1}^T  e_t= o_p(1)$, we have
\begin{equation*}
 |\psi_7|  \leq    = \frac{1}{C_1} \frac{1}{\sqrt{T}}\sum_{t=1}^T  e_t  = o_p(1).
\end{equation*}
Therefore, $\eta_6 = \psi_6 + \psi_7 = o_p(1) $.
 Hence,  $\widetilde{V}_T =  \overline{V}_T +o_p(T^{-1/2})$ hold.

\smallskip \noindent \textbf{Step 3:} Then, to show the asymptotic normality of the proposed value estimator under DREAM, based on the above two steps, it is sufficient to show 
\begin{equation}\label{thm4_step3_main_res}
\sqrt{T}(\widehat{V}_T -V^*)= \sqrt{T}(\overline{V}_T -V^*)+o_p(1)  \stackrel{D}{\longrightarrow}  \mathcal{N}\left(0,  \sigma_{DR}^2 \right),
\end{equation} 
as $T \rightarrow \infty$, using Martingale Central Limit Theorem. By $r_t = {\mu} \{\boldsymbol{x}_t ,a_t ) +e_t$, we define
\begin{eqnarray}\label{eqn:Zt}
\xi_t&&=  \frac{\mathbb{I}\{a_t={\pi}^*(\boldsymbol{x}_t )\} }{ \prob\{a_t={\pi}^*(\boldsymbol{x}_t)\}} \Big[ r_t - {\mu} \{\boldsymbol{x}_t ,{\pi}^*(\boldsymbol{x}_t )\}\Big] +  {\mu} \{\boldsymbol{x}_t ,{\pi}^*(\boldsymbol{x}_t ) \}\\\nonumber
&&=  \underbrace{ \frac{\mathbb{I}\{a_t={\pi}^*(\boldsymbol{x}_t )\} }{ \prob\{a_t={\pi}^*(\boldsymbol{x}_t)\}} e_t}_{Z_t} + \underbrace{  {\mu} \{\boldsymbol{x}_t ,{\pi}^*(\boldsymbol{x}_t ) \}-V^*}_{W_t}.  
\end{eqnarray} 
By \eqref{eqn:Zt}, we have
\begin{equation*}
\begin{aligned}
\Mean \{Z_t\mid \mathcal{H}_{t-1} \} & =\Mean \{\Mean \left[ Z_t\mid a_t \right] \mid \mathcal{H}_{t-1} \} =\Mean \left\{\Mean \left[ \frac{\mathbb{I}\{a_t={\pi}^*(\boldsymbol{x}_t )\} }{ \prob\{a_t={\pi}^*(\boldsymbol{x}_t)\}} e_t\mid a_t,\boldsymbol{x}_t  \right] \mid \mathcal{H}_{t-1} \right\}\\
& =\Mean \left\{ \frac{\mathbb{I}\{a_t={\pi}^*(\boldsymbol{x}_t )\} }{ \prob\{a_t={\pi}^*(\boldsymbol{x}_t)\}} \Mean \left[e_t\mid a_t,\boldsymbol{x}_t  \right] \mid \mathcal{H}_{t-1} \right\}.
\end{aligned}
\end{equation*}
Since $e_t$ is independent of $\mathcal{H}_{i-1}$ and $\boldsymbol{x}_i$ given $a_t$,  and $ \Mean  \left\{ e_t \mid a_t \right\}  = 0$,  we have
\begin{equation*}
\Mean \{Z_t\mid \mathcal{H}_{t-1} \}  =  \Mean  \left\{   \frac{\mathbb{I}\{a_t={\pi}^*(\boldsymbol{x}_t )\} }{ \prob\{a_t={\pi}^*(\boldsymbol{x}_t)\}}  \Mean  \left\{ e_t \mid a_t \right\} \mid \mathcal{H}_{t-1} \right\}   = 0.
\end{equation*}
Notice that $ \Mean[{\mu} \{\boldsymbol{x}_t ,{\pi}^*(\boldsymbol{x}_t ) \}]=V^* $ by the definition, we have 
\begin{equation*}
 \Mean \{W_t\mid \mathcal{H}_{t-1} \}=  \Mean  \left\{    {\mu} \{\boldsymbol{x}_t ,{\pi}^*(\boldsymbol{x}_t ) \}-V^* \right\}  = 0.
\end{equation*}
Thus, we have $\Mean \{\xi_t\mid \mathcal{H}_{t-1} \}  = 0$,  which implies that $\{Z_t\}_{t=1}^T$, $\{W_t\}_{t=1}^T$  and $\{\xi_t\}_{t=1}^T$ are Martingale difference sequences.  To prove Equation \eqref{thm4_step3_main_res},
it suffices to prove that $(1/ \sqrt{T}) \sum_{t=1}^T \xi_t   \overset{D}{\longrightarrow} \mathcal{N}( 0, \sigma_{DR}^2) $, as $T \rightarrow \infty$, using Martingale Central Limit Theorem. 

Firstly we calculate the conditional variance of $\xi_t$ given $ \mathcal{H}_{t-1}$. Note that
\begin{equation*}
\begin{aligned}
\Mean \left(Z_t^2 \mid \mathcal{H}_{t-1} \right)  & = \Mean \left\{  \left(\frac{\mathbb{I}\{a_t={\pi}^*(\boldsymbol{x}_t )\} }{ \prob\{a_t={\pi}^*(\boldsymbol{x}_t)\}} e_t  \right) ^2 \mid \mathcal{H}_{t-1} \right
\} =  \Mean\left( \frac{\mathbb{I}\{a_t=\pi^*(\boldsymbol{x}_t )\}}{[\prob\{a_t={\pi}^*(\boldsymbol{x}_t)\}]^2}  e_t^2\mid \mathcal{H}_{t-1}  \right) , \\
\end{aligned}
\end{equation*}
and
\begin{equation*}
\begin{aligned}
\Mean \left(Z_t^2 \mid \mathcal{H}_{t-1} \right)  & = \Mean  \left[\Mean \left(\frac{\mathbb{I}\{a_t=\pi^*(\boldsymbol{x}_t )\} }{ [\prob\{a_t={\pi}^*(\boldsymbol{x}_t)\}]^2}  e_t^2\  \mid a_t ,\boldsymbol{x}_t \right)\mid \mathcal{H}_{t-1}  \right] \\
& = \Mean  \left[\frac{\mathbb{I}\{a_t=\pi^*(\boldsymbol{x}_t )\} }{ [\prob\{a_t={\pi}^*(\boldsymbol{x}_t)\}]^2}   \Mean \left(e_t^2\  \mid a_t,\boldsymbol{x}_t  \right)\mid \mathcal{H}_{t-1}  \right].
\end{aligned}
\end{equation*}
Since $e_t$ is independent of $\mathcal{H}_{i-1}$ and $\boldsymbol{x}_i$ given $a_t$, we have
\begin{equation*}
\begin{aligned}
 \frac{1}{T}\sum_{t=1}^{T}\Mean \left(Z_t^2 \mid \mathcal{H}_{t-1} \right) & = \frac{1}{T}\sum_{t=1}^{T}\Mean  \left[\frac{\mathbb{I}\{a_t=\pi^*(\boldsymbol{x}_t )\} }{ [\prob\{a_t={\pi}^*(\boldsymbol{x}_t)\}]^2}   \Mean \left(e_t^2\  \mid a_t  \right) \mid \mathcal{H}_{t-1}  \right] \\
 & =\frac{1}{T}\sum_{t=1}^{T}\Mean\left( \frac{\mathbb{I}\{a_t=\pi^*(\boldsymbol{x}_t )\} }{ [\prob\{a_t={\pi}^*(\boldsymbol{x}_t)\}]^2}  \sigma_{a_t}^2  \mid \mathcal{H}_{t-1}  \right).\\
 & = \frac{1}{T}\sum_{t=1}^{T}\Mean \left[\frac{1 }{[\prob\{a_t={\pi}^*(\boldsymbol{x}_t)\}]^2} \sigma_{\pi^*(\boldsymbol{x}_t )}^2\Mean\left( \mathbb{I}\{a_t=\pi^*(\boldsymbol{x}_t )\}   \mid \boldsymbol{x}_t, \mathcal{H}_{t-1}   \right)  \right].
\end{aligned}
\end{equation*}
By the definition of Equation \eqref{eq:nudef}, 
\begin{equation*}
\nu_{i}\left(\boldsymbol{x}_{i},\mathcal{H}_{i-1}\right)
 \equiv \operatorname{Pr}\left\{a_{i} \neq \pi^{*}( \boldsymbol{x}_i) \right |\boldsymbol{x}_i,\mathcal{H}_{i-1} \}=\mathbb{E}\left[\mathbb{I}\left\{a_{i} \neq \pi^{*}( \boldsymbol{x}_i) \right\}|\boldsymbol{x}_i,\mathcal{H}_{i-1}\right],
\end{equation*}we have
\begin{equation*}
\begin{aligned}
\frac{1}{T}\sum_{t=1}^{T}\Mean \left(Z_t^2 \mid \mathcal{H}_{t-1} \right) & = \frac{1}{T}\sum_{t=1}^{T}\Mean \left[\frac{1 }{[\prob\{a_t={\pi}^*(\boldsymbol{x}_t)\}]^2}  \sigma_{\pi^*(\boldsymbol{x}_t )}^2\left\{ 1- \nu_{t}\left(\boldsymbol{x},\mathcal{H}_{t-1}\right) \right\} \right]\\
& = \frac{1}{T}\sum_{t=1}^{T}\int \frac{\left\{ 1- \nu_{t}\left(\boldsymbol{x},\mathcal{H}_{t-1} \right) \right\} }{[\prob\{a_t={\pi}^*(\boldsymbol{x})\}]^2} \sigma_{\pi^*(\boldsymbol{x})}^2d P_{\mathcal{X}} \\
& = \int \left[\frac{1}{T}\sum_{t=1}^{T}\frac{\left\{ 1- \nu_{t}\left(\boldsymbol{x},\mathcal{H}_{t-1} \right) \right\} }{[\prob\{a_t={\pi}^*(\boldsymbol{x})\}]^2}\right] \sigma_{\pi^*(\boldsymbol{x} )}^2d P_{\mathcal{X}}.
\end{aligned}
\end{equation*}
Similar as before, since $\lim_{i\rightarrow \infty}\prob\{a_i \neq \pi^{*}(\boldsymbol{x}) \} = \kappa_{\infty}(\boldsymbol{x})$, we have for any $\epsilon >0 $, there exist a constant $T_0 >0$ such that $\left|\prob\{a_i \neq \pi^{*}(\boldsymbol{x}) \} - \kappa_{\infty}(\boldsymbol{x})\right| < \epsilon$ for all $i \geq T_0$. \\
We firstly consider the expectation of $(1/T)\sum_{t=1}^{T}{\left\{ 1- \nu_{t}\left(\boldsymbol{x},\mathcal{H}_{t-1} \right) \right\} / [\prob\{a_t={\pi}^*(\boldsymbol{x})\}]^2}$. Note that $\prob\{a_t={\pi}^*(\boldsymbol{x})\}$ is not conditional on $\mathcal{H}_{t-1}$, thus we have
\begin{equation*}
\begin{aligned}
\Mean \left[ \frac{1}{T}\sum_{t=1}^{T} \frac{\left\{ 1- \nu_{t}\left(\boldsymbol{x},\mathcal{H}_{t-1} \right) \right\} }{[\prob\{a_t={\pi}^*(\boldsymbol{x})\}]^2} \right] & = \frac{1}{T}\sum_{t=1}^{T} \frac{\Mean \left\{ 1- \nu_{t}\left(\boldsymbol{x},\mathcal{H}_{t-1} \right) \right\} }{[\prob\{a_t={\pi}^*(\boldsymbol{x})\}]^2}  = \frac{1}{T}\sum_{t=1}^{T}\frac{\prob\{a_t={\pi}^*(\boldsymbol{x})\} }{[\prob\{a_t={\pi}^*(\boldsymbol{x})\}]^2}\\
&=\frac{1}{T}\sum_{t=1}^{T}\frac{1 }{\prob\{a_t={\pi}^*(\boldsymbol{x})\}}.
\end{aligned}
\end{equation*}
Therefore by the triangle inequality we have
\begin{equation*}
\begin{aligned}
& ~~~\left|\Mean \left[ \frac{1}{T}\sum_{t=1}^{T} \frac{\left\{ 1- \nu_{t}\left(\boldsymbol{x},\mathcal{H}_{t-1} \right) \right\} }{[\prob\{a_t={\pi}^*(\boldsymbol{x})\}]^2}\right] - \frac{1}{1-\kappa_{\infty}(\boldsymbol{x})} \right| 
= \left|\frac{1}{T}\sum_{t=1}^{T}\frac{1}{\prob\{a_t={\pi}^*(\boldsymbol{x})\}} - \frac{1}{1-\kappa_{\infty}(\boldsymbol{x})} \right|\\
& = \left|\frac{1}{T}\sum_{t=1}^{T} \left[\frac{1}{\prob\{a_t={\pi}^*(\boldsymbol{x})\}} - \frac{1}{1-\kappa_{\infty}(\boldsymbol{x})} \right]\right|
= \left|\frac{1}{T}\sum_{t=1}^{T} \frac{1-\kappa_{\infty}(\boldsymbol{x}) -  \prob\{a_t={\pi}^*(\boldsymbol{x})\}}{\prob\{a_t={\pi}^*(\boldsymbol{x})\}\left\{1- \kappa_{\infty}(\boldsymbol{x}) \right\}}\right| \\
& = \left|\frac{1}{T}\sum_{t=1}^{T} \frac{\prob\{a_t \neq {\pi}^*(\boldsymbol{x})-\kappa_{\infty}(\boldsymbol{x}) \}}{\prob\{a_t={\pi}^*(\boldsymbol{x})\}\left\{1- \kappa_{\infty}(\boldsymbol{x}) \right\}}\right|  \leq  \frac{1}{T}\sum_{t=1}^{T} \frac{\left|\prob\{a_t \neq {\pi}^*(\boldsymbol{x})-\kappa_{\infty}(\boldsymbol{x}) \}\right|}{\prob\{a_t={\pi}^*(\boldsymbol{x})\}\left\{1- \kappa_{\infty}(\boldsymbol{x}) \right\}}\\
& = \frac{1}{T}\sum_{t=1}^{T_0} \frac{\left|\prob\{a_t \neq {\pi}^*(\boldsymbol{x})-\kappa_{\infty}(\boldsymbol{x}) \}\right|}{\prob\{a_t={\pi}^*(\boldsymbol{x})\}\left\{1- \kappa_{\infty}(\boldsymbol{x}) \right\}} 
+ \frac{1}{T}\sum_{t=T_0}^{T}\frac {\left|\prob\{a_t \neq {\pi}^*(\boldsymbol{x})-\kappa_{\infty}(\boldsymbol{x}) \}\right| }{\prob\{a_t={\pi}^*(\boldsymbol{x})\}\left\{1- \kappa_{\infty}(\boldsymbol{x}) \right\}} \\
& < \frac{1}{T}\sum_{t=1}^{T_0} \frac{\left| \prob\{a_t \neq {\pi}^*(\boldsymbol{x})\}\right|+ \left|\kappa_{\infty}(\boldsymbol{x})\right|}{\prob\{a_t={\pi}^*(\boldsymbol{x})\}\left\{1- \kappa_{\infty}(\boldsymbol{x}) \right\}} 
+ \frac{1}{T}\sum_{t=T_0}^{T} \frac{\epsilon}{\prob\{a_t={\pi}^*(\boldsymbol{x})\}\left\{1- \kappa_{\infty}(\boldsymbol{x}) \right\}}. \\
\end{aligned}
\end{equation*}
Since the above equation holds for any $\epsilon >0$, we have 
\begin{equation*}
\begin{aligned}
\left|\Mean \left[ \frac{1}{T}\sum_{t=1}^{T}\frac{1-\nu_{t}\left(\boldsymbol{x},\mathcal{H}_{t-1}\right)}{[\prob\{a_t={\pi}^*(\boldsymbol{x})\}]^2}\right] - \frac{1}{1-\kappa_{\infty}(\boldsymbol{x})} \right| & \leq  \frac{1}{T}\sum_{t=1}^{T_0} \frac{\left| \prob\{a_t = {\pi}^*(\boldsymbol{x})\}\right|+ \left|\kappa_{\infty}(\boldsymbol{x})\right| }{\prob\{a_t={\pi}^*(\boldsymbol{x})\} \left\{1- \kappa_{\infty}(\boldsymbol{x}) \right\}} \\
& \leq  \frac{1}{T}\sum_{t=1}^{T_0} \frac{2 }{\prob\{a_t={\pi}^*(\boldsymbol{x})\} \left\{1- \kappa_{\infty}(\boldsymbol{x}) \right\}} .
\end{aligned}
\end{equation*}
Since $\kappa_{\infty}(\boldsymbol{x}) \leq C_2 <1 $ for sure for some constant $0< C_2 <1$ (by definition of a valid bandit algorithm an results for Theorem 1), and by Equation \eqref{eq:ptbound}, we have $\prob\{a_t={\pi}^*(\boldsymbol{x})\} > C_1$ for some constant $C_1>0$, therefore we have
\begin{equation*}
\begin{aligned}
\left|\Mean \left[ \frac{1}{T}\sum_{t=1}^{T}\frac{\left\{ 1- \nu_{t}\left(\boldsymbol{x},\mathcal{H}_{t-1} \right) \right\} }{[\prob\{a_t={\pi}^*(\boldsymbol{x})\}]^2}\right] - \frac{1}{1-\kappa_{\infty}(\boldsymbol{x})} \right| \leq   \frac{1}{T}\sum_{t=1}^{T_0} \frac{2 }{C_1 \left( 1- C_2 \right)}  = \frac{2 T_0}{C_1 \left( 1- C_2 \right)T} \rightarrow 0,
\end{aligned}
\end{equation*}
as $T \rightarrow \infty$. \\
Then we consider the variance of $(1/T)\sum_{t=1}^{T}{\left\{ 1- \nu_{t}\left(\boldsymbol{x},\mathcal{H}_{t-1} \right) \right\} / [\prob\{a_t={\pi}^*(\boldsymbol{x})\}]^2}$ over different histories. By Lemma \ref{lemma:bounded}, we have
\begin{equation*}
\begin{aligned}
 \operatorname{Var}\left[\nu_{t}\left(\boldsymbol{x},\mathcal{H}_{t-1}\right)\right]\leq \prob\{a_t={\pi}^*(\boldsymbol{x})\} \left[ 1-\prob\{a_t={\pi}^*(\boldsymbol{x})\}\right].
\end{aligned}
\end{equation*}
Therefore, we have 
\begin{equation*}
\begin{aligned}
& \Var \left[ \frac{1}{T}\sum_{t=1}^{T}\frac{\left\{ 1- \nu_{t}\left(\boldsymbol{x},\mathcal{H}_{t-1} \right) \right\} }{[\prob\{a_t={\pi}^*(\boldsymbol{x})\}]^2}\right]   = \frac{1}{T^2}\Var \left[ \sum_{t=1}^{T} \frac{\left\{ 1- \nu_{t}\left(\boldsymbol{x},\mathcal{H}_{t-1} \right) \right\} }{[\prob\{a_t={\pi}^*(\boldsymbol{x})\}]^2}\right]\\
  \leq & \frac{1}{T}\sum_{t=1}^{T}\Var \left[ \frac{\left\{ 1- \nu_{t}\left(\boldsymbol{x},\mathcal{H}_{t-1} \right) \right\}}{[\prob\{a_t={\pi}^*(\boldsymbol{x})\}]^2}\right]= \frac{1}{T}\sum_{t=1}^{T} \frac{\Var \left[\left\{ 1- \nu_{t}\left(\boldsymbol{x},\mathcal{H}_{t-1} \right) \right\}\right] }{[\prob\{a_t={\pi}^*(\boldsymbol{x})\}]^4} = \frac{1}{T}\sum_{t=1}^{T} \frac{\Var \left[\left\{  \nu_{t}\left(\boldsymbol{x},\mathcal{H}_{t-1} \right) \right\}\right] }{[\prob\{a_t={\pi}^*(\boldsymbol{x})\}]^4} \\
 \leq & \frac{1}{T}\sum_{t=1}^{T} \frac{\prob\{a_t={\pi}^*(\boldsymbol{x})\} \left[ 1-\prob\{a_t={\pi}^*(\boldsymbol{x})\}\right] }{[\prob\{a_t={\pi}^*(\boldsymbol{x})\}]^4}  = \frac{1}{T}\sum_{t=1}^{T} \frac{ 1-\prob\{a_t={\pi}^*(\boldsymbol{x})\}}{[\prob\{a_t={\pi}^*(\boldsymbol{x})\}]^3} \\
 \leq & \frac{1}{T}\sum_{t=1}^{T} \frac{ 1-\prob\{a_t={\pi}^*(\boldsymbol{x})\} }{[1-C_2]^3}  =  \frac{1}{[1-C_2]^3}\frac{1}{T}\sum_{t=1}^{T} \prob\{a_t \neq {\pi}^*(\boldsymbol{x})\}.
\end{aligned}
\end{equation*}
Similarly as before, we could proof  
\begin{equation*}
\begin{aligned}
\frac{1}{T}\sum_{t=1}^{T} \prob\{a_t \neq {\pi}^*(\boldsymbol{x})\} \rightarrow \kappa_{\infty}(\boldsymbol{x}),
\end{aligned}
\end{equation*}
which follows
\begin{equation*}
\Var \left[ \frac{1}{T}\sum_{t=1}^{T}\frac{\left\{ 1- \nu_{t}\left(\boldsymbol{x},\mathcal{H}_{t-1} \right) \right\} }{[\prob\{a_t={\pi}^*(\boldsymbol{x})\}]^2}\right] \rightarrow \frac{1}{[1-C_2]^3}\kappa_{\infty}(\boldsymbol{x}).
\end{equation*}
Therefore, as $T$ goes to infinity, we have
\begin{equation*}
\frac{1}{T}\sum_{t=1}^{T}\frac{\left\{ 1- \nu_{t}\left(\boldsymbol{x},\mathcal{H}_{t-1} \right) \right\} }{[\prob\{a_t={\pi}^*(\boldsymbol{x})\}]^2}\to \frac{1}{1-\kappa_{\infty}(\boldsymbol{x})},
\end{equation*}
and 
\begin{equation*}
\begin{aligned}
\frac{1}{T}\sum_{t=1}^{T}\Mean \left(Z_t^2 \mid \mathcal{H}_{t-1} \right)  \rightarrow  \int \frac{1}{1-\kappa_{\infty}(\boldsymbol{x})} \sigma_{\pi^*(\boldsymbol{x} )}^2d P_{\mathcal{X}}.
\end{aligned}
\end{equation*}
Using the same technique of conditioning on $a_t$ and $\boldsymbol{x}_t$, we  have
\begin{equation*}
\begin{aligned}
\Mean \left(W_t Z_t  \mid \mathcal{H}_{t-1} \right)  & =  \Mean \left\{ \left(  {\mu} \{\boldsymbol{x}_t ,{\pi}^*(\boldsymbol{x}_t ) \}-V^*\right)\frac{\mathbb{I}\{a_t={\pi}^*(\boldsymbol{x}_t )\} }{ \prob\{a_t={\pi}^*(\boldsymbol{x}_t)\}} e_t \mid \mathcal{H}_{t-1} \right\} \\
 & =  \Mean \left\{  \Mean  \left[\left(  {\mu} \{\boldsymbol{x}_t ,{\pi}^*(\boldsymbol{x}_t ) \}-V^*\right)\frac{\mathbb{I}\{a_t={\pi}^*(\boldsymbol{x}_t )\} }{ \prob\{a_t={\pi}^*(\boldsymbol{x}_t)\}} e_t \mid a_t,\boldsymbol{x}_t \right] \mid \mathcal{H}_{t-1} \right\} \\
& =  \Mean \left\{ \left(  {\mu} \{\boldsymbol{x}_t ,{\pi}^*(\boldsymbol{x}_t ) \}-V^*\right)\frac{\mathbb{I}\{a_t={\pi}^*(\boldsymbol{x}_t )\} }{ \prob\{a_t={\pi}^*(\boldsymbol{x}_t)\}}\Mean\left( e_t\mid  a_t \right) \mid \mathcal{H}_{t-1} \right\}  = 0.
\end{aligned}
\end{equation*}
Thus, we further have
\begin{equation*}
\Mean \left(\xi_t^2 \mid \mathcal{H}_{t-1} \right)  = \Mean \left\{ \left(Z_t + W_t\right) ^2  \mid \mathcal{H}_{t-1} \right\} = \Mean \left(Z_t^2 \mid \mathcal{H}_{t-1} \right) +\Mean \left(W_t^2 \mid \mathcal{H}_{t-1} \right),
\end{equation*}
and 
\begin{equation*}
\frac{1}{T}\sum_{t=1}^{T}\Mean \left(\xi_t^2 \mid \mathcal{H}_{t-1} \right)  = \int \frac{1}{1-\kappa_{\infty}(\boldsymbol{x})} \sigma_{\pi^*(\boldsymbol{x} )}^2d P_{\mathcal{X}}  + \frac{1}{T}\sum_{t=1}^{T}\Var \left[  {\mu} \{\boldsymbol{x}_t ,{\pi}^*(\boldsymbol{x}_t ) \} \mid \mathcal{H}_{t-1}\right].
\end{equation*}
Therefore as $T$ goes to infinity,  we  have  
\begin{equation*} 
\sum_{t=1}^T\Mean\left \{ \left( \frac{1}{\sqrt{T}} \xi_t \right)^2 \mid \mathcal{H}_{t-1} \right\} \longrightarrow \sigma_{DR}^2,
\end{equation*}
where
\begin{equation}\label{thm4_step3_final_res}
\sigma_{DR}^2 = \int_{\boldsymbol{x}} \frac{\sigma_{1}^2 \mathbb{I}\{\mu(\boldsymbol{x}, 1)> \mu(\boldsymbol{x}, 0) \}+\sigma_{0}^2 \mathbb{I}\{\mu(\boldsymbol{x}, 1)< \mu(\boldsymbol{x}, 0)\}}{1-\kappa_\infty(\boldsymbol{x})} d P_{\mathcal{X}} +\Var \left[  {\mu} \{\boldsymbol{x} ,{\pi}^*(\boldsymbol{x} ) \} \right].
\end{equation}

Then  we check the conditional Lindeberg condition.  For any $h >0$, we have 
\begin{equation*}
\begin{aligned}
&\sum_{t=1}^T\Mean\left \{ \left( \frac{1}{\sqrt{T}} \xi_t \right)^2\mathbb{I}\left\{ \left|\frac{1}{\sqrt{T}} \xi_t\right | >h \right\}\mid \mathcal{H}_{t-1} \right\} = \frac{1}{T}\sum_{t=1}^T\Mean\left \{ \xi_t ^2\mathbb{I}\left\{  \xi_t^2 >Th^2 \right\}\mid \mathcal{H}_{t-1} \right\}. \\
\end{aligned}
\end{equation*}
Since $ \xi_t ^2\mathbb{I}\left\{  \xi_t^2 >Th^2 \right\}$ converges to zero as $T$ goes to infinity and is dominated by $\xi_t^2$ given $\mathcal{H}_{t-1}$.  Therefore, by Dominated Convergence Theorem, we conclude that 
\begin{equation*} \sum_{t=1}^T\Mean\left \{ \left( \frac{1}{\sqrt{T}} \xi_t \right)^2\mathbb{I}\left\{ \left|\frac{1}{\sqrt{T}} \xi_t\right | >h \right\}\mid \mathcal{H}_{t-1} \right\}  \rightarrow 0,  \text{  as } t \rightarrow \infty.
\end{equation*}
Thus the  conditional Lindeberg condition is checked.

Next, recall the derived conditional variance in \eqref{thm4_step3_final_res}. By  Martingale Central Limit Theorem, we have 
\begin{equation*}
(1/ \sqrt{T}) \sum_{t=1}^T \xi_t   \overset{D}{\longrightarrow} \mathcal{N}( 0, \sigma_{DR}^2),
\end{equation*}
as $T \rightarrow \infty$. Hence, we complete the proof of Equation \eqref{thm4_step3_main_res}. 

\smallskip \noindent \textbf{Step 4:} Finally, to show the variance estimator in Equation (5) in the main paper is a consistent estimator of $\sigma_{DR}^2$. Recall that the variance estimator is 
\begin{equation}\label{sigma_dr_hat}
\begin{aligned}
\widehat{\sigma}_{T}^2  & =  \underbrace{\frac{1}{T}\sum_{t=1}^{T} \frac{\widehat{\sigma}_{1,t-1}^2(\boldsymbol{x}_t,1)  \mathbb{I}\{\widehat{\mu}_{t-1}(\boldsymbol{x}_t, 1)> \widehat{\mu}_{t-1}(\boldsymbol{x}_t, 0) \}+\widehat{\sigma}_{0,t-1}^2  \mathbb{I}\{\widehat{\mu}_{t-1}(\boldsymbol{x}_t, 1)< \widehat{\mu}_{t-1}(\boldsymbol{x}_t, 0)\}}{1-\widehat{\kappa}_{t}(\boldsymbol{x}_t)}}_{\widehat{\sigma}_{T,1}^2}\\
 &+   \underbrace{\frac{1}{T}\sum_{t=1}^{T} \left[  \widehat{\mu}_{T} \{\boldsymbol{x}_t ,\widehat{\pi}_{T}(\boldsymbol{x}_t ) \} -  \frac{1}{T}\sum_{t=1}^{T} \widehat{\mu}_{T} \{\boldsymbol{x}_t ,\widehat{\pi}_{T}(\boldsymbol{x}_t ) \}  \right]^2}_{\widehat{\sigma}_{T,2}^2}.
\end{aligned}
\end{equation}
Firstly we proof the first line of the above Equation \eqref{sigma_dr_hat} is a consistent  estimator for 
\begin{equation*}\int_{\boldsymbol{x}} \frac{\sigma_{1}^2 \mathbb{I}\{\mu(\boldsymbol{x}, 1)> \mu(\boldsymbol{x}, 0) \}+\sigma_{0}^2 \mathbb{I}\{\mu(\boldsymbol{x}, 1)< \mu(\boldsymbol{x}, 0)\}}{1-\kappa_\infty(\boldsymbol{x})} d P_{\mathcal{X}}.
\end{equation*}

Recall that we denote $\widehat{\Delta}_{\boldsymbol{x}_t} = \widehat{\mu}_{t-1}(\boldsymbol{x}_t, 1) -\widehat{\mu}_{t-1}(\boldsymbol{x}_t, 0) $, thus we can rewrite  $\widehat{\sigma}_{T}^2 $ as
\begin{equation*}
\begin{aligned}
\widehat{\sigma}_{T,1}^2  & = \frac{1}{T}\sum_{t=1}^{T} \frac{\widehat{\sigma}_{1,t-1}^2  \mathbb{I}\{\widehat{\mu}_{t-1}(\boldsymbol{x}_t, 1)> \widehat{\mu}_{t-1}(\boldsymbol{x}_t, 0) \}+\widehat{\sigma}_{0,t-1}^2 (\boldsymbol{x}_t,0)  \mathbb{I}\{\widehat{\mu}_{t-1}(\boldsymbol{x}_t, 1)< \widehat{\mu}_{t-1}(\boldsymbol{x}_t, 0)\}}{1-\widehat{\kappa}_{t}(\boldsymbol{x}_t)} \\
& = \frac{1}{T}\sum_{t=1}^{T} \frac{\widehat{\sigma}_{1,t-1}^2  \mathbb{I}\{\widehat{\Delta}_{\boldsymbol{x}_t}>0 \}+\widehat{\sigma}_{0,t-1}^2 \mathbb{I}\{\widehat{\Delta}_{\boldsymbol{x}_t} <0 \}}{1-\widehat{\kappa}_{t}(\boldsymbol{x}_t)}.
\end{aligned}
\end{equation*}
We decompose the proposed variance estimator by
\begin{equation*}
\begin{aligned}
\widehat{\sigma}_{T,1}^2 
 & = \frac{1}{T}\sum_{t=1}^{T} \frac{ \left\{ \widehat{\sigma}_{1,t-1}^2  -\sigma_{1}^2 \right\} \mathbb{I}\{\widehat{\Delta}_{\boldsymbol{x}_t} >0 \}+ \left\{ \widehat{\sigma}_{0,t-1}^2  -\sigma_{0}^2  \right\} \mathbb{I}\{\widehat{\Delta}_{\boldsymbol{x}_t} <0 \}}{1-\widehat{\kappa}_{t}(\boldsymbol{x}_t)}  \\
  & + \frac{1}{T}\sum_{t=1}^{T} \frac{\sigma_{1}^2  \left(  \mathbb{I}\{\widehat{\Delta}_{\boldsymbol{x}_t} > 0 \} - \mathbb{I}\{\Delta_{\boldsymbol{x}_t} > 0 \}\right) +\sigma_{0}^2  \left(  \mathbb{I}\{\widehat{\Delta}_{\boldsymbol{x}_t} < 0\}-\mathbb{I}\{\Delta_{\boldsymbol{x}_t} < 0 \} \right)}{1-\widehat{\kappa}_{t}(\boldsymbol{x}_t)} \\
    & + \frac{1}{T}\sum_{t=1}^{T} \left( \sigma_{1}^2   \mathbb{I}\{\Delta_{\boldsymbol{x}_t} > 0 \}+\sigma_{0}^2   \mathbb{I}\{\Delta_{\boldsymbol{x}_t} < 0\} \right) \left\{\frac{1}{1-\widehat{\kappa}_{t}(\boldsymbol{x}_t)}  - \frac{1}{1-\kappa_{t}(\boldsymbol{x}_t)}  \right\}\\
       & + \frac{1}{T}\sum_{t=1}^{T} \frac{\sigma_{1}^2   \mathbb{I}\{\Delta_{\boldsymbol{x}_t} > 0 \}+\sigma_{0}^2   \mathbb{I}\{\Delta_{\boldsymbol{x}_t} < 0\}}{1-\kappa_{t}(\boldsymbol{x}_t)} .
\end{aligned}
\end{equation*}
Our goal is to prove that the first three lines are all $o_p(1)$.

Firstly,  recall that
\begin{equation*}
\begin{aligned}
 \widehat{\sigma}_{a,t}^2&  = \{\sum_{i=1}^{t}\mathbb{I}(a_i =a)-d\}^{-1}\sum_{a_i =a}^{1\leq i\leq t}[\widehat{\mu}_{i} \{\boldsymbol{x}_i ,a_i\} - r_i]^2 \\
 & = \{\sum_{i=1}^{t}\mathbb{I}(a_i =a)-d\}^{-1}\sum_{a_i =a}^{1\leq i\leq t}[\boldsymbol{x}_i^\top\left\{\widehat{\boldsymbol{\beta}}_{i-1}(a)-\boldsymbol{\beta}_{i-1}(a)\right\} - e_i]^2.
\end{aligned}
\end{equation*}
By Lemma 4.1,  we have $\|\widehat{\boldsymbol{\beta}}_{i-1}(a)-\boldsymbol{\beta}_{i-1}(a)\|_1 = o_p(1)$. Under Assumption 4.1, we have $\boldsymbol{x}_i^\top\left\{\widehat{\boldsymbol{\beta}}_{i-1}(a)-\boldsymbol{\beta}_{i-1}(a)\right\} = o_p(1)$. Thus by Lemma 6 in \cite{luedtke2016statistical}, we have 
\begin{equation*}
\begin{aligned}
 \widehat{\sigma}_{a,t}^2&  =  \{\sum_{i=1}^{t}\mathbb{I}(a_i =a)-d\}^{-1}\sum_{a_i =a}^{1\leq i\leq t} e_i^2+o_p(1).
\end{aligned}
\end{equation*}
Since $e_i$ is i.i.d conditional on $a_i$, and $\Mean(e_i^2|a_i = a) = \sigma_{a}^2$, noting that 
\begin{equation*}
\lim_{t\rightarrow \infty} \frac{\sum_{i=1}^{t}\mathbb{I}(a_i =a)}{\sum_{i=1}^{t}\mathbb{I}(a_i =a)-d} = 1,
\end{equation*}
by Law of Large Numbers we have
\begin{equation*}
 \widehat{\sigma}_{a,t}^2 =  \sigma_{a}^2+o_p(1).
\end{equation*}
Therefore, the first line is $o_p(1)$.

 Secondly, denote the second line as 
\begin{equation*}
\begin{aligned}
\psi_8  & =  \frac{1}{T}\sum_{t=1}^{T} \frac{\sigma_{1}^2  \left(  \mathbb{I}\{\widehat{\Delta}_{\boldsymbol{x}_t} > 0 \} - \mathbb{I}\{\Delta_{\boldsymbol{x}_t} > 0 \}\right) +\sigma_{0}^2  \left(  \mathbb{I}\{\widehat{\Delta}_{\boldsymbol{x}_t} < 0\}-\mathbb{I}\{\Delta_{\boldsymbol{x}_t} < 0 \} \right)}{1-\widehat{\kappa}_{t}(\boldsymbol{x}_t)} \\
& =  \frac{1}{T}\sum_{t=1}^{T} \frac{\sigma_{1}^2  \left(  \mathbb{I}\{\widehat{\Delta}_{\boldsymbol{x}_t} > 0 \} - \mathbb{I}\{\Delta_{\boldsymbol{x}_t} > 0 \}\right) +\sigma_{0}^2  \left( \mathbb{I}\{\Delta_{\boldsymbol{x}_t} > 0 \} - \mathbb{I}\{\widehat{\Delta}_{\boldsymbol{x}_t} > 0\}\right)}{1-\widehat{\kappa}_{t}(\boldsymbol{x}_t)} \\
& =  \frac{\sigma_{1}^2 -\sigma_{0}^2}{T}\sum_{t=1}^{T} \frac{  \mathbb{I}\{\widehat{\Delta}_{\boldsymbol{x}_t} > 0 \} - \mathbb{I}\{\Delta_{\boldsymbol{x}_t} > 0 \} }{1-\widehat{\kappa}_{t}(\boldsymbol{x}_t)}. \\
\end{aligned}
\end{equation*} 
Since $\kappa_{t}(\boldsymbol{x}_t) \leq C_2 <1 $ for sure for some constant $0< C_2 <1$ (by definition of a valid bandit algorithm an results for Theorem 1), by the triangle inequality, we have
\begin{equation*}
 \begin{aligned}
|\psi_8|  & \leq \frac{\sigma_{1}^2 -\sigma_{0}^2}{T}\sum_{t=1}^{T}  \frac{  \left |\mathbb{I}\{\widehat{\Delta}_{\boldsymbol{x}_t} > 0 \} - \mathbb{I}\{\Delta_{\boldsymbol{x}_t} > 0 \} \right|}{ \left |1-\widehat{\kappa}_{t}(\boldsymbol{x}_t)\right|}  \leq \frac{\sigma_{1}^2 -\sigma_{0}^2}{T}\sum_{t=1}^{T}  \frac{  \left |\mathbb{I}\{\widehat{\Delta}_{\boldsymbol{x}_t} > 0 \} - \mathbb{I}\{\Delta_{\boldsymbol{x}_t} > 0 \} \right|}{1-C_2}  \\
& \leq \frac{\sigma_{1}^2 -\sigma_{0}^2}{1-C_2} \frac{1}{T}\sum_{t=1}^{T}   \left |\mathbb{I}\{\widehat{\Delta}_{\boldsymbol{x}_t} > 0 \} - \mathbb{I}\{\Delta_{\boldsymbol{x}_t} > 0 \} \right|. 
\end{aligned}
\end{equation*}
Since $\prob\left( \mathbb{I}\{\widehat{\Delta}_{\boldsymbol{x}_t} > 0 \} - \mathbb{I}\{\Delta_{\boldsymbol{x}_t} > 0 \}=0\right) = 1-ct^{- \alpha \gamma}$ by Lemma \ref {lemma}, there exists some constant $c$ such that 
\begin{equation*}
\prob\left( \frac{1}{T}\sum_{t=1}^{T}   \left |\mathbb{I}\{\widehat{\Delta}_{\boldsymbol{x}_t} > 0 \} - \mathbb{I}\{\Delta_{\boldsymbol{x}_t} > 0 \} \right| \neq 0\right) \leq \sum_{t=1}^{T}  \prob\left(   \left |\mathbb{I}\{\widehat{\Delta}_{\boldsymbol{x}_t} > 0 \} - \mathbb{I}\{\Delta_{\boldsymbol{x}_t} > 0 \} \right| \neq 0\right) \leq \sum_{t=1}^{T} ct^{- \alpha \gamma}.
\end{equation*}
By Lemma 6 in \cite{luedtke2016statistical}, we have $\sum_{t=1}^{T} ct^{- \alpha \gamma} = cT^{- \alpha \gamma}$, thus 
\begin{equation*}
\prob\left( |\psi_8|  \neq 0\right) = \prob\left( \frac{1}{T}\sum_{t=1}^{T}   \left |\mathbb{I}\{\widehat{\Delta}_{\boldsymbol{x}_t} > 0 \} - \mathbb{I}\{\Delta_{\boldsymbol{x}_t} > 0 \} \right| \neq 0\right) \leq cT^{- \alpha \gamma},
\end{equation*}
which follows $\prob\left( |\psi_8|  \neq 0\right) = o_p(1)$.

 Lastly, under the assumption that $\widehat{\kappa}_{t}(\boldsymbol{x}_t)$ is a consistent estimator for $\kappa_{t}(\boldsymbol{x}_t)$, we have the third line is $o_p(1)$ by the continuous mapping theorem.

Given the above results, we have 
\begin{equation*}
\widehat{\sigma}_{T,1}^2  = \frac{1}{T}\sum_{t=1}^{T} \frac{\sigma_{1}^2   \mathbb{I}\{\Delta_{\boldsymbol{x}_t} > 0 \}+\sigma_{0}^2   \mathbb{I}\{\Delta_{\boldsymbol{x}_t} < 0\}}{1-\kappa_{t}(\boldsymbol{x}_t)} +o_p(1)= \frac{1}{T}\sum_{t=1}^{T} \frac{\sigma_{1}^2   \mathbb{I}\{\Delta_{\boldsymbol{x}_t} > 0 \}+\sigma_{0}^2   \mathbb{I}\{\Delta_{\boldsymbol{x}_t} < 0\}}{1-\operatorname{Pr}\left\{a_{t} \neq \widehat{\pi}_{t}\left(x_{t}\right)\right\}} +o_p(1).
\end{equation*}
Thus, we can further express $\widehat{\sigma}_{T,1}^2 $ as 
\begin{equation} \label{eq:sigmaT}
\begin{aligned}
\widehat{\sigma}_{T,1}^2 & = \frac{1}{T}\sum_{t=1}^{T} \left(\sigma_{1}^2   \mathbb{I}\{\Delta_{\boldsymbol{x}_t} > 0 \}+\sigma_{0}^2   \mathbb{I}\{\Delta_{\boldsymbol{x}_t} < 0\} \right) \left( \frac{1}{1-\operatorname{Pr}\left\{a_t \neq \widehat{\pi}_{t}\left(x\right)\right\}}-\frac{1}{1-\operatorname{Pr}\left\{a_t \neq \pi^{*}\left(\boldsymbol{x}\right)\right\}}\right) \\
& + \frac{1}{T}\sum_{t=1}^{T} \frac{\sigma_{1}^2   \mathbb{I}\{\Delta_{\boldsymbol{x}_t} > 0 \}+\sigma_{0}^2   \mathbb{I}\{\Delta_{\boldsymbol{x}_t} < 0\}}{1-\operatorname{Pr}\left\{a_{t} \neq \pi^{*}\left(\boldsymbol{x}_t\right)\right\}} +o_p(1). \\
\end{aligned}
\end{equation}
Notice that 
\begin{equation*}
\left( \frac{1}{1-\operatorname{Pr}\left\{a_t \neq \widehat{\pi}_{t}\left(x\right)\right\}}-\frac{1}{1-\operatorname{Pr}\left\{a_t \neq \pi^{*}\left(\boldsymbol{x}\right)\right\}}\right) = \frac{\operatorname{Pr}\left\{a_t \neq \widehat{\pi}_{t}\left(x\right)\right\}-\operatorname{Pr}\left\{a_{t} \neq \pi^{*}\left(\boldsymbol{x}_t\right)\right\}}{ \left(1-\operatorname{Pr}\left\{a_t \neq \widehat{\pi}_{t}\left(x\right)\right\}\right)\left(1-\operatorname{Pr}\left\{a_t \neq \pi^{*}\left(\boldsymbol{x}\right)\right\}\right)},
\end{equation*}
where
\begin{equation*}
\operatorname{Pr}\left\{a_t \neq \widehat{\pi}_{t}\left(x\right)\right\}-\operatorname{Pr}\left\{a_{t} \neq \pi^{*}\left(\boldsymbol{x}_t\right)\right\}  = \Mean \left(\mathbb{I}\{ a_t \neq \widehat{\pi}_{t}\left(x\right)\} - \mathbb{I}\{a_{t} \neq \pi^{*}\left(\boldsymbol{x}_t\right) \}\right) = \Mean \left(\mathbb{I}\{  \widehat{\pi}_{t}\left(x\right)\neq \pi^{*}\left(\boldsymbol{x}_t\right) \}\right).
\end{equation*}
We also note that by Lemma \ref{lemma}, there exists some constant $c$  and $0 < \alpha < \frac{1}{2}$ such that $\alpha \gamma < \frac{1}{2}$ and
\begin{equation*}
\Mean \left(\mathbb{I}\{  \widehat{\pi}_{t}\left(x\right)= \pi^{*}\left(\boldsymbol{x}_t\right) \}\right) = \prob\left(\widehat{\pi}_{t}\left(x\right)= \pi^{*}\left(\boldsymbol{x}_t\right) \} \right) \leq ct^{- \alpha \gamma},
\end{equation*}
therefore we have the result that
\begin{equation*}
\operatorname{Pr}\left\{a_t \neq \widehat{\pi}_{t}\left(x\right)\right\}-\operatorname{Pr}\left\{a_{t} \neq \pi^{*}\left(\boldsymbol{x}_t\right)\right\}  = \Mean \left(\mathbb{I}\{  \widehat{\pi}_{t}\left(x\right)\neq \pi^{*}\left(\boldsymbol{x}_t\right) \}\right) = o_p(1).
\end{equation*}
Thus, Equation \eqref{eq:sigmaT} can be expressed as
\begin{equation*} 
\begin{aligned}
\widehat{\sigma}_{T,1}^2 & = \frac{1}{T}\sum_{t=1}^{T} \left(\sigma_{1}^2   \mathbb{I}\{\Delta_{\boldsymbol{x}_t} > 0 \}+\sigma_{0}^2   \mathbb{I}\{\Delta_{\boldsymbol{x}_t} < 0\} \right) o_p(1) + \frac{1}{T}\sum_{t=1}^{T} \frac{\sigma_{1}^2   \mathbb{I}\{\Delta_{\boldsymbol{x}_t} > 0 \}+\sigma_{0}^2   \mathbb{I}\{\Delta_{\boldsymbol{x}_t} < 0\}}{1-\operatorname{Pr}\left\{a_{t} \neq \pi^{*}\left(\boldsymbol{x}_t\right)\right\}} +o_p(1) \\
& = \frac{1}{T}\sum_{t=1}^{T} \frac{\sigma_{1}^2   \mathbb{I}\{\Delta_{\boldsymbol{x}_t} > 0 \}+\sigma_{0}^2   \mathbb{I}\{\Delta_{\boldsymbol{x}_t} < 0\}}{1-\operatorname{Pr}\left\{a_{t} \neq \pi^{*}\left(\boldsymbol{x}_t\right)\right\}} +o_p(1) \\
& = \underbrace{\frac{1}{T}\sum_{t=1}^{T} \left\{\sigma_{1}^2   \mathbb{I}\{\Delta_{\boldsymbol{x}_t} > 0 \}+\sigma_{0}^2   \mathbb{I}\{\Delta_{\boldsymbol{x}_t} < 0\}\right\}\left[ \frac{1}{1-\operatorname{Pr}\left\{a_{t} \neq \pi^{*}\left(\boldsymbol{x}_t\right)\right\}}- \frac{1}{1-\kappa_{\infty}(\boldsymbol{x}_t)}\right]}_{\psi_9}\\
&+\frac{1}{T}\sum_{t=1}^{T} \frac{\sigma_{1}^2   \mathbb{I}\{\Delta_{\boldsymbol{x}_t} > 0 \}+\sigma_{0}^2   \mathbb{I}\{\Delta_{\boldsymbol{x}_t} < 0\}}{1-\kappa_{\infty}(\boldsymbol{x}_t)} +o_p(1).\\
\end{aligned}
\end{equation*}
Note that 
\begin{equation*}
\begin{aligned}
\psi_9 & = \frac{1}{T}\sum_{t=1}^{T} \left\{\sigma_{1}^2   \mathbb{I}\{\Delta_{\boldsymbol{x}_t} > 0 \}+\sigma_{0}^2   \mathbb{I}\{\Delta_{\boldsymbol{x}_t} < 0\}\right\}\left[ \frac{1}{1-\operatorname{Pr}\left\{a_{t} \neq \pi^{*}\left(\boldsymbol{x}_t\right)\right\}}- \frac{1}{1-\kappa_{\infty}(\boldsymbol{x}_t)}\right]\\
& = \frac{1}{T}\sum_{t=1}^{T} \left\{\sigma_{1}^2   \mathbb{I}\{\Delta_{\boldsymbol{x}_t} > 0 \}+\sigma_{0}^2   \mathbb{I}\{\Delta_{\boldsymbol{x}_t} < 0\}\right\} \frac{\kappa_{\infty}(\boldsymbol{x}_t)-\operatorname{Pr}\left\{a_{t} \neq \pi^{*}\left(\boldsymbol{x}_t\right)\right\}}{\left[1-\operatorname{Pr}\left\{a_{t} \neq \pi^{*}\left(\boldsymbol{x}_t\right)\right\}\right]\left[1-\kappa_{\infty}(\boldsymbol{x}_t)\right]},
\end{aligned}
\end{equation*}
which follows
\begin{equation*}
\begin{aligned}
|\psi_9 |
& \leq \frac{1}{T}\sum_{t=1}^{T} \left\{\sigma_{1}^2   \mathbb{I}\{\Delta_{\boldsymbol{x}_t} > 0 \}+\sigma_{0}^2   \mathbb{I}\{\Delta_{\boldsymbol{x}_t} < 0\}\right\} \frac{\left|\kappa_{\infty}(\boldsymbol{x}_t)-\operatorname{Pr}\left\{a_{t} \neq \pi^{*}\left(\boldsymbol{x}_t\right)\right\}\right|}{\left[1-\operatorname{Pr}\left\{a_{t} \neq \pi^{*}\left(\boldsymbol{x}_t\right)\right\}\right]\left[1-\kappa_{\infty}(\boldsymbol{x}_t)\right]}.
\end{aligned}
\end{equation*}
By Equation \eqref{eq:ptbound}, for large enough $t$,  there exists some constant $C_1>0$ such that
\begin{equation*}
    \prob\{a_t\neq{\pi}^*(\boldsymbol{x}_t)\} < C_1.
\end{equation*}
Since $\kappa_{t}(\boldsymbol{x}_t) \leq C_2 <1 $ for sure for some constant $0< C_2 <1$ (by the definition of a valid bandit algorithm as results shown in Theorem 1), we also have 
\begin{equation*}
    \kappa_{\infty}(\boldsymbol{x}_t)<  C_2.
\end{equation*}
Therefore,
\begin{equation*}
\frac{\sigma_{1}^2   \mathbb{I}\{\Delta_{\boldsymbol{x}_t} > 0 \}+\sigma_{0}^2   \mathbb{I}\{\Delta_{\boldsymbol{x}_t} < 0\}}{\left[1-\operatorname{Pr}\left\{a_{t} \neq \pi^{*}\left(\boldsymbol{x}_t\right)\right\}\right]\left[1-\kappa_{\infty}(\boldsymbol{x}_t)\right]} < \frac{\max\{\sigma_{0}^2  ,\sigma_{1}^2  \} }{(1-C_1)(1-C_2)} \triangleq C,
\end{equation*}
which follows immediately that 
\begin{equation*}
\begin{aligned}
|\psi_9 |
& \leq \frac{1}{T}\sum_{t=1}^{T} C \left|\kappa_{\infty}(\boldsymbol{x}_t)-\operatorname{Pr}\left\{a_{t} \neq \pi^{*}\left(\boldsymbol{x}_t\right)\right\}\right|.
\end{aligned}
\end{equation*}
Since $\lim_{t\rightarrow \infty}\operatorname{Pr}\left\{a_{t} \neq \pi^{*}\left(\boldsymbol{x}\right)\right\} = \kappa_{\infty}(\boldsymbol{x})$ for any $\boldsymbol{x}$, therefore for any $\epsilon>0$, there exists some constant $T_0$, such that $\left| \operatorname{Pr}\left\{a_{t} \neq \pi^{*}\left(\boldsymbol{x}\right)\right\} - \kappa_{t}(\boldsymbol{x})\right| < \epsilon$ for any $\boldsymbol{x}$ with $t \geq T_0$, thus we have 
\begin{equation*}
\begin{aligned}
|\psi_9 |
& \leq \frac{1}{T}\sum_{t=1}^{T} C \left|\kappa_{\infty}(\boldsymbol{x}_t)-\operatorname{Pr}\left\{a_{t} \neq \pi^{*}\left(\boldsymbol{x}_t\right)\right\}\right|\\
& = \frac{1}{T}\sum_{t=1}^{T_0} C \left|\kappa_{\infty}(\boldsymbol{x}_t)-\operatorname{Pr}\left\{a_{t} \neq \pi^{*}\left(\boldsymbol{x}_t\right)\right\}\right| + \frac{1}{T}\sum_{t=T_0}^{T} C \left|\kappa_{\infty}(\boldsymbol{x}_t)-\operatorname{Pr}\left\{a_{t} \neq \pi^{*}\left(\boldsymbol{x}_t\right)\right\}\right|\\
& < \frac{1}{T}\sum_{t=1}^{T_0} C \left|\kappa_{\infty}(\boldsymbol{x}_t)-\operatorname{Pr}\left\{a_{t} \neq \pi^{*}\left(\boldsymbol{x}_t\right)\right\}\right| + \frac{1}{T}\sum_{t=T_0}^{T} C \epsilon.
\end{aligned}
\end{equation*}
Note that by the triangle inequality,
\begin{equation*}
\left|\kappa_{\infty}(\boldsymbol{x}_t)-\operatorname{Pr}\left\{a_{t} \neq \pi^{*}\left(\boldsymbol{x}_t\right)\right\}\right| \leq \left|\kappa_{\infty}(\boldsymbol{x}_t)\right|+ \left|\operatorname{Pr}\left\{a_{t} \neq \pi^{*}\left(\boldsymbol{x}_t\right)\right\}\right| < C_1 + C_2, 
\end{equation*}
thus we have
\begin{equation*}
|\psi_9 |
 < \frac{1}{T}\sum_{t=1}^{T_0} C \left(C_1 + C_2\right) + \frac{1}{T}\sum_{t=T_0}^{T} C \epsilon =\frac{T_0C \left(C_1 + C_2\right)}{T}  + \frac{T-T0}{T}C \epsilon.
\end{equation*}
Since the above equation holds for any $\epsilon>0$, we have 
\begin{equation*}
|\psi_9 | \leq  \frac{1}{T}\sum_{t=1}^{T_0} C \left(C_1 + C_2\right) + \frac{1}{T}\sum_{t=T_0}^{T} C \epsilon =\frac{T_0C \left(C_1 + C_2\right)}{T} = \mathcal{O}(\frac{1}{T}),
\end{equation*}
which follows $\psi_9 =0_p(1)$. Therefore,  we have
\begin{equation*}
\widehat{\sigma}_{T,1}^2  = \frac{1}{T}\sum_{t=1}^{T} \frac{\sigma_{1}^2   \mathbb{I}\{\Delta_{\boldsymbol{x}_t} > 0 \}+\sigma_{0}^2   \mathbb{I}\{\Delta_{\boldsymbol{x}_t} < 0\}}{1-\kappa_{\infty}(\boldsymbol{x}_t)} +o_p(1).
\end{equation*}
By Law of Large Numbers, we further have
\begin{equation*}
\widehat{\sigma}_{T,1}^2  = \int_{\boldsymbol{x}} \frac{\sigma_{1}^2 \mathbb{I}\{\mu(\boldsymbol{x}, 1)> \mu(\boldsymbol{x}, 0) \}+\sigma_{0}^2 \mathbb{I}\{\mu(\boldsymbol{x}, 1)< \mu(\boldsymbol{x}, 0)\}}{1-\kappa_\infty(\boldsymbol{x})} d P_{\mathcal{X}} +o_p(1).
\end{equation*}
Next, we proof the second  line of Equation \eqref{sigma_dr_hat} is a consistent  estimator for $\Var \left[  {\mu} \{\boldsymbol{x} ,{\pi}^*(\boldsymbol{x} ) \} \right]$. By Central Limit Theorem and Continuous Mapping Theorem, we have
\begin{equation*}
\widehat{\sigma}_{T,2}^2  = \frac{1}{T}\sum_{t=1}^{T} \left[  \widehat{\mu}_{T} \{\boldsymbol{x}_t ,\widehat{\pi}_{T}(\boldsymbol{x}_t ) \} -  \frac{1}{T}\sum_{t=1}^{T} \widehat{\mu}_{T} \{\boldsymbol{x}_t ,\widehat{\pi}_{T}(\boldsymbol{x}_t ) \}  \right]^2 .
\end{equation*}
Since $\boldsymbol{x}_t$ are i.i.d, $ \widehat{\mu}_{T} \{\boldsymbol{x}_t ,\widehat{\pi}_{T}(\boldsymbol{x}_t ) \} $ are i.i.d as well. Thus by the Law of Large Numbers, we have
\begin{equation*}
\frac{1}{T}\sum_{t=1}^{T} \widehat{\mu}_{T} \{\boldsymbol{x}_t ,\widehat{\pi}_{T}(\boldsymbol{x}_t ) \}  = \Mean \left[  \widehat{\mu}_{T} \{\boldsymbol{x} ,\widehat{\pi}_{T}(\boldsymbol{x} ) \} \right] + o_p(1),
\end{equation*}
and
\begin{equation*}
\widehat{\sigma}_{T,2}^2  = \Var \left[  \widehat{\mu}_{T} \{\boldsymbol{x} ,\widehat{\pi}_{T}(\boldsymbol{x} ) \} \right] + o_p(1).
\end{equation*}
Note that 
\begin{equation*}
 \widehat{\mu}_{T} \{\boldsymbol{x} ,\widehat{\pi}_{T}(\boldsymbol{x} ) \} =   \mathbb{I}\{\widehat{\Delta}_{\boldsymbol{x}} > 0 \} \widehat{\mu}_{T} \{\boldsymbol{x} ,1 \} + \left[1- \mathbb{I}\{\widehat{\Delta}_{\boldsymbol{x}} > 0 \}\right] \widehat{\mu}_{T} \{\boldsymbol{x} ,0 \},
\end{equation*}
since $  \mathbb{I}\{\widehat{\Delta}_{\boldsymbol{x}} > 0 \} - \mathbb{I}\{\Delta_{\boldsymbol{x}} > 0 \} = o_p(1)$ and the fact that $\widehat{\mu}_{T}\{\boldsymbol{x} ,0 \}$ and $\widehat{\mu}_{T}\{\boldsymbol{x} ,1 \}$ are consistent, we have 
\begin{equation*}
\widehat{\mu}_{T} \{\boldsymbol{x} ,\widehat{\pi}_{T}(\boldsymbol{x} ) \}=   \mathbb{I}\{\Delta_{\boldsymbol{x}} > 0 \} \mu \{\boldsymbol{x} ,1 \} + \left[1- \mathbb{I}\{\Delta_{\boldsymbol{x}} > 0 \}\right] \mu\{\boldsymbol{x} ,0 \} = \mu \{\boldsymbol{x} ,\pi(\boldsymbol{x} ) \} +  o_p(1).
\end{equation*}
Therefore 
\begin{equation*}
\widehat{\sigma}_{T,2}^2  = \Var \left[ \mu \{\boldsymbol{x} ,\pi(\boldsymbol{x} ) \} +  o_p(1)\right] = \Var \left[ \mu \{\boldsymbol{x} ,\pi(\boldsymbol{x} ) \}\right] +  o_p(1).
\end{equation*}
by Continuous Mapping Theorem. The proof for Theorem 3 is thus completed.

\subsection{Results and Proof for Auxiliary Lemmas}

\begin{lemma}\label{lemma:bounded}
Suppose a random variable  $X$ is restricted to $[a, b]$ and $\mu=E[X]$, then the variance of  $X$ is bounded by $(b-\mu)(\mu-a)$.
\end{lemma}

\textbf{Proof:} 
Firstly consider the case that $a=0, b=1$. Notice that we have $E\left[X^2\right] \leq E[X]$ since for all $x \in[0,1], x^2 \leq x$. Therefore, 
\begin{equation*}
\operatorname{Var}[X]=E\left[X^2\right]-\left(E[X]^2\right)=E\left[X^2\right]-\mu^2 \leq \mu-\mu^2=\mu(1-\mu).
\end{equation*}
Then we consider general interval $[a, b]$. Define $Y=\frac{X-a}{b-a}$, which is restricted in $[0,1]$. Equivalently, $X=(b-a) Y+a$, which follows immediate that \begin{equation*}
\operatorname{Var}[X]=(b-a)^2 \operatorname{Var}[Y] \leq(b-a)^2 \mu_X\left(1-\mu_Y\right),
\end{equation*}
where the inequality is based on the first result. Now, by substituting $\mu_Y=\frac{\mu_X-a}{b-a}$, the bound equals
\begin{equation*}
(b-a)^2 \frac{\mu_X-a}{b-a}\left(1-\frac{\mu_X-a}{b-a}\right)=(b-a)^2 \frac{\mu_X-a}{b-a} \frac{b-\mu_X}{b-a}=\left(\mu_X-a\right)\left(b-\mu_X\right),
\end{equation*}
which is the desired result.

\begin{lemma} \label{lemma}
Suppose the conditions in Theorem  2 hold with Assumption 4.3, then there exists some constant $c$  and $0 < \alpha < \frac{1}{2}$ such that $\alpha \gamma < \frac{1}{2}$ and $\prob\left(\widehat{\pi}_{t}\left(x\right)\neq \pi^{*}\left\{\boldsymbol{x}_t\right)\right \}  \geq 1-ct^{- \alpha \gamma}$.
\end{lemma}
\textbf{Proof:} By Theorem 2, we have $\widehat{\Delta}_{\boldsymbol{x}_t}  - \Delta_{\boldsymbol{x}_t} =\mathcal{O}_p(t^{-\frac{1}{2}})$, thus $ \mathbb{I}\{\widehat{\Delta}_{\boldsymbol{x}_t} > 0 \}  =  \mathbb{I}\{ \Delta_{\boldsymbol{x}_t} + \mathcal{O}_p(t^{-\frac{1}{2}})> 0 \} $. \\ By Assumption 4.3, there exists some constant $c$  and $0 < \alpha < \frac{1}{2}$ such that $\alpha \gamma < \frac{1}{2}$ and
\begin{equation*}
\prob\{0 <  |\Delta_{\boldsymbol{x}_t}|  < t^{-\alpha} \} \leq ct^{- \alpha \gamma}.
\end{equation*}
Thus, with probability greater than $1-ct^{- \alpha \gamma}$, we have $|\Delta_{\boldsymbol{x}_t}| > t^{-\alpha} $, which further implies $ \mathbb{I}\{ \Delta_{\boldsymbol{x}_t} + \mathcal{O}_p(t^{-\frac{1}{2}})> 0 \}  =  \mathbb{I}\{ \Delta_{\boldsymbol{x}_t} > 0 \} $. In other words, $  \mathbb{I}\{\widehat{\Delta}_{\boldsymbol{x}_t} > 0 \} - \mathbb{I}\{\Delta_{\boldsymbol{x}_t} > 0 \} = 0$ with probability  greater than $1-ct^{- \alpha \gamma}$, which convergences to 1 as $t \rightarrow \infty$. Therefore, as $t \rightarrow \infty$, we have
\begin{equation*}
\prob\left( \mathbb{I}\{\widehat{\Delta}_{\boldsymbol{x}_t} > 0 \} - \mathbb{I}\{\Delta_{\boldsymbol{x}_t} > 0 \}=0\right) \geq  1-ct^{- \alpha \gamma},
\end{equation*}
i.e.,
\begin{equation*}
\prob\left(\widehat{\pi}_{t}\left(x\right)= \pi^{*}\left\{\boldsymbol{x}_t\right)  \right\} \geq 1-ct^{- \alpha \gamma} .
\end{equation*}
\begin{lemma} \label{lemma2}
Suppose conditions in Lemma 4.1 hold. Assuming Assumptions 4.3 with $tp_t^2 \rightarrow \infty$ as $t\rightarrow \infty$, we have 
\begin{equation*}
 { {T}^{-1/2} } \sum_{t=1}^T\left|\mu\{\boldsymbol{x}_t, \widehat{\pi}_t(\boldsymbol{x}_t ) \} - \mu\{\boldsymbol{x}_t, \pi^*\} \right|= o_p (1).
\end{equation*}
\end{lemma}

\textbf{Proof:} Without loss of generality, suppose $\widehat{\pi}_t(\boldsymbol{x}_t ) =\mathbb{I}\{\boldsymbol{x}_t^{\top}  \boldsymbol{\beta}(1) >\boldsymbol{x}_t^{\top} \boldsymbol{\beta}(0) \} $. Since $\mu(\boldsymbol{x}_t, a) =\boldsymbol{x}_t^{\top}  \boldsymbol{\beta}(a)  $, we have
\begin{equation}\label{thm4_pf_step2_e1}
\begin{aligned}
 \mu(\boldsymbol{x}_t, \widehat{\pi}_t(\boldsymbol{x}_t )
 & = \mathbb{I}\{\boldsymbol{x}_t^{\top} \widehat{\boldsymbol{\beta}}_t(1) >\boldsymbol{x}_t^{\top}\widehat{\boldsymbol{\beta}}_t(0) \} \boldsymbol{x}_t^{\top}  \boldsymbol{\beta}(1)  +\mathbb{I}\{\boldsymbol{x}_t^{\top} \widehat{\boldsymbol{\beta}}_t(1) \leq \boldsymbol{x}_t^{\top}\widehat{\boldsymbol{\beta}}_t(0) \} \boldsymbol{x}_t^{\top}  \boldsymbol{\beta}(0)  \\
  & = \mathbb{I}\{\boldsymbol{x}_t^{\top} \widehat{\boldsymbol{\beta}}_t(1) >\boldsymbol{x}_t^{\top}\widehat{\boldsymbol{\beta}}_t(0) \} \boldsymbol{x}_t^{\top} \left\{  \boldsymbol{\beta}(1)- \boldsymbol{\beta}(0)\right\}+\boldsymbol{x}_t^{\top}  \boldsymbol{\beta}(0). 
\end{aligned}
\end{equation}
Similarly to \eqref{thm4_pf_step2_e1}, we have
\begin{equation}\label{thm4_pf_step2_e2}
\begin{aligned}
\mu(\boldsymbol{x}_t, \pi^*) = \mathbb{I}\{\boldsymbol{x}_t^{\top}  \boldsymbol{\beta}(1) >\boldsymbol{x}_t^{\top} \boldsymbol{\beta}(0) \} \boldsymbol{x}_t^{\top} \left\{  \boldsymbol{\beta}(1)- \boldsymbol{\beta}(0)\right\}+\boldsymbol{x}_t^{\top}  \boldsymbol{\beta}(0).
\end{aligned}
\end{equation}
Combining \eqref{thm4_pf_step2_e1} and \eqref{thm4_pf_step2_e2}, we have
\begin{equation}\label{thm4_pf_step2_e3}
 \mu(\boldsymbol{x}_t, \widehat{\pi}_t(\boldsymbol{x}_t ) - \mu(\boldsymbol{x}_t, \pi^*) = \left[\mathbb{I}\{\boldsymbol{x}_t^{\top} \widehat{\boldsymbol{\beta}}_t(1) >\boldsymbol{x}_t^{\top}\widehat{\boldsymbol{\beta}}_t(0) \} - \mathbb{I}\{\boldsymbol{x}_t^{\top}  \boldsymbol{\beta}(1) >\boldsymbol{x}_t^{\top} \boldsymbol{\beta}(0) \}  \right]\boldsymbol{x}_t^{\top} \left\{  \boldsymbol{\beta}(1)- \boldsymbol{\beta}(0)\right\}.  
\end{equation}
Since $ \mathbb{I}\{\boldsymbol{x}_t^{\top}  \boldsymbol{\beta}(1) >\boldsymbol{x}_t^{\top} \boldsymbol{\beta}(0) \} = 1$ by assumption, \eqref{thm4_pf_step2_e3} can be simplified as
\begin{equation*}
\mu(\boldsymbol{x}_t, \widehat{\pi}_t(\boldsymbol{x}_t ) - \mu(\boldsymbol{x}_t, \pi^*)
= -\mathbb{I}\{\boldsymbol{x}_t^{\top} \widehat{\boldsymbol{\beta}}_t(1) -\boldsymbol{x}_t^{\top}\widehat{\boldsymbol{\beta}}_t(0) \leq 0\}\boldsymbol{x}_t^{\top} \left\{  \boldsymbol{\beta}(1)- \boldsymbol{\beta}(0)\right\}= -\mathbb{I}\{\widehat{\Delta}_{\boldsymbol{x}_t} \leq 0\}\Delta_{\boldsymbol{x}_t} \leq 0,
\end{equation*}
where $\Delta_{\boldsymbol{x}_t} = \boldsymbol{x}_t^{\top}  \{  \boldsymbol{\beta}(1)- \boldsymbol{\beta}(0) \}$ and $\widehat{\Delta}_{\boldsymbol{x}_t} = \boldsymbol{x}_t^{\top} \{  \widehat{\boldsymbol{\beta}}(1)- \widehat{\boldsymbol{\beta}}(0) \}$. Thus, we have
\begin{equation*}
 { {T}^{-1/2} } \sum_{t=1}^T\left|\mu\{\boldsymbol{x}_t, \widehat{\pi}_t(\boldsymbol{x}_t ) \} - \mu\{\boldsymbol{x}_t, \pi^*\} \right| = \frac{1}{\sqrt{T}}  \sum_{t=1}^T  \mathbb{I}\{\widehat{\Delta}_{\boldsymbol{x}_t} \leq 0\}\Delta_{\boldsymbol{x}_t}.
\end{equation*}
To show $ { {T}^{-1/2} } \sum_{t=1}^T\left|\mu\{\boldsymbol{x}_t, \widehat{\pi}_t(\boldsymbol{x}_t ) \} - \mu\{\boldsymbol{x}_t, \pi^*\} \right| =  o_p (1)$,  it suffices to show that  $(1/ \sqrt{T}) \sum_{t=1}^T \mathbb{I}\{\widehat{\Delta}_{\boldsymbol{x}_t} \leq 0\}\Delta_{\boldsymbol{x}_t} $ has an  upper bound. Since $\mathbb{I}\{\widehat{\Delta}_{\boldsymbol{x}_t} \leq 0\}\widehat{\Delta}_{\boldsymbol{x}_t} \leq 0$, it suffices to show 
\begin{equation*}
\zeta = \frac{1}{\sqrt{T}}  \sum_{t=1}^T \mathbb{I}\{\widehat{\Delta}_{\boldsymbol{x}_t} \leq 0\}(\widehat{\Delta}_{\boldsymbol{x}_t} - \Delta_{\boldsymbol{x}_t} )
\end{equation*}
has a  lower bound. We further notice that for any $\alpha>0$,
\begin{equation}\label{proof_t4_zeta}
\begin{aligned} 
 \zeta & = \underbrace{\prob\{0 <  \Delta_{\boldsymbol{x}_t}  < T^{-\alpha} \}\frac{1}{\sqrt{T}}  \sum_{t=1}^T \mathbb{I}\{0 <  \Delta_{\boldsymbol{x}_t}  < T^{-\alpha} \}\mathbb{I}\{\widehat{\Delta}_{\boldsymbol{x}_t} \leq 0\}(\widehat{\Delta}_{\boldsymbol{x}_t} - \Delta_{\boldsymbol{x}_t} ) }_{\zeta_1} \\
& +  \underbrace{\prob\{\Delta_{\boldsymbol{x}_t} > T^{-\alpha} \} \frac{1}{\sqrt{T}}  \sum_{t=1}^T \mathbb{I}\{\Delta_{\boldsymbol{x}_t} > T^{-\alpha} \} \mathbb{I}\{\widehat{\Delta}_{\boldsymbol{x}_t} \leq 0\}(\widehat{\Delta}_{\boldsymbol{x}_t} - \Delta_{\boldsymbol{x}_t} ) }_{\zeta_2} .
\end{aligned}
\end{equation}
To show $\zeta$ has a  lower bound, it is sufficient to show  $\zeta_1 = o_p (1) $ and $\zeta_2 = o_p (1)$ correspondingly.

Firstly, we are going to show $\zeta_1 =  o_p (1)$. By Theorem  2, $\widehat{\Delta}_{\boldsymbol{x}_t} - \Delta_{\boldsymbol{x}_t} = \mathcal{O}_p (t^{-\frac{1}{2}})$, which implies
\begin{equation*}
 \widehat{\Delta}_{\boldsymbol{x}_t} - \Delta_{\boldsymbol{x}_t} = o_p \{t^{-(\frac{1}{2}- \alpha \gamma)}\}.  
\end{equation*}
Thus we have
\begin{equation*}
\begin{aligned}
 & |\frac{1}{\sqrt{T}}  \sum_{t=1}^T \mathbb{I}\{0 <  \Delta_{\boldsymbol{x}_t}  < T^{-\alpha} \}\mathbb{I}\{\widehat{\Delta}_{\boldsymbol{x}_t} \leq 0\}(\widehat{\Delta}_{\boldsymbol{x}_t} - \Delta_{\boldsymbol{x}_t} )| \\
  & \leq \frac{1}{\sqrt{T}}  \sum_{t=1}^T |\mathbb{I}\{0 <  \Delta_{\boldsymbol{x}_t}  < T^{-\alpha} \}||\mathbb{I}\{\widehat{\Delta}_{\boldsymbol{x}_t} \leq 0\}||(\widehat{\Delta}_{\boldsymbol{x}_t} - \Delta_{\boldsymbol{x}_t} )| \\
 & \leq \frac{1}{\sqrt{T}}  \sum_{t=1}^T |(\widehat{\Delta}_{\boldsymbol{x}_t} - \Delta_{\boldsymbol{x}_t} )| \leq   \frac{\sqrt{T}}{T} \sum_{t=1}^T o_p (t^{-(\frac{1}{2}- \alpha \gamma)}) \overset{*}{=} \sqrt{T} o_p (T^{-(\frac{1}{2}- \alpha \gamma)})=o_p (T^{\alpha \gamma}), 
\end{aligned}  
\end{equation*}
where the equation (*) is derived by Lemma 6 in \cite{luedtke2016statistical}.
By Assumption 4.3, there exists some constant $c$  and $0 < \alpha < \frac{1}{2}$ such that $\alpha \gamma < \frac{1}{2}$ and
\begin{equation*}
\prob\{0 <  \Delta_{\boldsymbol{x}_t}  < T^{-\alpha} \} \leq cT^{- \alpha \gamma}.
\end{equation*}
Therefore we have
\begin{equation}\label{proof_t4_e1}
|\zeta_1| \leq \prob\{0 <  \Delta_{\boldsymbol{x}_t}  < T^{-\alpha} \}o_p (T^{-(\frac{1}{2}- \alpha \gamma)}) = cT^{- \alpha \gamma}  o_p (T^{\alpha \gamma})  =o_p(1).
\end{equation}
Notice that
\begin{equation} \label{proof_t4_e2}
\mathbb{I}\{0 <  \Delta_{\boldsymbol{x}_t}  < T^{-\alpha} \}\mathbb{I}\{\widehat{\Delta}_{\boldsymbol{x}_t} \leq 0\}(\widehat{\Delta}_{\boldsymbol{x}_t} - \Delta_{\boldsymbol{x}_t} ) \leq \mathbb{I}\{\widehat{\Delta}_{\boldsymbol{x}_t} \leq 0\}\widehat{\Delta}_{\boldsymbol{x}_t} \leq 0,
\end{equation}
where the first inequality holds since $\Delta_{\boldsymbol{x}_t}\geq 0$. Combining \eqref{proof_t4_zeta}, \eqref{proof_t4_e1}, and  \eqref{proof_t4_e2}, we have
\begin{equation} \label{proof_t4_zeta1}
0 \geq \zeta_1 =  o_p(1).
\end{equation}
Next, we consider the second part $\zeta_2$. Note that
\begin{equation*}
\mathbb{I}\{\widehat{\Delta}_{\boldsymbol{x}_t} \leq 0\} = \mathbb{I}\{\widehat{\Delta}_{\boldsymbol{x}_t} - \Delta_{\boldsymbol{x}_t} \leq -\Delta_{\boldsymbol{x}_t} \} = \mathbb{I}\{|\widehat{\Delta}_{\boldsymbol{x}_t} - \Delta_{\boldsymbol{x}_t}| >  \Delta_{\boldsymbol{x}_t}\},
\end{equation*}
we have
\begin{equation}\label{thm4_pf_step2_e4}
\left| \mathbb{I}\{\widehat{\Delta}_{\boldsymbol{x}_t} \leq 0\}(\widehat{\Delta}_{\boldsymbol{x}_t} - \Delta_{\boldsymbol{x}_t} ) \right| =\mathbb{I}\{|\widehat{\Delta}_{\boldsymbol{x}_t} - \Delta_{\boldsymbol{x}_t}| >  \Delta_{\boldsymbol{x}_t}\} | \widehat{\Delta}_{\boldsymbol{x}_t} - \Delta_{\boldsymbol{x}_t} | \leq  \frac{| \widehat{\Delta}_{\boldsymbol{x}_t} - \Delta _{\boldsymbol{x}_t}| ^2}{\Delta_{\boldsymbol{x}_t}},
\end{equation}
where the inequality holds since 
\begin{equation*}
   \mathbb{I}\{|\widehat{\Delta}_{\boldsymbol{x}_t} - \Delta_{\boldsymbol{x}_t}| >  \Delta_{\boldsymbol{x}_t}\} \leq  \frac{| \widehat{\Delta}_{\boldsymbol{x}_t} - \Delta _{\boldsymbol{x}_t}| }{\Delta_{\boldsymbol{x}_t}}.
\end{equation*}
Thus, by \eqref{thm4_pf_step2_e4}, we further have
\begin{equation}\label{thm4_pf_step2_e5}
\mathbb{I}\{\widehat{\Delta}_{\boldsymbol{x}_t} \leq 0\}(\widehat{\Delta}_{\boldsymbol{x}_t} - \Delta_{\boldsymbol{x}_t} )  \geq - \frac{(\widehat{\Delta}_{\boldsymbol{x}_t} - \Delta_{\boldsymbol{x}_t})^2}{\Delta_{\boldsymbol{x}_t}}.
\end{equation}
Since $\widehat{\Delta}_{\boldsymbol{x}_t} - \Delta_{\boldsymbol{x}_t} <0$, based on \eqref{thm4_pf_step2_e5}, we have
\begin{equation}\label{thm4_pf_step2_e6}
0 \geq \zeta_2  \geq \frac{1}{\sqrt{T}} \sum_{t=1}^T \mathbb{I}\{\Delta_{\boldsymbol{x}_t} > T^{-\alpha} \} \mathbb{I}\{\widehat{\Delta}_{\boldsymbol{x}_t} \leq 0\}(\widehat{\Delta}_{\boldsymbol{x}_t} - \Delta_{\boldsymbol{x}_t} ) \geq - \frac{1}{\sqrt{T}}  \sum_{t=1}^T\mathbb{I}\{\Delta_{\boldsymbol{x}_t}  > T^{-\alpha} \} \frac{(\widehat{\Delta}_{\boldsymbol{x}_t} - \Delta_{\boldsymbol{x}_t})^2}{\Delta_{\boldsymbol{x}_t}} .
\end{equation}
Notice that $\mathbb{I}\{\Delta_{\boldsymbol{x}_t}  > T^{-\alpha} \} \leq { \Delta_{\boldsymbol{x}_t}}{T^{\alpha}}$, combining with \eqref{thm4_pf_step2_e6}, we further have
\begin{equation*}
0 \geq \zeta_2 \geq - \frac{1 }{\Delta_{\boldsymbol{x}_t} }\frac{1}{\sqrt{T}}  \sum_{t=1}^T \frac{ \Delta_{\boldsymbol{x}_t}}{T^{-\alpha}}(\widehat{\Delta}_{\boldsymbol{x}_t} - \Delta_{\boldsymbol{x}_t})^2 = - T^{-\frac{1}{2}+\alpha}\sum_{t=1}^T (\widehat{\Delta}_{\boldsymbol{x}_t} - \Delta_{\boldsymbol{x}_t})^2= - T^{\frac{1}{2}+\alpha} \left\{ \frac{1}{T}  \sum_{t=1}^T (\widehat{\Delta}_{\boldsymbol{x}_t} - \Delta_{\boldsymbol{x}_t})^2 \right\}.
\end{equation*}
By Theorem 2, $\widehat{\Delta}_{\boldsymbol{x}_t} - \Delta_{\boldsymbol{x}_t} = \mathcal{O}_p (t^{-\frac{1}{2}})$, which implies $(\widehat{\Delta}_{\boldsymbol{x}_t} - \Delta_{\boldsymbol{x}_t})^2 = o_p (T^{-(\frac{1}{2}+\alpha)})$.And by Lemma 6 in \cite{luedtke2016statistical}, $T^{-1} \sum_{t=1}^T o_p \{t^{-(\frac{1}{2}+\alpha)}\} = o_p \{T^{-(\frac{1}{2}+\alpha)}\}$, we have
\begin{equation}\label{proof_t4_zeta2}
0 \geq \zeta_2 \geq - T^{\frac{1}{2}+\alpha} o_p (T^{-(\frac{1}{2}+\alpha)}) = o_{p}(1) .
\end{equation}
Therefore, combining  Equation \eqref{proof_t4_zeta1} and Equation \eqref{proof_t4_zeta2}, we have 
\begin{equation*}
 0 \geq \zeta= \zeta_1 + \zeta_2 =  o_{p}(1).
\end{equation*}
Thus, we have
\begin{equation*}
 { {T}^{-1/2} } \sum_{t=1}^T\left|\mu\{\boldsymbol{x}_t, \widehat{\pi}_t(\boldsymbol{x}_t ) \} - \mu\{\boldsymbol{x}_t, \pi^*\} \right|=  o_{p}(1).
\end{equation*}

\end{document}